\documentclass[dvipsnames]{article}
    
    \usepackage[eandd, final]{v4_neurips_2026/neurips_2026}

    \usepackage[utf8]{inputenc} % allow utf-8 input
    \usepackage[T1]{fontenc}    % use 8-bit T1 fonts
    \usepackage{hyperref}       % hyperlinks
    \usepackage{url}            % simple URL typesetting
    \usepackage{booktabs}       % professional-quality tables
    \usepackage{amsfonts}       % blackboard math symbols
    \usepackage{nicefrac}       % compact symbols for 1/2, etc.
    \usepackage{microtype}      % microtypography
    \usepackage{xcolor}         % colors

    \usepackage{amsmath}
    \usepackage{wrapfig}

    \usepackage{fontawesome5}   % \faGithub icon
\usepackage[most]{tcolorbox}

\newtcolorbox{promptbox}{
  enhanced,
  breakable,
  colback=gray!10,
  colframe=gray!50,
  boxrule=0.5pt,
  arc=2mm,
  left=2mm,
  right=2mm,
  top=1mm,
  bottom=1mm,
  fontupper=\normalfont\ttfamily,
  before upper={\parindent0pt}
}
    \usepackage{graphicx}
    \usepackage{xcolor}
    \usepackage{pifont}
    \definecolor{rowgray}{gray}{0.95}
    \definecolor{rowwhite}{gray}{1.0}
    \usepackage[table]{xcolor}
    \usepackage{multirow}
    \usepackage[inline]{enumitem}
    \usepackage{tabularx}
    \usepackage{float}

    \usepackage{cleveref}

    \definecolor{ExplainColor}{RGB}{0,70,140} % dark blue

    \newcommand{\explain}[1]{\textcolor{ExplainColor}{\textit{#1}}}

    \definecolor{archEDC}{HTML}{DCEEFF}   % ~blue!12-ish
    \definecolor{archEDCp}{HTML}{DCF7E8}  % ~green!12-ish
    \definecolor{archKG}{HTML}{FFE9D6}    % ~orange!12-ish
    \definecolor{archRAKG}{HTML}{EEE2FF}  % ~violet!12-ish
    \definecolor{archOther}{HTML}{EEEEEE} % ~black!7-ish
    \definecolor{archRELIK}{HTML}{FFF2CC} % ~warm yellow!12-ish
    \definecolor{archKGaware}{HTML}{FCE1EC} % ~pink!12-ish (KG-aware baseline)
    \definecolor{rowStripe}{HTML}{F7F7F7} % very light gray
    \newcommand{\ie}{i.e., }
    \newcommand{\eg}{e.g., }

    \newcommand{\op}[1]{\textsc{#1}}

    \newcommand{\opadd}{\op{Add}}
    \newcommand{\opmintadd}{\op{Mint+Add}}
    \newcommand{\opinfer}{\op{Infer}}
    \newcommand{\opdeprecate}{\op{Deprecate}}
    \newcommand{\opexists}{\op{Exists}}
    \newcommand{\datasetname}{EMERGE}

    \newcommand{\prsemanticcoverage}{semantic coverage}
    \newcommand{\prexecutable}{executable}

    \newcommand{\mprexecutable}{Executable}

    \newcommand{\Figref}[1]{Figure~\ref{#1}}  % start of sentence
    \newcommand{\Tabref}[1]{Table~\ref{#1}}
    \newcommand{\Tabrefs}[2]{Tables~\ref{#1}--\ref{#2}}

    \newcommand{\opaddAcronym}{\op{A}}
    \newcommand{\opmintaddAcronym}{\op{M+A}}
    \newcommand{\opinferAcronym}{\op{I}}
    \newcommand{\opdeprecateAcronym}{\op{D}}
    \newcommand{\opexistsAcronym}{\op{E}}

    \newcommand{\statsNuminstances}{233K}
    \newcommand{\statsTotnrtkgus}{1.19M}
    \newcommand{\statsTotnrtkgusNoM}{1.19}
    \newcommand{\statsTestTotnrtkgus}{23,107}

    \newcommand{\statsTestNrDTriples}{1,628}
    \newcommand{\revklim}[1]{#1}

    \newcommand{\cmark}{\textcolor{green!60!black}{\scalebox{0.9}{\ding{51}}}}
    \newcommand{\xmark}{\textcolor{red!80!black}{\scalebox{0.9}{\ding{55}}}}

    \title{\datasetname: A Benchmark for Updating Knowledge Graphs with
    Emerging Textual Knowledge}

\author{\parbox{\linewidth}{\centering\normalfont
  \textbf{Klim Zaporojets}\,$^{1}$ \quad
  \textbf{Daniel Daza}\,$^{2,3,4}$ \quad
  \textbf{Edoardo Barba}\,$^{5,6}$ \quad \\[0.2em]
  \textbf{Ira Assent}\,$^{1}$ \quad
  \textbf{Roberto Navigli}\,$^{5,7}$ \quad
  \textbf{Paul Groth}\,$^{8}$ \\[0.5em]
   $^{1}$Aarhus University \quad $^{2}$Vrije Universiteit Amsterdam \quad $^{3}$Eindhoven University of Technology \\
  $^{4}$Translational AI Laboratory, Department of Laboratory Medicine, \\
  Amsterdam University Medical Center, Vrije Universiteit Amsterdam \\
  $^{5}$Sapienza University of Rome \quad $^{6}$Principled Intelligence \\
  $^{7}$Babelscape \quad $^{8}$University of Amsterdam \\[0.5em]
  \texttt{klimzaporojets@gmail.com} \\[0.8em]
  \faGithub\;\textbf{Code}: \href{https://github.com/klimzaporojets/emerge}{\texttt{github.com/klimzaporojets/emerge}} \\[0.3em]
  \raisebox{-0.15em}{\includegraphics[height=1em]{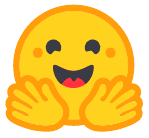}}\;\textbf{Dataset}: \href{https://huggingface.co/datasets/klimzaporojets/emerge-benchmark}{\texttt{klimzaporojets/emerge-benchmark}}
}}
    
\begin{document}

    \maketitle

    \begin{abstract}
      Knowledge Graphs (KGs) are
    structured knowledge repositories
    containing entities and
    relations between them.~In this paper, we
    study
    the
    problem
    of automatically updating KGs over time in response to evolving knowledge in unstructured textual sources.~Addressing this problem requires identifying a wide range of update operations
    based on the state of an existing KG at a
    given time
    and the information extracted from text.~This contrasts with traditional information extraction pipelines, which extract knowledge from text independently of the current state of a KG.
    To address this challenge, we propose a generic and extensible pipeline that pairs textual passages with the KG edit operations they induce on a given KG snapshot. We instantiate this pipeline on the Wikidata knowledge graph and the English Wikipedia corpus to create \datasetname{}, a dataset of \statsNuminstances~Wikipedia passages associated with a total of \statsTotnrtkgusNoM~million KG edits across seven yearly Wikidata snapshots from 2019 to 2025.~Our experimental results highlight key challenges in updating KG snapshots based on emerging textual knowledge, particularly in integrating knowledge expressed in text with the existing KG structure.~\revklim{We further show that KG access substantially improves executable updates and enables triple deprecation.}~These findings position \datasetname{} as a benchmark for developing methods that keep KGs up to date with emerging textual knowledge.
    %These findings position the dataset as a valuable benchmark for future research.
    % The code and dataset are available at \url{https://github.com/klimzaporojets/emerge.
    \end{abstract}

    \section{Introduction}
    % following figure used: https://docs.google.com/drawings/d/1AarHGkv68_ORh6GyC4ZQ70oraQcvqCedq6aRORgT6-M/edit
    \begin{figure}
        \begin{center}
        \centerline{\includegraphics[trim=5 410 25 25, clip,  width=1.0\linewidth]{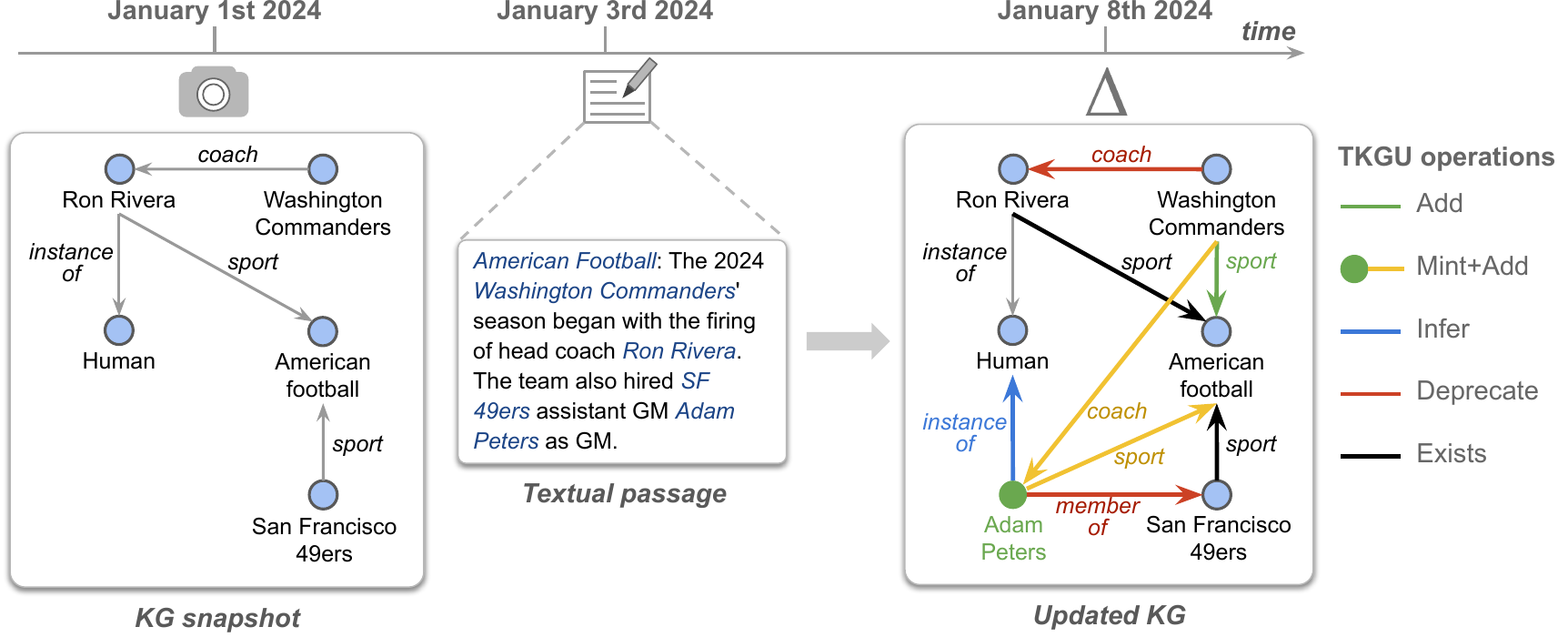}}
        % \centerline{\includegraphics[trim=5 410 40 25, clip,  width=1.0\linewidth]{figures/2025.09.21_MSCA_KG_latent_evolution.pdf}}
        \end{center}
        \caption{One \datasetname~instance.~A reference \textit{KG snapshot} (Jan~1,~2024) is updated with new knowledge from an incoming \textit{textual passage} (Jan~3,~2024). We introduce text-driven KG updating (TKGU) operations to: detect existing relations (\opexists), add relations between known entities (\opadd), create emerging entities and connect them (\opmintadd), infer links between emerging entities and the rest of the graph (\opinfer), and deprecate outdated relations (\opdeprecate).}
        \label{fig:intro-figure}
    \end{figure}
    % Knowledge graphs (KGs) 
    Knowledge graphs (KGs)~\citep{hogan2021knowledge} are graph-structured knowledge bases that represent information about entities and the relations between them using triples of the form (subject, predicate, object). Their ability to compactly encode structured knowledge and support complex queries has led to widespread adoption in applications such as question answering~\citep{wang2024knowledge, dong2025effiqa},
    recommender systems~\citep{zhang2024review, wang2025knowledge}, information retrieval~\citep{reinanda2020knowledge}, fact-checking \citep{kim2023factkg, hao2025fact}, and healthcare~\citep{jiang2025reasoning}, among others \citep{zou2020survey}.

    Constructing a KG from unstructured sources is a challenging task~\citep{tamavsauskaite2023defining}.~It typically requires processing large volumes of domain-specific text, such as scientific publications or news articles, to identify relevant entities and the relations implied between them. To reduce the cost and effort of this process, a large body of work in Information Extraction (IE) has focused on automating KG construction from text~\citep{lingfeng2023kgcsurvey,xu2024large}. Prominent examples of IE tasks include Named Entity Recognition (NER), which identifies entity mentions in text; Relation Extraction (RE), which extracts relations between entities mentioned in a passage; and Entity Linking (EL) and Entity Disambiguation (ED), which map textual mentions to canonical entities in an existing knowledge base. Collectively, the development of methods to tackle these tasks has substantially lowered the barrier to building knowledge graphs from textual data~\citep{lingfeng2023kgcsurvey, pan2024unifying}.

    While KG construction has received significant attention, the problem of \emph{maintaining} KGs over time remains comparatively underexplored.
    This limitation is critical, as many real-world domains are inherently dynamic: facts change, entities acquire new attributes, and previously valid relations become outdated~\cite{polleres2023does}.
    As a concrete example, consider a KG describing political offices worldwide. Queries over such a graph may become outdated whenever a politician changes office, unless the KG is updated accordingly.
    As a result, the utility of a KG in real-world deployments depends not only on its initial construction, but also on effective maintenance and the availability of methods that can update it as underlying domain knowledge evolves.

    Despite their success in extracting structured facts from text, existing IE methods are not designed to support KG maintenance. In particular, IE systems operate largely independently of the current state of a knowledge graph.
    Given a textual passage, such systems can produce a semantically valid triple, yet they do not address key questions required for updating a KG:
    Can the extracted triple be directly added to the graph? Does it invalidate or supersede existing triples? Does it require introducing a previously unseen entity? Addressing such questions requires reasoning jointly over textual evidence and the content of an existing KG, which lies outside the scope of standard IE formulations.

    To address this gap, we introduce \textbf{\datasetname}, a dataset for evaluating methods for updating knowledge graphs from textual evidence, along with a general pipeline for constructing such datasets. Each instance in \datasetname~consists of a textual passage, a KG snapshot, and update operations induced by the passage on that snapshot (see \Figref{fig:intro-figure}).
    % Each instance in \datasetname~consists of a textual passage, a KG snapshot, and update operations induced by the passage (see \Figref{fig:intro-figure}). 
    We refer to the pairing between textual evidence and the KG updates it supports as an \emph{alignment}. These operations include adding new triples, introducing new entities, deprecating outdated triples, and identifying triples already present in the graph.~To construct \datasetname, we align KG updates with textual evidence using entity mentions linked to KG entities, and \revklim{use LLM-based curation to flag updates not supported by their corresponding passages}.~By modeling updates with respect to a KG snapshot, \datasetname~goes beyond traditional IE datasets and captures the requirements of KG maintenance over time. To the best of our knowledge, \datasetname~is the first dataset to systematically study this problem.
    % apply LLM-based verification to retain only text-supported updates.~

    We use EMERGE to benchmark several state-of-the-art methods from the IE literature. Our results show that existing approaches fall short of supporting the range of operations required for KG updates. In particular, these methods rely exclusively on textual signals and remain unaware of how knowledge is structured within an existing KG. As a result, the extracted triples, while often semantically correct, fail to align with the graph structure. In summary, our contributions are as follows:
    \begin{enumerate}
        \item We formalize the task of KG maintenance from text as a set of update operations applied to a given KG snapshot, going beyond traditional information extraction formulations.
        \item We introduce EMERGE, a new dataset that pairs Wikipedia textual passages with
        Wikidata KG update operations, enabling systematic evaluation of KG update methods over time.
        \item We propose a principled pipeline for constructing such datasets by aligning KG updates with textual evidence, 
        \revklim{followed by LLM-guided curation and manual verification.}
        % followed by LLM-guided filtering and manual curation.
        % \item We benchmark state-of-the-art IE methods on EMERGE and introduce a KG-aware baseline, demonstrating both the limitations of existing approaches and the gains from incorporating KG context.
        \item \revklim{We benchmark existing state-of-the-art IE methods and introduce a \textit{KG-aware} baseline, showing both their limitations and the gains from incorporating KG context.}
    \end{enumerate}

    % ------------------------------------------------------------------------------------------------------------------
    % Table 1 (dataset comparison)
    % Script: emerge-prod:src/stats/dataset/table_dataset_comparison.ipynb  (writes this LaTeX table)
    %         other datasets: numbers hard-coded there, computed once by emerge-prod:src/stats/dataset/other_datasets/<dataset>_stats.py
    % EMERGE row (camera-ready, v1.0.1 corpus): Inst. 233.11K and Triples = stats macros; Rels. 833 / Ents. 174.51K = distinct
    %         relations / entities over all corpus tkgu_triples (jq; the notebook still reads the old pre-re-issue pickle)
    % ------------------------------------------------------------------------------------------------------------------
    \begin{table}
    \caption{
    Overview of major Information Extraction datasets that ground extracted triples in a KG, compared with \datasetname.
    }
    \label{tab:comparison-benchmarks-main}
    \rowcolors{6}{rowwhite}{rowgray} % start from row 2
    \resizebox{\textwidth}{!}{\begin{tabular}{l|cc|ccccc| cccc}
    \toprule
    & \multicolumn{2}{c|}{Evolution}
    & \multicolumn{5}{c|}{TKGU operations}
    & \multicolumn{4}{c}{Annotation statistics} \\
    \cmidrule(lr){2-3}\cmidrule(lr){4-8}\cmidrule(lr){9-12}
    Dataset & KG & Text & E & A & M+A & I & D &  Instances & Rel. types & Entities & Triples \\
    % Dataset & KG & Text & E & A & M+A & I & D &  Inst. & Rels. & Ents. & Triples \\
    \midrule
    WebNLG \citep{gardent2017webnlg} & \xmark & \xmark & \cmark & \cmark & \xmark & \xmark & \xmark & ~~38.87K & ~~~411 & ~~~~3.62K & 115.28K \\
    % FewRel (\citeyear{han2018fewrel}) & \xmark & \xmark & \cmark & \cmark & \xmark & \xmark & \xmark & \xmark &  70.85K & 124 & 84.60K & 68.35K \\
    T-REX \citep{elsahar2019t} & \xmark & \xmark & \cmark & \cmark & \xmark & \xmark & \xmark & ~~~~6.20M & ~~~642 & ~~~~4.65M & ~~11.00M \\
    Wiki-NRE \citep{distiawan2019neural} & \xmark & \xmark & \cmark & \cmark & \xmark & \xmark & \xmark & 255.65K & ~~~158 & 279.89K & 330.01K \\
    DocRED \citep{yao2019docred} & \xmark & \xmark & \cmark & \cmark & \xmark & \xmark & \xmark & 101.87K & ~~~~~96 & 782.46K & ~~~~1.51M \\
    BioRel \citep{xing2020biorel} & \xmark & \xmark & \cmark & \cmark & \xmark & \xmark & \xmark & 533.56K & ~~~125 & ~~17.82K & 578.01K \\
    Wiki20 \citep{han2020more} & \xmark & \xmark & \cmark & \cmark & \cmark & \xmark & \xmark & 901.31K & ~~~~~81 & 392.26K & 901.31K  \\
    REBEL \citep{cabot2021rebel} & \xmark & \xmark & \cmark & \cmark & \xmark & \xmark & \xmark & ~~~~3.06M & 1,155 & ~~~~5.64M & ~~10.31M \\
    CDG \citep{zhang2022distant} & \xmark & \xmark & \cmark & \cmark & \xmark & \xmark & \xmark & ~~72.33K & ~~~~~14 & ~~15.72K & 175.42K \\
    SynthIE-text \citep{josifoski2023exploiting}  & \xmark & \xmark & \cmark & \cmark & \cmark & \xmark & \xmark & ~~70.27K & ~~~888 & 121.26K & 241.80K \\
    \textsc{Text2KG} \citep{mihindukulasooriya2023text2kgbench} & \xmark & \xmark & \cmark & \cmark & \xmark & \xmark & \xmark & ~~18.42K & ~~~291 & ~~~~1.91K & ~~~~8.62K \\
    \midrule
    \datasetname~(ours) & \cmark & \cmark & \cmark & \cmark & \cmark & \cmark & \cmark & 233.11K & ~~~833 & 174.51K & ~~~~\statsTotnrtkgus \\
    \bottomrule
    \end{tabular}}
    \end{table}

    \section{Related Work}
    \label{sec:related-work}
    We review work most relevant to updating KGs from textual evidence; Appendix \ref{sec:appendix:comparison} provides a detailed comparison of \datasetname{} with existing IE datasets.

    \textbf{Information extraction from text.} A large body of work in Information Extraction (IE) studies how to extract structured knowledge from unstructured text.~Early datasets such as MUC-7~\citep{chinchor1998muc}, CoNLL04~\citep{roth2004linear}, and ACE 2005~\citep{walker2006ace} focused on identifying entities and relations within individual sentences. Subsequent datasets expanded the scope in terms of size, relation diversity, document length, and domain, including ERE~\citep{aguilar2014comparison,song2015light}, BC5CDR~\citep{li2016biocreative}, TACRED~\citep{zhang2017position}, SciERC~\citep{luan2018multi}, SemEval-2010 and SemEval-2017~\citep{hendrickx2010semeval,augenstein2017semeval}, DWIE~\citep{zaporojets2021dwie}, and BioRED~\citep{luo2022biored}. Several datasets explicitly align textual evidence with KG triples.~Distantly supervised resources such as NYT~\citep{riedel2010modeling} link text to Freebase triples, while FACC1~\citep{gabrilovich2013facc1} aligns ClueWeb documents with Freebase entity mentions at scale.~More recent datasets, including WebNLG~\citep{gardent2017webnlg},
    % FewRel~\citep{han2018fewrel},
    DocRED~\citep{yao2019docred}, Wiki/GEO-NRE~\citep{distiawan2019neural},
    % BioRel~\citep{xing2020biorel},
    T-REX~\citep{elsahar2019t},
    KELM~\citep{agarwal2021knowledge},
    REBEL~\citep{cabot2021rebel},
    Text2KG~\citep{mihindukulasooriya2023text2kgbench}, SciNLP~\citep{duan2025scinlp}, and MINE~\citep{mo2025kggen}
    further strengthen the connection between text and KG triples.

    These datasets have been instrumental in advancing models that map text to entities, relations, and triples. However, they frame extraction independently of the current state of a KG. 
    The output of an IE system is typically a set of candidate triples, without specifying how they relate to an existing KG: whether to add them, supersede existing triples, or introduce new entities.
    %The output of an IE system is typically a set of candidate triples, without specifying how these triples relate to an existing KG: whether they should be added, whether they supersede existing facts, or whether they require introducing new entities. 
    In contrast, our work considers a more holistic problem in which textual evidence is interpreted relative to a specific KG snapshot in order to determine the concrete update operations required to maintain the graph over time.~\Tabref{tab:comparison-benchmarks-main} compares \datasetname~with major IE datasets along three axes: (1) \textit{Evolution}: whether the dataset reflects naturally occurring changes over time in both the \textit{KG} and \textit{text}; (2) coverage of text-driven KG update \textit{(TKGU) operations} necessary to update a KG from textual sources, and introduced in Section \ref{sec:problem-definition};
    and (3) \textit{Annotation statistics}, including number of \textit{instances}, \textit{relation types}, \textit{entities} and annotated \textit{triples}.
    % including number of instances (\textit{Inst.}), relation types (\textit{Rels.}), entities (\textit{Ents.}) and  annotated \textit{triples}.

    \textbf{Temporal Knowledge Graph completion.} Our task is related to but distinct from temporal KG completion (TKGC)~\citep{boschee2015icews, leetaru2013gdelt, gastinger2024tgb, zhang2024dtgb, yi2025tgb}, which models the temporal evolution of facts by associating triples with timestamps and predicting missing or future edges from historical patterns within the graph.
    In contrast, we introduce text-driven KG updating (TKGU): given a KG snapshot and a new passage, the goal is to infer the updates required for the KG to reflect the information expressed in that passage.
    This shift introduces several key differences: (i) updates are explicitly grounded in a passage, (ii) previously unseen entities may need to be added to the KG, and (iii) outputs are update operations relative to a specific snapshot rather than timestamped triples.~We view TKGU and TKGC as complementary: TKGC captures how the KG evolves from internal graph patterns, whereas TKGU captures how external textual evidence triggers concrete updates to an existing KG.

    \textbf{Verification of textual claims and triples using knowledge graphs.} A related but distinct line of work studies the use of knowledge graphs to verify information expressed in text. Claim verification datasets such as FactKG~\citep{kim2023factkg} and subsequent extensions~\citep{kim2023kggpt,yuan2024zero,lourencco2026kg,upadhyay2026evidence,zhou2026assessing, sawczyn2026factselfcheck} pair textual claims with KG evidence and frame the task as classifying a claim as supported, refuted, or unverifiable. These approaches explicitly contrast text against an existing KG, but their goal is to assess the validity of individual claims rather than to update the graph itself.

    \section{Problem Definition}
    \label{sec:problem-definition}

    We define the problem of text-driven knowledge graph updating (TKGU) as determining the set of graph update operations required to maintain a knowledge graph (KG) given new textual evidence.

    Formally, a KG snapshot at time $t$ is defined as a tuple $G_t = (V_t, R_t, T_t)$, where $V_t$ is a set of entities, $R_t$ is a set of relation types, and $T_t \subseteq V_t \times R_t \times V_t$ is a set of triples of the form $(s, p, o)$, with $s,o \in V_t$ and $p \in R_t$. Given a textual passage $d_{t'}$ created at time $t' > t$, with $M(d_{t'})$ denoting the set of entities explicitly mentioned in $d_{t'}$, the goal of TKGU is to infer the set of update operations that should be applied to $G_t$ to reflect the knowledge expressed in $d_{t'}$. In this paper, we introduce the fundamental TKGU operations, formalized below and illustrated with concrete examples.

    \textbf{\opadd~(\opaddAcronym):} Addition of a new triple involving entities that already exist in the KG. This operation applies when $(s,p,o) \notin T_t$, $\{s, o\} \subseteq M(d_{t'})$, $(s,p,o)$ is entailed by $d_{t'}$, and $s \in V_t \wedge o \in V_t$.

    For example, in \Figref{fig:intro-figure}, the triple
    \textit{(Washington Commanders, sport, American Football)} is added, even though both entities already exist in the KG snapshot.

    \textbf{\opmintadd~(\opmintaddAcronym):} Creation of one or more new (emerging) entities, followed by the addition of a triple involving them. This operation applies when $(s,p,o) \notin T_t$, $\{s, o\} \subseteq M(d_{t'})$, $s \notin V_t \vee o \notin V_t$, and $(s,p,o)$ is entailed by $d_{t'}$.

    For example, in \Figref{fig:intro-figure}, the triple
    \textit{(Washington Commanders, coach, Adam Peters)} introduces the previously unseen entity \textit{Adam Peters}, which must first be minted (\ie added to the KG) before the triple can be added; both \textit{Washington Commanders} and \textit{Adam Peters} are mentioned in the passage.

    \textbf{\opinfer~(\opinferAcronym):} Addition of a triple linking a newly introduced (emerging) entity mentioned in the passage to an existing KG entity that is not mentioned in the passage.~Formally, this operation applies when $(s,p,o) \notin T_t$, exactly one of $\{s, o\}$ belongs to $M(d_{t'}) \setminus V_t$, and the other belongs to $V_t \setminus M(d_{t'})$. The relation $p$ must therefore be inferred jointly from passage $d_{t'}$ and KG context, testing the ability of a model to integrate emerging entities into the broader KG.

    For example, in \Figref{fig:intro-figure}, the triple \textit{(Adam Peters, instance of, Human)} links the newly minted entity \textit{Adam Peters} (explicitly mentioned in the passage) to the pre-existing entity \textit{Human} (not mentioned in the passage), and must be inferred from the KG structure.

    % \textbf{\opdeprecate~(\opdeprecateAcronym):} Invalidation of an existing triple based on updated information expressed in the passage $d_{t'}$. Formally, this operation applies when $(s,p,o) \in T_t$, $\{s, o\} \subseteq M(d_{t'})$, and the evidence in $d_{t'}$ indicates that $(s,p,o)$ is no longer valid.

    \textbf{\opdeprecate~(\opdeprecateAcronym):} \revklim{Invalidation of a triple based on updated information expressed in the passage $d_{t'}$. Formally, this operation applies when at least one of $\{s, o\}$ belongs to $M(d_{t'})$, at least one belongs to $V_t$, and the evidence in $d_{t'}$ indicates that $(s,p,o)$ is no longer valid. In most cases (91.4\% of the dataset), $(s,p,o) \in T_t$ and both entities are mentioned in $d_{t'}$. In the remaining cases, $(s,p,o) \notin T_t$, as when the triple is added after $t$, potentially involving a newly minted entity (e.g., \textit{Adam Peters} in \Figref{fig:intro-figure}).}

    For example, in \Figref{fig:intro-figure}, the triples
    \textit{(Adam Peters, member of, San Francisco 49ers)} and
    \textit{(Washington Commanders, coach, Ron Rivera)}
    are deprecated based on newly emerging information expressed in the textual passage.

    \textbf{\opexists~(\opexistsAcronym):} Identification of a triple already present in the KG that is supported by the textual passage, i.e., $(s,p,o) \in T_t$, $\{s, o\} \subseteq M(d_{t'})$, and $(s,p,o)$ is entailed by $d_{t'}$. This operation evaluates the ability of a model to recognize and confirm existing knowledge rather than redundantly re-adding it. For example, in \Figref{fig:intro-figure}, the triple
    \textit{(San Francisco 49ers, sport, American Football)}
    is both supported by the passage and already present in the original KG snapshot.

\begin{wraptable}[20]{r}{0.50\textwidth}
\vspace{-2em}
\caption{Comparison of existing IE models by supported TKGU operations:
(1) \opexists{} (\textit{\opexistsAcronym}), (2) \opadd{} (\textit{\opaddAcronym}), (3) \opmintadd{} (\textit{\opmintaddAcronym}), (4) \opinfer{} (\textit{\opinferAcronym}), and (5) \opdeprecate{} (\textit{\opdeprecateAcronym}).
\textit{KG Linked} indicates whether extracted triples are mapped to canonical KG identifiers.
}
\label{tab:comparison_current_architectures}
\vspace{\baselineskip}
% \vspace{-0.5em}
\setlength{\tabcolsep}{3pt}
\rowcolors{1}{gray!15}{white}
\begin{tabularx}{\linewidth}{
  p{0.31\linewidth}
  >{\centering\arraybackslash}p{0.13\linewidth}
  *{2}{>{\centering\arraybackslash}m{0.06\linewidth}}*{1}{>{\centering\arraybackslash}m{0.11\linewidth}}*{2}{>{\centering\arraybackslash}m{0.06\linewidth}}}
\toprule
\rowcolor{white}
\multirow{2}{*}{\raisebox{-0.4ex}{Model}}
& \raisebox{-0.55ex}{KG}
& \multicolumn{5}{c@{}}{TKGU operations} \\
\cmidrule(lr){3-7}
\rowcolor{white}
& \raisebox{0.55ex}{Linked}
& \textit{\opexistsAcronym}
& \textit{\opaddAcronym}
& \textit{\opmintaddAcronym}
& \textit{\opinferAcronym}
& \textit{\opdeprecateAcronym} \\
\midrule
\rowcolors{1}{gray!7}{white}
REBEL \citep{cabot2021rebel} & \xmark & -- & -- & -- & -- & -- \\
GenIE \citep{josifoski2021genie} & \cmark & \cmark & \cmark & \xmark & \xmark & \xmark \\
KnowGL \citep{rossiello2023knowgl} & \cmark & \cmark & \cmark & \xmark & \xmark & \xmark \\
GCD \citep{geng2023grammar} & \cmark & \cmark & \cmark & \xmark & \xmark & \xmark \\
ReLiK cIE \citep{orlando2024relik} & \cmark & \cmark & \cmark & \xmark & \xmark & \xmark \\
ReLiK RE \citep{orlando2024relik} & \xmark & -- & -- & -- & -- & -- \\
EDC \citep{zhang2024extract} & \xmark & -- & -- & -- & -- & -- \\
ATG \citep{zaratiana2024autoregressive} & \xmark & -- & -- & -- & -- & -- \\
CodeKGC \citep{bi2024codekgc} & \xmark & -- & -- & -- & -- & -- \\
KARMA \citep{lu2025karma} & \xmark & -- & -- & -- & -- & -- \\
RAKG \citep{zhang2025rakg} & \xmark & -- & -- & -- & -- & -- \\
KGGen \citep{mo2025kggen} & \xmark & -- & -- & -- & -- & -- \\
\bottomrule
\end{tabularx}
\vspace{-5.95em}
\end{wraptable}
\Tabref{tab:comparison_current_architectures} compares
    % existing
    widely used and representative IE architectures according to the TKGU operations they support.~While some methods can identify \opexists~and \opadd~operations, most struggle with \opmintadd, and none explicitly support \opinfer, which requires integrating newly introduced entities into the existing KG structure.~Furthermore, existing IE methods are not designed to identify \opdeprecate~operations based on emerging textual evidence.~Finally, most existing IE models do not canonicalize extracted triples with respect to a target KG (\textsc{KG Linked} column in \Tabref{tab:comparison_current_architectures}).
    Consequently, while these models can extract triples that cover the knowledge involved in TKGU operations, they are not designed to act on an existing KG.

    \section{EMERGE}
    \label{sec:our-dataset}
    \begin{figure}[t]
    \centerline{\includegraphics[trim=0 325 105 10, clip,  width=1.0\linewidth] {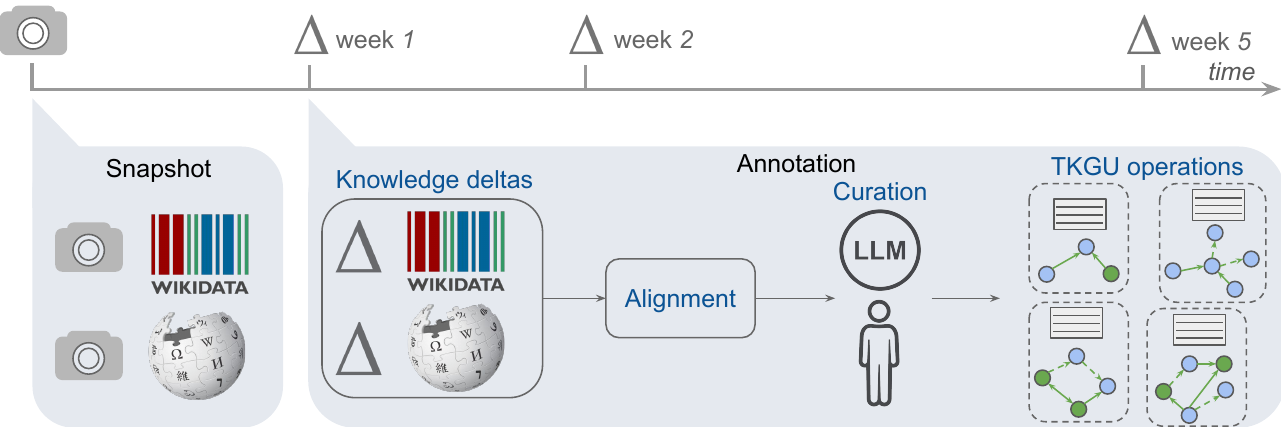}}
    \caption{
    \datasetname{} pipeline: weekly knowledge deltas ($\Delta$) are extracted from Wikipedia (text) and Wikidata (KG) relative to a fixed snapshot, then paired in the \textit{Alignment} step.
    During \textit{Curation}, an LLM flags KG updates not supported by text, verified with manual annotations on a subsample. This results in high-quality text-KG update pairs (\textit{TKGU operations}) as in Figure~\ref{fig:intro-figure}.
    }
    \label{fig:pipeline}
    \end{figure}

    % We describe the construction of \datasetname, a large-scale dataset supporting all text-driven KG updating (TKGU) operations defined in Section~\ref{sec:problem-definition}.
    % We describe the construction of \datasetname, a large-scale dataset that supports all operations defined in Section~\ref{sec:problem-definition}.
    We describe the construction of \datasetname, a large-scale dataset that supports all operations defined in Section~\ref{sec:problem-definition}, with the pipeline illustrated in \Figref{fig:pipeline}.

    \subsection{Data Collection}
    \label{sec:data-collection}
    % Below, we describe the main steps of the \datasetname\ creation pipeline (see also~\Figref{fig:pipeline}).
    % also illustrated in \Figref{fig:pipeline}.

    \noindent \textbf{Wikipedia and Wikidata dumps.} We begin by downloading the historical revision logs from the Wikipedia and Wikidata dumps available at \url{https://dumps.wikimedia.org/}. These logs provide complete access to the revision history of Wikipedia and Wikidata, enabling fine-grained tracking of temporal changes. This allows us to construct \datasetname~using \textit{any number of arbitrarily defined KG snapshots and delta windows, with temporal precision down to the second}.
     % These logs provide complete access to the revision history of Wikipedia and Wikidata, enabling fine-grained tracking of temporal changes that allows us to construct \datasetname~using \textit{any number of arbitrarily defined KG snapshots and delta windows, with temporal precision down to the second}.

    \noindent \textbf{Snapshot generation.}
    We construct a dataset from seven snapshots \revklim{(January~1, 00{:}00~CET)} spanning 2019--2025. For each snapshot timestamp $t$, we process Wikipedia and Wikidata revision histories to obtain:
    \begin{enumerate*}[label=(\arabic*)]
        \item the Wikidata KG snapshot $G_t$,
        \item a dictionary of entities present in Wikipedia at $t$, and
        \item a dictionary of relation types present in Wikidata at $t$.
    \end{enumerate*}
    Following Wikidata5M~\citep{wang2021kepler}, we restrict $G_t$ to entities that appear in Wikipedia.

    \noindent \textbf{KG deltas generation.}~To evaluate model performance when updating increasingly outdated KGs with emerging knowledge, we generate cumulative \textit{deltas} over weekly windows of up to five weeks for each snapshot (see \Figref{fig:pipeline}). Each delta is the difference between two KG snapshots, $G_{t+\Delta} - G_t$, where $\Delta$ denotes the window length.

    We design the TKGU operations in Section~\ref{sec:problem-definition} to capture all the KG changes observed in these deltas.
    Concretely, \textit{\opexists} covers triples present in both $G_t$ and $G_{t+\Delta}$, \textit{\opadd} captures emerging relations in $G_{t+\Delta}$ between entities already in $G_t$, and \textit{\opmintadd} and \textit{\opinfer} span emerging relations where the subject or object does not exist in $G_t$ but is introduced in $G_{t+\Delta}$.
    Finally, \opdeprecate~includes (i) edges removed between snapshots (present in $G_t$ but not in $G_{t+\Delta}$), and (ii) edges explicitly deprecated within the delta window, as indicated by Wikidata triple qualifiers (Appendix~\ref{sec:appendix:qualifiers}).
    We mark these triples as deprecated rather than removing them, since the underlying fact does not change but expires within the delta interval.

    \noindent \textbf{Aligning KG deltas with text.} For each delta in a given snapshot $t$, we retrieve the newly introduced Wikipedia passages within the temporal window corresponding to that delta. Following the approach of \cite{cabot2021rebel, elsahar2019t}, we then \textit{align} these passages with triples in each of the KG deltas by matching the annotated hyperlinked entity mentions in each of the passages to the corresponding entities in the triples.
    We refer to this distant supervision process as the \textit{alignment} step (see \Figref{fig:pipeline}).~\revklim{Restricting alignment to the delta's temporal window reduces spurious matches between KG changes and passages introduced at unrelated times. However, facts appearing in Wikipedia and Wikidata in different windows may be missed, reducing coverage (see the discussion of this and other limitations in Appendix~\ref{sec:appendix:limitations}). The resulting text-triple pairs are subsequently annotated during \textit{curation} (Section~\ref{sec:quality-control}) to flag TKGU operations not supported by their textual passages.}
    % ~\revklim{Restricting alignment to the same window avoids unrestricted matching across time. As a consequence, a fact that reaches Wikipedia and Wikidata in different windows yields no aligned instance, thus potentially reducing coverage without introducing incorrect alignments (Appendix~\ref{sec:appendix:limitations}). The resulting text-triple pairs are subsequently annotated during \textit{curation} (Section~\ref{sec:quality-control}) to flag TKGU operations not supported by their textual passages.} 
    % The resulting text-triple pairs are subsequently refined in the \textit{curation} step (see Section~\ref{sec:quality-control}) to retain only those pairs in which the textual content supports the associated KG changes.
    \subsection{Quality Assurance and Control}
    \label{sec:quality-control}
    During the \textit{alignment} step of the \datasetname~creation pipeline (see \Figref{fig:pipeline}) we use multiple heuristics to ensure the quality of the aligned textual passages with KG updates. For instance, we filter out passages with a low proportion of English words and those containing wikitext special symbols used for constructing elements such as tables and images. Furthermore, we discard updates in Wikidata and Wikipedia that are quickly rolled back, as these often indicate incorrect or vandalized changes.~A complete list of preprocessing and cleaning steps can be found in Appendix~\ref{sec:appendix:cleaning}.

    During the \textit{curation} step of the \datasetname~pipeline (see \Figref{fig:pipeline}), we use the \texttt{Meta-Llama-3.1-405B} LLM to validate that all TKGU operations can be derived from the corresponding textual passage.~The full prompt design and illustrative examples are provided in Appendix~\ref{sec:appendix:prompts}.~This step flags KG updates not supported by the text, rather than removing them, enabling future use of more powerful LLMs for additional verification and curation.
    \revklim{For all reported metrics (Section~\ref{sec:metrics-and-evaluation}), the ground truth includes only TKGU operations marked as supported by their corresponding textual passages.}
    Appendix \ref{sec:appendix:annotation-stats} reports additional statistics \revklim{on the fraction of TKGU operations marked as supported.}

    To evaluate curation reliability, we manually annotate 500 triple-text pairs (100 per each TKGU operation) and compare human judgments with the LLM. Agreement is strong to almost perfect, with pairwise Cohen's $\kappa$ of 0.61--0.88 (overall 0.74--0.79), and multi-rater Fleiss' $\kappa$ and Krippendorff's $\alpha$ of 0.67--0.83 (overall 0.77).  This indicates that LLM-based curation is sufficiently consistent with human judgment to act as a dataset-level safeguard.~\revklim{These judgments concern textual support, whereas TKGU operation types are assigned deterministically from KG snapshots and deltas, entity membership in $V_t$, and hyperlink-based mentions $M(d_{t'})$ (Section~\ref{sec:problem-definition}).} Detailed annotation guidelines and per-operation agreement statistics are provided in Appendix~\ref{sec:appendix:human-annotation}.

    % ------------------------------------------------------------------------------------------------------------------
    % KG sizes in this sentence (same numbers as Table 8, columns Entities / Rel. Types / Triples)
    % Script:  emerge-prod:src/stats/dataset/table_dataset_statistics.ipynb  (per snapshot: entities, relation types, triples)
    % Triples: rows of the v1.0.1 KG snapshots (2019 26,346,379; 2025 38,590,522), recounted for the camera-ready by
    %          emerge-stage:scripts/stats/camera_ready_reissue_20261028/analysis_2021_dip_reissue_20261028.py (analysis2021__kg_relation_counts.json)
    % Entities: lines of the yearly Wikipedia entity dictionaries. Rel. types: lines of the yearly relation dictionary
    %          (all Wikidata properties at t = R_t), data/indices/relik_edc_relation_indexes/<year>/documents.jsonl -- NOT the
    %          ~1K relations used in the snapshot triples. Entities and relation types are unchanged in the re-issue.
    % ------------------------------------------------------------------------------------------------------------------
    \subsection{Dataset Statistics}
    \label{sec:dataset-statistics}
    \datasetname~consists of \statsNuminstances~instances across seven yearly KG snapshots (2019--2025), with a total of \statsTotnrtkgus~TKGU update operations (see \Tabref{tab:comparison-benchmarks-main} for comparison with other IE datasets).
    Both the KG size (\ie number of entities and edges) and the schema (\ie the relation types) evolve across snapshots.~For instance, the 2019 KG snapshot contains 5.96M entities, \revklim{26.35}M relations, and 5,646 relation types, while the 2025 snapshot includes 6.93M entities, \revklim{38.59}M relations, and 12,304 relation types. This dynamic setting enables the evaluation of model robustness under evolving KG knowledge and schema changes, thereby \textit{reflecting real-world KG evolution}. Additional tables and figures in Appendix~\ref{sec:appendix:additional-dataset-statistics} provide a detailed overview of the size and distribution of TKGU operations in \datasetname. Furthermore, \Tabrefs{tab:x-triples-examples}{tab:d-triples-examples} in Appendix~\ref{sec:appendix:qualitative-analysis} present illustrative examples of each TKGU operation type introduced in Section~\ref{sec:problem-definition}.

    \subsection{Extensibility}
    \label{sec:extensibility}
    EMERGE is extensible by design.~Its temporal KG delta extraction, text-KG alignment, TKGU operation annotation (Section~\ref{sec:data-collection}), and LLM-based curation (Section~\ref{sec:quality-control}) apply beyond Wikidata and Wikipedia wherever a KG evolves over time and text describes those changes. Concretely, we envision two main extension scenarios. First, Wikidata can be retained while Wikipedia is replaced with domain-specific corpora: through links to over 100 million entities and specialized resources such as UMLS, DrugBank, PubChem, UniProt, and GeoNames, Wikidata can serve as a bridge to specialized KGs.~Second, the pipeline can be applied directly to other KGs, such as UMLS, DrugBank, or OpenAlex, given versioned snapshots for temporal delta computation.
    In both cases, hyperlink-based alignment to textual sources (\eg scientific articles) would be replaced by entity linking using Wiki-based methods~\citep{wu2020scalable, van2020rel, orlando2024relik} or domain-specific linkers~\citep{mishra2025medpath, christof2025medlinkde, kim2025learning, zhang2024entity, graciotti2025ke, semnani2025lemonade, sarkar2025mahanama}.
    Although noisier, unsupported alignments can be
    % filtered
    \revklim{flagged}
    by LLM-based curation as showcased in Section~\ref{sec:quality-control}.~Potential real-world applications include news-driven updates to event-centric KGs, \revklim{publication-driven updates to biomedical KGs, and legislation-driven updates to regulatory KGs.}
    % scientific publications updating biomedical KGs, and legislation updating regulatory KGs.
    
    \revklim{The pipeline could also be extended to other languages. Since Wikipedia editions in other languages link their pages to the same Wikidata items, our hyperlink-based alignment applies to them directly, while other non-English corpora would require entity linkers adapted to the target language. Literal-valued attributes could be supported by normalizing dates and quantities and fuzzy-matching strings, extending alignment beyond entity-to-entity triples. These extensions remain future work.}

    % ------------------------------------------------------------------------------------------------------------------
    % Test-set numbers in this section: 3,500 instances (500 per snapshot, 100 per delta); \statsTestTotnrtkgus = 23,107
    %   405B-supported operations; Deprecate 7% = \statsTestNrDTriples / \statsTestTotnrtkgus = 1,628 / 23,107 = 7.0%
    % Script: emerge-stage:scripts/stats/camera_ready_reissue_20261028/figures_dataset_stats_reissue_20261028.py (counts JSON, see stats-macros block)
    % Data:   v1.0.1 test set. Subsampling: emerge-prod:src/dataset/emerge/s07b_subsample_dataset.py (config: emerge-prod:config/dataset/emerge/s07b_subsample_dataset/)
    % ------------------------------------------------------------------------------------------------------------------
    \section{Evaluating Existing IE Models}
    \label{sec:experimental-setup}
    We benchmark \datasetname{} on a subset of 3,500 instances obtained by subsampling 500 instances per snapshot (100 per delta), comprising \statsTestTotnrtkgus{} TKGU operations. To ensure sufficient coverage of the \opdeprecate{} operation, we retained up to 40 \opdeprecate{} instances per delta, resulting in \opdeprecate{} operations accounting for 7\% (\statsTestNrDTriples{}) of all TKGU operations in the test set. 
% Each TKGU operation type is evaluated independently, ensuring stable metrics despite this imbalance (see \Tabref{tab:dataset_re_results}). 
\revklim{Each TKGU operation type is evaluated independently, so this up-weighting does not affect the per-operation metrics (see \Tabref{tab:dataset_re_results}). As a trade-off, the operation distribution of the test set is less representative of the full dataset.}
A full comparison of TKGU operation distributions is provided in~Section~\ref{sec:appendix:tkgu-operations-distribution}.

    \subsection{Models}
    \label{sec:models}
    %%%%%%%
    % script used to generate: /home/klim/work/w/repositories/emerge-stage/src/stats/evaluation/figure_zs_vs_icl_v2_empty_pred_fix.ipynb
    %%%%%%%
    % \begin{wrapfigure}[17]{r}{0.6\linewidth}
    % \centering
    % % trim=left bottom right top
    % \includegraphics[trim=0 10 0 10,clip,width=\linewidth]{figures/statistics/zs_vs_icl_v2_empty_pred_fix_hlegend.pdf}
    % \caption{
    % \revklimtodo{Completeness and G-BERTScore-R of EDC+ with \texttt{GPT-5.1} under \textit{in-context} learning versus \textit{zero-shot} prompting across TKGU operation types: \opexists~(\opexistsAcronym), \opadd~(\opaddAcronym), \opmintadd~(\opmintaddAcronym), \opinfer~(\opinferAcronym), and \opdeprecate~(\opdeprecateAcronym).
    % }}
    % \label{fig:in-context-zero-shot-performance}
    %  \vspace{0.0em}
    % \end{wrapfigure}
    To assess performance on \datasetname{}, we evaluate three widely used IE paradigms:~autoregressive seq2seq extraction \citep{sutskever2014sequence}, extractive span-based modeling \citep{lee2017end}, and LLM-based generative approaches \citep{dagdelen2024structured,xu2024large,zhang2025survey}.
    We benchmark each paradigm using the following models: REBEL \citep{cabot2021rebel} for seq2seq extraction, ReLiK \citep{orlando2024relik} for span-based modeling, and the LLM-driven methods EDC \citep{zhang2024extract}, KGGen \citep{mo2025kggen}, and RAKG \citep{zhang2025rakg}. We evaluate LLMs of varying scale, including \texttt{Mistral-small}, \texttt{Mistral-Large-2411},~and \texttt{GPT-5.1}; due to its multi-stage, retrieval-augmented design and associated inference cost, RAKG is not evaluated with \texttt{GPT-5.1}. See Appendix~\ref{sec:appendix:compared-models} for details.

    Additionally, we extend EDC \citep{zhang2024extract} to cover the knowledge involved in the \opinfer{} and \opdeprecate{} TKGU operations. Concretely, we modify the original EDC prompt to (i) extract triples involving entities present in a Wikidata snapshot but not explicitly mentioned in the input text, and (ii) identify candidate triples for deprecation based on the input text. We refer to this variant as \textbf{EDC+}. The EDC+ prompts are described in Appendix~\ref{sec:appendix:edc-prompts}.

    \revklim{To evaluate TKGU with explicit access to the current KG state, we introduce a \textbf{KG-aware (KGA)} baseline built on the same \texttt{GPT-5.1} as EDC+. KGA is prompted separately for each TKGU operation present in an instance, producing triples with canonical identifiers for existing entities and relation types, while emerging entities are minted without identifiers. We evaluate three variants. \textbf{KGA-RAG} receives the top $k{=}32$ retrieved KG candidates (entities, relation types, and triples) from $G_t$. \textbf{KGA-Oracle (KGA-O)} receives gold triples for \opexists{} and \opdeprecate{}, and gold existing entities and relation types for \opadd{}, \opmintadd{}, and \opinfer{}. For \opinfer{}, KGA-O additionally receives up to six triples from $G_t$ linking another entity to the same existing entity via the same relation.~These examples illustrate how the relation is used in $G_t$, guiding the model toward analogous links for the emerging entity. \textbf{KGA-Oracle+RAG (KGA-O+RAG)} combines gold and retrieved context in random order (details in Appendix~\ref{sec:appendix:compared-models}). These variants allow us to assess the effects of KG context and retrieval on TKGU performance.}
    % \revklim{To evaluate TKGU with explicit access to the current KG state, we introduce a \emph{KG-aware} (KGA) baseline built on the same \texttt{GPT-5.1} as EDC+. KGA augments the model with context from the reference snapshot $G_t$, enabling state-dependent operations such as \opdeprecate{}. We evaluate three variants: \textbf{KGA-RAG}, using the top $k{=}32$ retrieved KG triples; \textbf{KGA-Oracle} (KGA-O), supplied with gold KG triples from the ground-truth TKGU operations; and \textbf{KGA-Oracle+RAG} (KGA-O+RAG), combining gold and retrieved triples. This allows us to evaluate how KG access affects TKGU performance.}

    \subsection{Metrics and Evaluation}
    \label{sec:metrics-and-evaluation}

%%%%%%%%%%% BEGIN V2
    We distinguish between two complementary evaluation protocols. The \textbf{\prsemanticcoverage{}} protocol measures how closely predicted triples match the knowledge in the ground-truth update, independent of canonicalization to KG identifiers.~Complementarily, we define the \textbf{\prexecutable{}} protocol to measure the correctness of the KG edits required by TKGU operations.~This protocol requires canonicalizing model-generated triples to KG identifiers and applying them as edits to a KG snapshot, conditioned on its current state (Appendix~\ref{sec:appendix:executable-protocol}).~\revklim{Under the \textit{\prsemanticcoverage{}} protocol, we use two metrics. \textbf{Completeness} \citep{jiang2024genres} measures the fraction of ground-truth triples whose cosine similarity with a predicted triple exceeds a fixed threshold (Appendix~\ref{sec:appendix:completeness}). \textbf{G-BERTScore-R} \citep{saha2021explagraphs} extends BERTScore \citep{zhang2020bertscore} to graph matching by treating each graph as a set of edges. It computes an optimal alignment between predicted and gold edges and reports recall.}
% Under the \textit{\prsemanticcoverage{}} protocol, we use \textbf{Completeness} \citep{jiang2024genres}, the fraction of ground-truth triples whose cosine similarity with a predicted triple exceeds a fixed threshold (Appendix~\ref{sec:appendix:completeness}); and \textbf{G-BERTScore-R} \citep{saha2021explagraphs}, an extension of BERTScore \citep{zhang2020bertscore} to graph matching that treats each graph as a set of edges, computes an optimal alignment between predicted and gold edges, and reports recall.
%%%%%%%%%%%%%%%%%%%%% START CAMERA READY CHANGE
%% BEGIN OLD PRE-CAMERA READY
% Under the \textit{\prexecutable{}} protocol, we define \textbf{Executable-R}, the recall of canonical $(s,p,o)$ exact matches against the ground-truth KG update, restricted to \opexists{} and \opadd{}, the operations naturally induced by closed-IE outputs from models such as ReLiK cIE \citep{orlando2024relik}.
%
%% END OLD PRE-CAMERA READY
%% BEGIN NEW PRE-CAMERA READY
% Executable-R is well-defined for \opexists{} and \opadd{}, whose entities already carry canonical identifiers in $V_t$, and for \opdeprecate{} triples present in $T_t$. We compute it over all text-supported gold triples; the 8.7\% of \opdeprecate{} test triples absent from $T_t$ (Section~\ref{sec:problem-definition}) therefore count as misses, making the reported \opdeprecate{} Executable-R conservative.
% Executable-R is well-defined for \opexists{}, \opadd{}, and \opdeprecate{}, whose entities already carry canonical identifiers in $V_t$.

\revklim{%Under the \textit{\prexecutable{}} protocol, we define \textbf{Executable-R}, the recall of canonical $(s,p,o)$ exact matches against the ground-truth KG update. Executable-R is well-defined for \opexists{} and \opadd{}, whose entities already contain canonical identifiers in $V_t$, and for \opdeprecate{} triples present in $T_t$, since a deprecation can only be executed on an existing triple; the 8.7\% of test \opdeprecate{} triples absent from $T_t$ are therefore scored only by the two \textit{\prsemanticcoverage{}} metrics. This differs from \opmintadd{} and \opinfer{}, whose emerging entities have no canonical identifier at inference time, for which we rely only on the two \textit{\prsemanticcoverage{}} metrics (\ie \textit{Completeness} and \textit{G-BERTScore-R}) above.%
Under the \textit{\prexecutable{}} protocol, we define \textbf{Executable-R}, the recall of canonical $(s,p,o)$ exact matches against the ground-truth KG update. Executable-R is computed for \opexists{} and \opadd{}, whose entities have canonical identifiers in $V_t$, and for the 91.3\% of test \opdeprecate{} triples present in $T_t$. All TKGU operations, including the remaining 8.7\% of \opdeprecate{} triples absent from $T_t$, are evaluated using the two \textit{\prsemanticcoverage{}} metrics (\ie \textit{Completeness} and \textit{G-BERTScore-R}). These are the only applicable metrics for \opmintadd{} and \opinfer{}, whose emerging entities lack canonical identifiers.%
}
%% END NEW PRE-CAMERA READY
%%%%%%%%%%%%%%%%%%%%% END CAMERA READY CHANGE

All three metrics are recall-based, reflecting the open-world assumption~\citep{razniewski2024completeness,yang2022rethinking,safavi2020evaluating}: predictions absent from our annotations may still be valid KG triples, making precision and F1 unreliable. \revklim{Moreover, the ground truth for every metric consists only of TKGU operations not flagged as unsupported during curation (Section~\ref{sec:quality-control}). An unsupported operation cannot be inferred from its passage, so including it would penalize models for not predicting facts absent from the text.}

    % ------------------------------------------------------------------------------------------------------------------
    % Table 3. C / G-R cells: emerge-prod:scripts/stats/generate_table3.py --pkl <D3 results file> (reproduces all 110 cells)
    % D3 = IE rows, config emerge-prod:config/evaluation/s04_evaluate_predictions/20261023_v1_0_1_reissue/config.json
    % D4 = KGA rows, config emerge-prod:config/evaluation/s04_evaluate_predictions/20261028_kg_aware_v1_0_1_reissue/config.json
    % E-R column (all 11 cells): emerge-stage:scripts/stats/camera_ready_reissue_20261028/executable_recall_deprecate_in_snapshot_20261030.py --config emerge-stage:config/stats/20261030_executable_recall_deprecate_in_snapshot/config.json
    %   Deprecate E-R only over gold triples in T_t (1,487 of 1,628); reads the D3 + D4 results files; stops unless it reproduces the pre-change cells.
    % KGA rows C / G-R: UNCERTAIN, read from the D4 evaluation log (no extraction script saved). Results files: see legend.
    % ------------------------------------------------------------------------------------------------------------------

    \begin{table}
    \caption{
    \textit{Completeness} (C), \textit{G-BERTScore-R} (G-R), and \textit{Executable-R} (E-R) for IE baselines and KG-aware (KGA) variants on the TKGU operations of Section~\ref{sec:problem-definition}. LLMs use \texttt{Mistral-small} (M-Sm), \texttt{Mistral-Large-2411} (M-Lg), and \texttt{GPT-5.1}.~All KGA variants use \texttt{GPT-5.1}. Best results are \textbf{bold}, second-best \underline{underlined}.~Gold-context variants KGA-Oracle (KGA-O) and KGA-Oracle+RAG (KGA-O+RAG) ($\dagger$) are excluded from the ranking.~\revklim{E-R is computed for \opexists{}, \opadd{}, and the \opdeprecate{} triples present in $T_t$ (Section~\ref{sec:metrics-and-evaluation}), but not for \opmintadd{} or \opinfer{} due to emerging entities without identifiers. Dashes indicate inapplicable metrics or unsupported operations.}
        % E-R is defined for \opexists{}, \opadd{}, and \opdeprecate{}, but not for \opmintadd{} or \opinfer{} due to emerging entities without identifiers.
}
    \label{tab:dataset_re_results}
\centering
    \rowcolors{4}{rowwhite}{rowgray}
    {\setlength{\tabcolsep}{5pt}
    \resizebox{\textwidth}{!}{\begin{tabular}{l|ccc|ccc|cc|cc|ccc}
      \toprule
      & \multicolumn{3}{c|}{\opexists} & \multicolumn{3}{c|}{\opadd} & \multicolumn{2}{c|}{\opmintadd} & \multicolumn{2}{c|}{\opinfer} & \multicolumn{3}{c}{\opdeprecate} \\
      \cmidrule(lr){2-4}\cmidrule(lr){5-7}\cmidrule(lr){8-9}\cmidrule(lr){10-11}\cmidrule(lr){12-14}
      Model & C$\uparrow$ & G-R$\uparrow$ & E-R$\uparrow$ & C$\uparrow$ & G-R$\uparrow$ & E-R$\uparrow$ & C$\uparrow$ & G-R$\uparrow$ & C$\uparrow$ & G-R$\uparrow$ & C$\uparrow$ & G-R$\uparrow$ & E-R$\uparrow$ \\
\midrule
\cellcolor{archEDCp} EDC+ GPT-5.1 & \textbf{25.8} & \textbf{66.5} & -- & \textbf{39.9} & \textbf{77.3} & -- & \textbf{35.6} & \textbf{77.3} & \textbf{21.9} & \textbf{72.4} & \textbf{34.8} & \textbf{53.2} & -- \\
\cellcolor{archEDCp} EDC+ M-Lg & 16.2 & 53.4 & -- & \underline{29.9} & 70.4 & -- & 27.9 & \underline{71.5} & 19.9 & 68.8 & \underline{31.7} & \underline{53.1} & -- \\
\cellcolor{archEDCp} EDC+ M-Sm & ~~8.8 & 45.5 & -- & 20.8 & 63.3 & -- & 16.3 & 62.6 & \underline{21.4} & \underline{70.8} & 14.5 & 28.6 & -- \\
\cellcolor{archKG} KGGen GPT-5.1 & 12.7 & 62.7 & -- & 15.3 & 68.1 & -- & 13.3 & 67.0 & ~~3.3 & 59.1 & -- & -- & -- \\
\cellcolor{archKG} KGGen M-Lg & 11.8 & 60.5 & -- & 16.1 & 67.7 & -- & 13.7 & 65.9 & ~~2.1 & 56.9 & -- & -- & -- \\
\cellcolor{archKG} KGGen M-Sm & ~~8.1 & 57.0 & -- & 13.1 & 65.8 & -- & 12.6 & 65.3 & ~~2.1 & 56.1 & -- & -- & -- \\
\cellcolor{archRAKG} RAKG M-Lg & 11.7 & 64.1 & -- & 19.3 & 71.0 & -- & 16.5 & 70.7 & ~~2.5 & 62.6 & -- & -- & -- \\
\cellcolor{archRAKG} RAKG M-Sm & 11.7 & \underline{64.1} & -- & 19.6 & \underline{71.3} & -- & 16.0 & 70.7 & ~~2.5 & 62.6 & -- & -- & -- \\
\cellcolor{archRELIK} ReLiK RE & \underline{21.9} & 59.4 & -- & 20.7 & 61.7 & -- & 17.1 & 60.9 & ~~3.8 & 48.8 & -- & -- & -- \\
\cellcolor{archRELIK} ReLiK cIE & 15.2 & 31.0 & \underline{16.2} & 18.2 & 57.5 & \textbf{15.3} & -- & -- & -- & -- & -- & -- & -- \\
\cellcolor{archOther} REBEL & 12.3 & 44.0 & -- & 13.6 & 56.3 & -- & 11.6 & 55.1 & ~~1.5 & 37.6 & -- & -- & -- \\
% realistic KG-aware baseline -- competes for bold/underline with the IE models above the rule
\cellcolor{archKGaware} KGA-RAG & 13.9 & 27.3 & \textbf{19.3} & 15.6 & 19.0 & \underline{12.7} & \underline{30.6} & 49.7 & 11.3 & 52.1 & 12.3 & 12.7 & \textbf{12.5} \\
\midrule
% ---- KG-aware UPPER BOUNDS (supplied gold KG pieces; $\dagger$; excluded from bold/underline ranking) ----
\cellcolor{archKGaware} KGA-O$^\dagger$ & 60.2 & 49.9 & 50.0 & 46.6 & 42.9 & 39.4 & 43.2 & 47.6 & 50.3 & 75.9 & 37.4 & 35.8 & 37.6 \\
\cellcolor{archKGaware} KGA-O+RAG$^\dagger$ & 52.5 & 49.0 & 42.9 & 27.3 & 27.0 & 18.1 & 39.3 & 53.1 & 45.0 & 75.5 & 18.7 & 18.4 & 18.5 \\
      \bottomrule
    \end{tabular}}}
    \end{table}

    \section{Experiments and Analysis}
    % \Tabref{tab:dataset_re_results} reports model performance across all TKGU operations. Bootstrap resampling (100 draws with replacement from the 3{,}500-instance test set) confirms these comparisons are stable:
    % \revklim{median 95\% CI width $\leq 4.3$ pts}
    % % median 95\% CI width $\leq 1.9$ pts
    % across all (metric, operation) pairs~\revklim{of Completeness and G-BERTScore-R for the eleven IE baselines,} and
    % % Spearman $\rho \geq 0.92$ between
    % \revklim{Spearman $\rho \geq 0.87$ between}
    % the original test-set ranking and bootstrap rankings (Appendix~\ref{sec:appendix:bootstrap-stability}).
    \Tabref{tab:dataset_re_results} reports model performance across all TKGU operations. Bootstrap resampling (100 draws with replacement from the 3{,}500-instance test set) confirms stable comparisons \revklim{for Completeness and G-BERTScore-R across the eleven IE baselines. Specifically, median 95\% CI widths are $\leq 4.3$ points across all (metric, operation) pairs. Spearman rank correlations between the original test-set and bootstrap rankings are $\rho \geq 0.87$} (Appendix~\ref{sec:appendix:bootstrap-stability}). The following paragraphs address key research questions and lay groundwork for future studies using the introduced TKGU operations.

    \noindent\textbf{What is the overall performance?}
    Overall, \textit{Completeness} scores are low (\textsc{C} in \Tabref{tab:dataset_re_results}), as this metric rewards only near-exact matches (above an established threshold) to the ground-truth triples representing KG changes. In contrast, G-BERTScore-R (\textsc{G-R} in \Tabref{tab:dataset_re_results}) is comparatively higher, indicating that models often produce semantically similar triples that capture relevant knowledge but do not exactly match the specific triples expected in the ground-truth KG updates,
    as illustrated in Section~\ref{sec:appendix:qualitative-predictions}.
    We hypothesize that this discrepancy arises 
    \revklim{because IE baselines without KG context lack access to KG content and structure},
    % because the models lack access to KG content and structure, 
    which prevents them from determining the precise nature of the knowledge being added and the relation types involved.~This highlights a promising direction for future work: developing IE models that can also leverage the internal structure of KGs.

    \noindent \textbf{How does performance vary across TKGU operations and between models?} 
    \revklim{Among models without gold KG context, EDC+ with \texttt{GPT-5.1} achieves the highest Completeness and G-BERTScore-R across all TKGU operations. Its advantage over the other IE baselines is particularly pronounced for \opdeprecate{} and \opinfer{}, which its adapted prompt explicitly addresses.~With gold KG context, however, KGA-O achieves higher Completeness than EDC+ on all operations, while KGA-O+RAG also outperforms it on \opinfer{} in both metrics. 
    In contrast, KGA-RAG remains below EDC+ on both metrics across all operations, partly because it depends on the quality of the retrieved KG context, whereas EDC+ is not bound to the KG schema.
    % In contrast, KGA-RAG remains below EDC+ on both metrics across all operations, highlighting the challenge of retrieving relevant KG context. 
    On \opinfer{}, for instance, Completeness rises from 11.3 with retrieved context (KGA-RAG) to 50.3 with gold context (KGA-O), compared with 21.9 for EDC+. Beyond the semantic performance captured by Completeness and G-BERTScore-R, KGA offers an additional advantage over EDC+: it produces canonicalized predictions for \opexists{}, \opadd{}, and \opdeprecate{}, enabling their direct execution against the KG and evaluation using Executable-R.
    %(see the analysis of executable KG updates below).
    }
    % Overall, our adapted EDC+ model consistently outperforms the other benchmarked approaches. This improvement is particularly pronounced for the \opdeprecate{} and \opinfer{} operations. While existing IE models fail to identify deprecations of KG triples, we show that a simple prompt-based adaptation enables EDC+ to produce meaningful predictions. Similarly, for \opinfer{}, EDC+ substantially outperforms all other models, highlighting the strong potential of LLM-based approaches to capture the knowledge underlying TKGU operations.

    EDC+ outperforms KGGen and RAKG, particularly on the \textit{Completeness} metric, when evaluated with the same underlying LLMs. We hypothesize that this advantage is related to the use of five in-context demonstrations drawn outside the test set.
    Disabling this in-context learning (zero-shot) for EDC+ with \texttt{GPT-5.1} substantially reduces performance (Appendix~\ref{sec:appendix:icl-ablation}), with the largest drops on \opdeprecate{} (up to 33.4 points in Completeness and 50.0 in G-BERTScore-R).
    This suggests that the in-context demonstrations provide benefits beyond standard few-shot formatting, potentially reflecting implicit exposure to KG structure or content.
    However, the current ablation cannot disentangle this from general few-shot effects, leaving this for future work. We also note that the span-based ReLiK model remains consistently competitive, particularly on Completeness, despite relying on BERT-based architectures with substantially lower inference cost than LLM-based approaches.

    % ------------------------------------------------------------------------------------------------------------------
    % Figure 3 (also appendix Figure 7, same notebook): emerge-stage:src/stats/evaluation/figure_deltas_snapshots_reissue_20261028.ipynb
    % Run: PKL_PATH=<D3 results file> SAVE_DIR=<out> jupyter nbconvert --to notebook --execute <notebook>   (D3: see legend)
    % Output: plot_combined_single_row_experimental_v2_empty_pred_fix.pdf; test: check_figure_values.py (overall values = Table 3)
    % ------------------------------------------------------------------------------------------------------------------

    \begin{figure}[t]
    \centering
    \includegraphics[trim=0 0 0 0,clip,width=\linewidth]{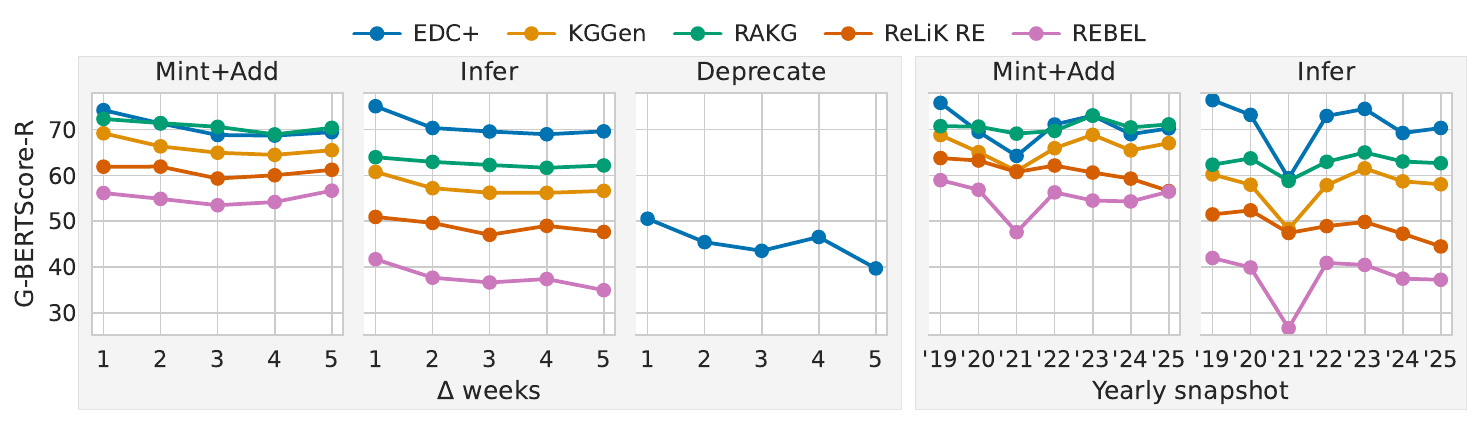}
    \caption{G-BERTScore-R across KG weekly deltas (left panel) for \opmintadd{}, \opinfer{}, and \opdeprecate{}, and across yearly KG snapshots (right panel) for \opmintadd{} and \opinfer{}. Performance declines with larger weekly deltas, indicating increased difficulty with stale KGs.~Across yearly snapshots, we observe a performance drop in 2021 for both \opmintadd{} and \opinfer{} TKGU operations.
    }
    \label{fig:deltas-performance}
    \end{figure}

    \noindent\textbf{What is the performance across different snapshots?}
    Performance is broadly stable across snapshots, with a consistent drop around 2021 for emerging-entity operations (\opmintadd{}, \opinfer{}).
    This effect is shown in the right panel of \Figref{fig:deltas-performance}, which reports average G-BERTScore-R across yearly KG snapshots for \opmintadd{} and \opinfer{} (see Section~\ref{sec:appendix:quantitative-predictions} for results on all operations and metrics).
    % BEGIN new version
    A breakdown by relation type (Appendix~\ref{sec:appendix:2021-dip}) shows that KG changes on the 2021 snapshot are dominated by
    % event-driven
    relations (\eg \textit{significant event}, \textit{target}, \textit{has cause}) that are rare in the KG: the Spearman correlation between dataset and KG relation frequencies drops to $\rho \approx 0.05$ for \opinfer{} and \opmintadd{} in 2021
    \revklim{(vs.\ $0.31$--$0.58$ in other years)}.
    % (vs.\ $0.30$--$0.57$ in other years).
    % These relations account for $\sim$45\% of 2021
    \revklim{These relations account for $\sim$40\% of 2021}
    \opinfer{} and \opmintadd{} triples,
    % triples (and similarly for \opmintadd{})
    and yield lower scores \revklim{($-17.8$ G-BERTScore-R for} \opinfer{}, $-8.0$ for \opmintadd{}), explaining the observed dip. They are largely associated with
    % contemporaneous
    political events
    % (e.g., the U.S. presidential transition and the January 2021 Capitol attack),
    (e.g., the January 2021 Capitol attack),
    which introduced a burst of atypical relations into Wikidata.
    % END new version

    \noindent\textbf{What is the performance on increasing temporal KG deltas?}~We observe a general performance decline as the temporal gap between updates and the reference KG snapshot increases.~This effect is most pronounced for \opmintadd{}, \opinfer{}, and \opdeprecate{}, as shown in the left panel of \Figref{fig:deltas-performance}, which reports average G-BERTScore-R across increasing weekly KG deltas for these operations (see Section~\ref{sec:appendix:quantitative-predictions} for results on all operations and metrics).
    This trend indicates that extracting relevant knowledge from text becomes increasingly challenging as the underlying KG grows more outdated.

    \textbf{To what extent can existing IE models execute KG updates?} \textit{\mprexecutable{}} updating requires three capabilities that, to the best of our knowledge, are absent or only partially supported in current state-of-the-art models: (i) \textit{linking entities to canonical KG identifiers}, required for all TKGU operations and available only in KG-linked models (\textsc{KG Linked} column, \Tabref{tab:comparison_current_architectures}); (ii) \textit{generating and integrating emerging entities}, required for \opmintadd{} and \opinfer{}, and only partially supported by our \textit{KG-aware} (KGA) baseline; and (iii) \textit{identifying existing triples for deprecation},
    required for \opdeprecate{}, \revklim{and supported by our KGA baseline through access to the current KG state.}
    % (iii) \textit{identifying existing triples for deprecation}, required for \opdeprecate{} and supported by our \textit{KG-aware} (KGA) baseline, which conditions on the current KG state to identify candidate triples for deprecation.
    % required for \opdeprecate{}, and likewise unsupported.

Per-operation feasibility under the \textit{\prexecutable{}} evaluation protocol (Section~\ref{sec:metrics-and-evaluation}) is detailed in Appendix~\ref{sec:appendix:executable-protocol}: exact-match canonical-ID evaluation is well-defined for \opexists{}, \opadd{}, and \opdeprecate{} triples in $T_t$, whose entities already exist in $V_t$. In contrast, \opmintadd{} and \opinfer{} involve emerging entities and thus require post-hoc canonicalization for executable evaluation. Since current IE models do not produce executable outputs for these operations, we evaluate them with \textit{semantic coverage} Completeness and G-BERTScore-R. Notably, the alignment underlying G-BERTScore-R could also serve as a basis for post-hoc canonicalization in future executable models (see Appendix~\ref{sec:appendix:executable-protocol}).

\revklim{Only ReLiK cIE \citep{orlando2024relik} and our \emph{KG-aware} (KGA) method produce canonicalized predictions, which can be executed as KG edits and scored with Executable-R.
% ReLiK cIE \citep{orlando2024relik} and our \emph{KG-aware} (KGA) method produce canonicalized predictions. 
By incorporating reference KG context into \texttt{GPT-5.1}, KGA can also produce executable \opdeprecate{} updates. KGA-O achieves the highest Executable-R: 50.0 on \opexists{}, 39.4 on \opadd{}, and 37.6 on \opdeprecate{}. However, even with gold triples provided for \opexists{} and \opdeprecate{}, these modest scores suggest that identifying which triples the passage supports or invalidates remains challenging.
% Yet even with gold triples provided for \opexists{} and \opdeprecate{}, its scores suggest that identifying which triples the passage supports or invalidates remains challenging. 
KGA-RAG reaches only 19.3, 12.7, and 12.5, respectively, suggesting that retrieval is another major bottleneck. Compared with ReLiK cIE, KGA-RAG performs better on \opexists{} (19.3 vs.\ 16.2) but worse on \opadd{} (12.7 vs.\ 15.3), showing that this much lighter span-based approach remains competitive, even though it requires supervised training. KGA-O+RAG also underperforms KGA-O, suggesting that retrieved distractors can hurt even with gold context. Related work such as \cite{lu2025karma} also integrates text with graph structure, but constructs the graph from the processed corpus rather than updating a fixed external KG with canonical identifiers.}

    \section{Conclusion}
    In this work, we introduced \datasetname, the first dataset to cover all text-driven knowledge graph updating (TKGU) operations required to keep KGs aligned with emerging knowledge from textual sources.~Evaluation of five representative information extraction (IE) methods
    % on a dataset subset
    revealed a gap in current IE models to extract new information from text while accounting for existing KG content and structure.~This suggests that future work should focus on designing IE methods capable of interacting with both emerging knowledge in text and the evolving content and structure of KGs. 
    % Limitations and future directions are discussed in Appendix~\ref{sec:appendix:limitations}. 
    Furthermore, we see \datasetname{} and the TKGU task introduced in this paper as a starting point for future work on text-driven KG updating across \textit{specialized} domains and knowledge sources. Additional \textit{limitations} of our work, along with potential \textit{directions for future research}, are discussed in Appendix~\ref{sec:appendix:limitations}.

\begin{ack}
   %  \section*{Acknowledgments}
\revklim{Klim Zaporojets acknowledges support from the European Union's Horizon Europe programme through a Marie Skłodowska-Curie Actions (MSCA) Postdoctoral Fellowship (KG-PROVENANCE, Grant Agreement No.~101146515). Daniel Daza was funded by a gift from Accenture LLP. Ira Assent was partially supported by the Pioneer Centre for AI, DNRF grant number
P1.~The authors thank Nicholas Monath for valuable discussions.}
\end{ack}
    \medskip

    {
    \small
    \bibliographystyle{plain}
    \bibliography{v4_neurips_2026/references}
    }

    \newpage
    \section*{NeurIPS Paper Checklist}

\begin{enumerate}

\item {\bf Claims}
    \item[] Question: Do the main claims made in the abstract and introduction accurately reflect the paper's contributions and scope?
    \item[] Answer: \answerYes{} % \answerTODO{} % Replace by \answerYes{}, \answerNo{}, or \answerNA{}.
    \item[] Justification: The abstract and the introduction state our four contributions: (i) a formalization of the task of KG maintenance from text by defining text-driven KG updating (TKGU) operations (Section~\ref{sec:problem-definition}) that can be applied to a given KG snapshot, going beyond traditional information extraction formulations, (ii) a benchmark dataset (\datasetname{}) pairing Wikipedia passages with Wikidata KG updates across the five TKGU operation types, (iii) a pipeline for constructing such datasets by aligning KG updates with textual evidence, followed by LLM-guided filtering and manual curation, and (iv) an evaluation of representative IE models on these operations, identifying a gap in KG-aware updating. These claims match the experimental results in Section~\ref{sec:experimental-setup} and \Tabref{tab:dataset_re_results}. The scope (general-domain Wikipedia/Wikidata, seven yearly snapshots 2019--2025, IE evaluation on a 3{,}500-instance subset) and assumptions (e.g., hyperlink-based alignment, English Wikipedia, entity-only triples) are outlined in abstract and introduction, and explicitly delimited in Sections~\ref{sec:data-collection} and~\ref{sec:experimental-setup}, with extensions and limitations discussed in Section~\ref{sec:our-dataset} (Extensibility subsection) and Appendix~\ref{sec:appendix:limitations}.
    \item[] Guidelines:
    \begin{itemize}
        \item The answer \answerNA{} means that the abstract and introduction do not include the claims made in the paper.
        \item The abstract and/or introduction should clearly state the claims made, including the contributions made in the paper and important assumptions and limitations. A \answerNo{} or \answerNA{} answer to this question will not be perceived well by the reviewers. 
        \item The claims made should match theoretical and experimental results, and reflect how much the results can be expected to generalize to other settings. 
        \item It is fine to include aspirational goals as motivation as long as it is clear that these goals are not attained by the paper. 
    \end{itemize}

\item {\bf Limitations}
    \item[] Question: Does the paper discuss the limitations of the work performed by the authors?
    \item[] Answer: \answerYes{} % Replace by \answerYes{}, \answerNo{}, or \answerNA{}.
    \item[] Justification: A dedicated limitations discussion is provided in Appendix~\ref{sec:appendix:limitations}, covering: (i) the focus on textual evidence only (multimodal sources are out of scope); (ii) the temporal-window mismatch between Wikipedia text and Wikidata KG updates (some events are documented in text months/years before they enter the KG); (iii) the entities-only scope (literal-valued attributes such as dates, lengths, counts are excluded by construction); and (iv) coverage limited to the subset of Wikidata changes grounded in Wikipedia text via hyperlink alignment. Each limitation is paired with a concrete future-work direction.
    \item[] Guidelines:
    \begin{itemize}
        \item The answer \answerNA{} means that the paper has no limitation while the answer \answerNo{} means that the paper has limitations, but those are not discussed in the paper. 
        \item The authors are encouraged to create a separate ``Limitations'' section in their paper.
        \item The paper should point out any strong assumptions and how robust the results are to violations of these assumptions (e.g., independence assumptions, noiseless settings, model well-specification, asymptotic approximations only holding locally). The authors should reflect on how these assumptions might be violated in practice and what the implications would be.
        \item The authors should reflect on the scope of the claims made, e.g., if the approach was only tested on a few datasets or with a few runs. In general, empirical results often depend on implicit assumptions, which should be articulated.
        \item The authors should reflect on the factors that influence the performance of the approach. For example, a facial recognition algorithm may perform poorly when image resolution is low or images are taken in low lighting. Or a speech-to-text system might not be used reliably to provide closed captions for online lectures because it fails to handle technical jargon.
        \item The authors should discuss the computational efficiency of the proposed algorithms and how they scale with dataset size.
        \item If applicable, the authors should discuss possible limitations of their approach to address problems of privacy and fairness.
        \item While the authors might fear that complete honesty about limitations might be used by reviewers as grounds for rejection, a worse outcome might be that reviewers discover limitations that aren't acknowledged in the paper. The authors should use their best judgment and recognize that individual actions in favor of transparency play an important role in developing norms that preserve the integrity of the community. Reviewers will be specifically instructed to not penalize honesty concerning limitations.
    \end{itemize}

\item {\bf Theory assumptions and proofs}
    \item[] Question: For each theoretical result, does the paper provide the full set of assumptions and a complete (and correct) proof?
    \item[] Answer: \answerNA{} % Replace by \answerYes{}, \answerNo{}, or \answerNA{}.
    \item[] Justification: The paper introduces a benchmark dataset and an empirical evaluation of existing IE models; it does not contain theorems or formal proofs. Section~\ref{sec:problem-definition} provides \textit{formal definitions} of the five TKGU operations (\opexists{}, \opadd{}, \opmintadd{}, \opinfer{}, \opdeprecate{}) over a KG snapshot $G_t = (V_t, T_t)$ and a delta window, but no theoretical claim is made about these definitions.
    % \justificationTODO{}
    \item[] Guidelines:
    \begin{itemize}
        \item The answer \answerNA{} means that the paper does not include theoretical results. 
        \item All the theorems, formulas, and proofs in the paper should be numbered and cross-referenced.
        \item All assumptions should be clearly stated or referenced in the statement of any theorems.
        \item The proofs can either appear in the main paper or the supplemental material, but if they appear in the supplemental material, the authors are encouraged to provide a short proof sketch to provide intuition. 
        \item Inversely, any informal proof provided in the core of the paper should be complemented by formal proofs provided in appendix or supplemental material.
        \item Theorems and Lemmas that the proof relies upon should be properly referenced. 
    \end{itemize}

    \item {\bf Experimental result reproducibility}
    \item[] Question: Does the paper fully disclose all the information needed to reproduce the main experimental results of the paper to the extent that it affects the main claims and/or conclusions of the paper (regardless of whether the code and data are provided or not)?
    \item[] Answer: \answerYes{} % \answerTODO{} % Replace by \answerYes{}, \answerNo{}, or \answerNA{}.
    \item[] Justification: The full pipeline (dataset construction, LLM-based curation, evaluation of all benchmarked IE models) is described in Sections~\ref{sec:our-dataset}, \ref{sec:data-collection}, \ref{sec:quality-control}, and~\ref{sec:experimental-setup}. Per-baseline configurations and modifications (in particular, the EDC+ extension) are detailed in Appendix~\ref{sec:appendix:compared-models}; the EDC+ prompts are reproduced in Appendix~\ref{sec:appendix:edc-prompts}; the LLM-curation prompts are reproduced in Appendix~\ref{sec:appendix:prompts}; the executable-protocol per-operation feasibility is in Appendix~\ref{sec:appendix:executable-protocol}; the completeness threshold and bootstrap protocol are in Appendices~\ref{sec:appendix:completeness} and~\ref{sec:appendix:bootstrap-stability}. In addition, the open-source code at \url{https://github.com/klimzaporojets/emerge} releases the SLURM sbatch wrappers, JSON configs, 
    % (under \texttt{config/dataset/emerge/}), 
    per-step Python pipeline,
    % (\texttt{src/dataset/emerge/}), 
    and evaluation scripts. 
    % (\texttt{src/evaluation/}). 
    The dataset is hosted on Hugging Face (\url{https://huggingface.co/datasets/klimzaporojets/emerge-benchmark}) and accompanied by a Datasheet (Appendix~\ref{sec:appendix:dataset-sheet:composition}, following \cite{gebru2021datasheets}) plus validated Croissant 1.1 + Responsible AI metadata. % \justificationTODO{}
    \item[] Guidelines:
    \begin{itemize}
        \item The answer \answerNA{} means that the paper does not include experiments.
        \item If the paper includes experiments, a \answerNo{} answer to this question will not be perceived well by the reviewers: Making the paper reproducible is important, regardless of whether the code and data are provided or not.
        \item If the contribution is a dataset and\slash or model, the authors should describe the steps taken to make their results reproducible or verifiable. 
        \item Depending on the contribution, reproducibility can be accomplished in various ways. For example, if the contribution is a novel architecture, describing the architecture fully might suffice, or if the contribution is a specific model and empirical evaluation, it may be necessary to either make it possible for others to replicate the model with the same dataset, or provide access to the model. In general. releasing code and data is often one good way to accomplish this, but reproducibility can also be provided via detailed instructions for how to replicate the results, access to a hosted model (e.g., in the case of a large language model), releasing of a model checkpoint, or other means that are appropriate to the research performed.
        \item While NeurIPS does not require releasing code, the conference does require all submissions to provide some reasonable avenue for reproducibility, which may depend on the nature of the contribution. For example
        \begin{enumerate}
            \item If the contribution is primarily a new algorithm, the paper should make it clear how to reproduce that algorithm.
            \item If the contribution is primarily a new model architecture, the paper should describe the architecture clearly and fully.
            \item If the contribution is a new model (e.g., a large language model), then there should either be a way to access this model for reproducing the results or a way to reproduce the model (e.g., with an open-source dataset or instructions for how to construct the dataset).
            \item We recognize that reproducibility may be tricky in some cases, in which case authors are welcome to describe the particular way they provide for reproducibility. In the case of closed-source models, it may be that access to the model is limited in some way (e.g., to registered users), but it should be possible for other researchers to have some path to reproducing or verifying the results.
        \end{enumerate}
    \end{itemize}

\item {\bf Open access to data and code}
    \item[] Question: Does the paper provide open access to the data and code, with sufficient instructions to faithfully reproduce the main experimental results, as described in supplemental material?
    \item[] Answer: \answerYes{} % \answerTODO{} % Replace by \answerYes{}, \answerNo{}, or \answerNA{}.
    \item[] Justification: \datasetname{} is publicly released under CC BY-SA 4.0 on Hugging Face (\url{https://huggingface.co/datasets/klimzaporojets/emerge-benchmark}), accompanied by a validated Croissant 1.1 + Responsible AI metadata file. The code is open-sourced at \url{https://github.com/klimzaporojets/emerge} (Apache 2.0 license, \texttt{LICENSE}; data license \texttt{LICENSE-DATA}, CC BY-SA 4.0). The repository contains the per-step Python pipelines, JSON configurations with hyperparameters, SLURM sbatch wrappers, a dataset-download script, and an end-to-end test suite. End-to-end usage instructions are in the in-repo \texttt{README.md} file.
    % \item[] Justification: \datasetname{} is publicly released under CC BY-SA 4.0 on Hugging Face (\url{https://huggingface.co/datasets/klimzaporojets/emerge-benchmark}), accompanied by a validated Croissant 1.1 + Responsible AI metadata file. The code is open-sourced at \url{https://github.com/klimzaporojets/emerge} (Apache 2.0 license, \texttt{LICENSE}; data license \texttt{LICENSE-DATA}, CC BY-SA 4.0). The repository contains the per-step Python pipelines (\texttt{src/dataset/emerge/}, \texttt{src/dataset/wikidata/}, \texttt{src/dataset/wikipedia/}, \texttt{src/evaluation/}), JSON configurations (\texttt{config/dataset/emerge/}, \texttt{config/evaluation/}, \texttt{config/benchmarks/}), SLURM sbatch wrappers (\texttt{scripts/slurm/}), a dataset-download script (\texttt{scripts/download\_data.sh}), an end-to-end test (\texttt{scripts/test\_end\_to\_end.sh}), and the evaluation runner (\texttt{scripts/run/evaluate.sh}). End-to-end usage instructions are in the in-repo top-level \texttt{README.md} and \texttt{src/dataset/README.md} (with a separate \texttt{src/evaluation/README.md} for evaluation).
    \item[] Guidelines:
    \begin{itemize}
        \item The answer \answerNA{} means that paper does not include experiments requiring code.
        \item Please see the NeurIPS code and data submission guidelines (\url{https://neurips.cc/public/guides/CodeSubmissionPolicy}) for more details.
        \item While we encourage the release of code and data, we understand that this might not be possible, so \answerNo{} is an acceptable answer. Papers cannot be rejected simply for not including code, unless this is central to the contribution (e.g., for a new open-source benchmark).
        \item The instructions should contain the exact command and environment needed to run to reproduce the results. See the NeurIPS code and data submission guidelines (\url{https://neurips.cc/public/guides/CodeSubmissionPolicy}) for more details.
        \item The authors should provide instructions on data access and preparation, including how to access the raw data, preprocessed data, intermediate data, and generated data, etc.
        \item The authors should provide scripts to reproduce all experimental results for the new proposed method and baselines. If only a subset of experiments are reproducible, they should state which ones are omitted from the script and why.
        \item At submission time, to preserve anonymity, the authors should release anonymized versions (if applicable).
        \item Providing as much information as possible in supplemental material (appended to the paper) is recommended, but including URLs to data and code is permitted.
    \end{itemize}

\item {\bf Experimental setting/details}
    \item[] Question: Does the paper specify all the training and test details (e.g., data splits, hyperparameters, how they were chosen, type of optimizer) necessary to understand the results?
    \item[] Answer: \answerYes{} % \answerTODO{} % Replace by \answerYes{}, \answerNo{}, or \answerNA{}.
    \item[] Justification: The benchmark evaluation setting (test-set construction by stratified subsampling of 100 instances per delta per snapshot, resulting in a total of 3{,}500 instances) is described in Section~\ref{sec:experimental-setup}. The metrics (Completeness threshold, G-BERTScore-R, Executable-R) and their definitions are in Section~\ref{sec:metrics-and-evaluation} and Appendices~\ref{sec:appendix:completeness} and~\ref{sec:appendix:executable-protocol}. The benchmarked models and their configurations are in Section~\ref{sec:models} and Appendix~\ref{sec:appendix:compared-models}; the EDC+ extension (in-context demonstrations, prompts) is fully specified in Appendix~\ref{sec:appendix:edc-prompts}, and the in-context-vs-zero-shot ablation is reported in Appendix~\ref{sec:appendix:icl-ablation}. The recommended train/validation/test split (earlier snapshots for training, later snapshots for testing) is described in the Datasheet (Appendix~\ref{sec:appendix:dataset-sheet:composition}). All run-time hyperparameters (LLM API parameters, retry policies, prompt templates) are specified in JSON config files 
    % under \texttt{config/dataset/emerge/} 
    in the public repository (\url{https://github.com/klimzaporojets/emerge}). % \justificationTODO{}
    \item[] Guidelines:
    \begin{itemize}
        \item The answer \answerNA{} means that the paper does not include experiments.
        \item The experimental setting should be presented in the core of the paper to a level of detail that is necessary to appreciate the results and make sense of them.
        \item The full details can be provided either with the code, in appendix, or as supplemental material.
    \end{itemize}

\item {\bf Experiment statistical significance}
    \item[] Question: Does the paper report error bars suitably and correctly defined or other appropriate information about the statistical significance of the experiments?
    \item[] Answer: \answerYes{} % Replace by \answerYes{}, \answerNo{}, or \answerNA{}.
    \item[] Justification: We assess the stability of \Tabref{tab:dataset_re_results} via bootstrap resampling, reported in Appendix~\ref{sec:appendix:bootstrap-stability}: $B=100$ samples drawn with replacement from the 3{,}500-instance test set, per-cell median and maximum 95\% confidence-interval widths across the eleven benchmarked models, and Spearman rank correlations $\rho$ (minimum and mean over the 100 bootstrap rankings) against the full-test ranking. The variability source is test-set resampling; the CI computation is the empirical 2.5\%/97.5\% percentiles of the 100 bootstrap means per cell; the Spearman $\rho$ summarizes whether model rankings are preserved. Numerical results (Table~\ref{tab:bootstrap-stability}) show 
    % median 95\% CI width $\leq 1.9$ pts and Spearman $\rho \geq 0.92$ across 
    median 95\% CI width $\leq 4.3$ pts and Spearman $\rho \geq 0.87$ across
    all (metric, operation) cells, which is also summarized in the main-text discussion of the experimental results.
    % \justificationTODO{}
    \item[] Guidelines:
    \begin{itemize}
        \item The answer \answerNA{} means that the paper does not include experiments.
        \item The authors should answer \answerYes{} if the results are accompanied by error bars, confidence intervals, or statistical significance tests, at least for the experiments that support the main claims of the paper.
        \item The factors of variability that the error bars are capturing should be clearly stated (for example, train/test split, initialization, random drawing of some parameter, or overall run with given experimental conditions).
        \item The method for calculating the error bars should be explained (closed form formula, call to a library function, bootstrap, etc.)
        \item The assumptions made should be given (e.g., Normally distributed errors).
        \item It should be clear whether the error bar is the standard deviation or the standard error of the mean.
        \item It is OK to report 1-sigma error bars, but one should state it. The authors should preferably report a 2-sigma error bar than state that they have a 96\% CI, if the hypothesis of Normality of errors is not verified.
        \item For asymmetric distributions, the authors should be careful not to show in tables or figures symmetric error bars that would yield results that are out of range (e.g., negative error rates).
        \item If error bars are reported in tables or plots, the authors should explain in the text how they were calculated and reference the corresponding figures or tables in the text.
    \end{itemize}

\item {\bf Experiments compute resources}
    \item[] Question: For each experiment, does the paper provide sufficient information on the computer resources (type of compute workers, memory, time of execution) needed to reproduce the experiments?
    \item[] Answer: \answerYes{} % Replace by \answerYes{}, \answerNo{}, or \answerNA{}.
    \item[] Justification: the compute resources required for both the dataset creation and execution of benchmarks are described in Appendix~\ref{sec:appendix:compute-resources}.
    % \justificationTODO{}
    \item[] Guidelines:
    \begin{itemize}
        \item The answer \answerNA{} means that the paper does not include experiments.
        \item The paper should indicate the type of compute workers CPU or GPU, internal cluster, or cloud provider, including relevant memory and storage.
        \item The paper should provide the amount of compute required for each of the individual experimental runs as well as estimate the total compute. 
        \item The paper should disclose whether the full research project required more compute than the experiments reported in the paper (e.g., preliminary or failed experiments that didn't make it into the paper). 
    \end{itemize}
    
\item {\bf Code of ethics}
    \item[] Question: Does the research conducted in the paper conform, in every respect, with the NeurIPS Code of Ethics \url{https://neurips.cc/public/EthicsGuidelines}?
    \item[] Answer: \answerYes{} % Replace by \answerYes{}, \answerNo{}, or \answerNA{}.
    \item[] Justification: We have reviewed the NeurIPS Code of Ethics. The dataset is built exclusively from publicly available, factual content (Wikipedia and Wikidata under CC BY-SA and CC0, respectively) without scraping private or restricted sources. No personal data beyond what is already publicly published in Wikipedia/Wikidata is collected (see Datasheet, Appendix~\ref{sec:appendix:dataset-sheet:composition}). Human annotation was performed by two expert co-authors (researchers with PhDs); no external crowdsourcing was used and no personal identifiers were collected from annotators.
    \item[] Guidelines:
    \begin{itemize}
        \item The answer \answerNA{} means that the authors have not reviewed the NeurIPS Code of Ethics.
        \item If the authors answer \answerNo, they should explain the special circumstances that require a deviation from the Code of Ethics.
        \item The authors should make sure to preserve anonymity (e.g., if there is a special consideration due to laws or regulations in their jurisdiction).
    \end{itemize}

\item {\bf Broader impacts}
    \item[] Question: Does the paper discuss both potential positive societal impacts and negative societal impacts of the work performed?
    \item[] Answer: \answerYes{} % Replace by \answerYes{}, \answerNo{}, or \answerNA{}.
    \item[] Justification: Positive impacts are discussed throughout the paper: \datasetname{} enables systematic evaluation of IE models on knowledge-graph maintenance under evolving textual evidence, supporting the downstream goal of keeping public knowledge bases (Wikidata-style) in sync with newly emerging facts (introduction; Section~\ref{sec:our-dataset}).
    Potential extensibility to other socially relevant domains beyond Wikipedia/Wikidata are discussed in Section~\ref{sec:extensibility}.
    Limitations and caveats around scope, coverage, and applicability are discussed in Appendix~\ref{sec:appendix:limitations}. Potential negative-impact considerations specific to a benchmark dataset are addressed item-by-item in the Datasheet (Appendix~\ref{sec:appendix:dataset-sheet:composition}). A short broader-impact paragraph is provided in Appendix~\ref{sec:appendix:broader-impact}.
    \item[] Guidelines:
    \begin{itemize}
        \item The answer \answerNA{} means that there is no societal impact of the work performed.
        \item If the authors answer \answerNA{} or \answerNo, they should explain why their work has no societal impact or why the paper does not address societal impact.
        \item Examples of negative societal impacts include potential malicious or unintended uses (e.g., disinformation, generating fake profiles, surveillance), fairness considerations (e.g., deployment of technologies that could make decisions that unfairly impact specific groups), privacy considerations, and security considerations.
        \item The conference expects that many papers will be foundational research and not tied to particular applications, let alone deployments. However, if there is a direct path to any negative applications, the authors should point it out. For example, it is legitimate to point out that an improvement in the quality of generative models could be used to generate Deepfakes for disinformation. On the other hand, it is not needed to point out that a generic algorithm for optimizing neural networks could enable people to train models that generate Deepfakes faster.
        \item The authors should consider possible harms that could arise when the technology is being used as intended and functioning correctly, harms that could arise when the technology is being used as intended but gives incorrect results, and harms following from (intentional or unintentional) misuse of the technology.
        \item If there are negative societal impacts, the authors could also discuss possible mitigation strategies (e.g., gated release of models, providing defenses in addition to attacks, mechanisms for monitoring misuse, mechanisms to monitor how a system learns from feedback over time, improving the efficiency and accessibility of ML).
    \end{itemize}
    
\item {\bf Safeguards}
    \item[] Question: Does the paper describe safeguards that have been put in place for responsible release of data or models that have a high risk for misuse (e.g., pre-trained language models, image generators, or scraped datasets)?
    \item[] Answer: \answerNA{} % Replace by \answerYes{}, \answerNo{}, or \answerNA{}.
    \item[] Justification: \datasetname{} consists of factual, public content from Wikipedia and Wikidata (released by their upstream maintainers under CC BY-SA 3.0+ and CC0, respectively); we redistribute it under CC BY-SA 4.0. We do not release any new pre-trained models, generators, or scraped content posing dual-use or high-misuse risk. The dataset is text-only and contains no offensive, sensitive, or biometric content (Datasheet, Appendix~\ref{sec:appendix:dataset-sheet:composition}). The LLMs used for curation (Llama-3.1-405B-Instruct-GPTQ-INT4) and benchmark evaluation (GPT-5.1, Mistral-Large-2411, Mistral-small) are not redistributed; they are accessed via their respective official channels under the providers' usage terms.
    \item[] Guidelines:
    \begin{itemize}
        \item The answer \answerNA{} means that the paper poses no such risks.
        \item Released models that have a high risk for misuse or dual-use should be released with necessary safeguards to allow for controlled use of the model, for example by requiring that users adhere to usage guidelines or restrictions to access the model or implementing safety filters. 
        \item Datasets that have been scraped from the Internet could pose safety risks. The authors should describe how they avoided releasing unsafe images.
        \item We recognize that providing effective safeguards is challenging, and many papers do not require this, but we encourage authors to take this into account and make a best faith effort.
    \end{itemize}

\item {\bf Licenses for existing assets}
    \item[] Question: Are the creators or original owners of assets (e.g., code, data, models), used in the paper, properly credited and are the license and terms of use explicitly mentioned and properly respected?
    \item[] Answer: \answerYes{} % Replace by \answerYes{}, \answerNo{}, or \answerNA{}.
    \item[] Justification:  Upstream data sources are credited and their licenses respected: Wikipedia (CC BY-SA 3.0+) and Wikidata (CC0 1.0); both are referenced via \url{https://dumps.wikimedia.org/} (Section~\ref{sec:data-collection}) and recorded in the Datasheet (Appendix~\ref{sec:appendix:dataset-sheet:composition}) and the Croissant + RAI metadata. Our derived release of \datasetname{} is licensed CC BY-SA 4.0, which is compatible with the upstream Wikipedia ShareAlike requirement. Benchmarked baselines are properly cited with their original publications and repository URLs: REBEL~\citep{cabot2021rebel}, ReLiK~\citep{orlando2024relik}, EDC~\citep{zhang2024extract}, KGGen~\citep{mo2025kggen}, RAKG~\citep{zhang2025rakg}; metric implementations are likewise cited (Completeness from \cite{jiang2024genres}; G-BERTScore from \cite{saha2021explagraphs} extending BERTScore~\citep{zhang2020bertscore}). LLMs used are cited and accessed under their respective terms: Llama-3.1-405B-Instruct-GPTQ-INT4 (Llama 3.1 Community License), Mistral-Large-2411 and Mistral-small (Mistral AI usage terms), GPT-5.1 (OpenAI usage policy). 
    % Our own released code is licensed Apache 2.0 (\texttt{LICENSE}); our released data is licensed CC BY-SA 4.0 (\texttt{LICENSE-DATA}).
    \item[] Guidelines:
    \begin{itemize}
        \item The answer \answerNA{} means that the paper does not use existing assets.
        \item The authors should cite the original paper that produced the code package or dataset.
        \item The authors should state which version of the asset is used and, if possible, include a URL.
        \item The name of the license (e.g., CC-BY 4.0) should be included for each asset.
        \item For scraped data from a particular source (e.g., website), the copyright and terms of service of that source should be provided.
        \item If assets are released, the license, copyright information, and terms of use in the package should be provided. For popular datasets, \url{paperswithcode.com/datasets} has curated licenses for some datasets. Their licensing guide can help determine the license of a dataset.
        \item For existing datasets that are re-packaged, both the original license and the license of the derived asset (if it has changed) should be provided.
        \item If this information is not available online, the authors are encouraged to reach out to the asset's creators.
    \end{itemize}

\item {\bf New assets}
    \item[] Question: Are new assets introduced in the paper well documented and is the documentation provided alongside the assets?
    \item[] Answer: \answerYes{} % Replace by \answerYes{}, \answerNo{}, or \answerNA{}.
    \item[] Justification: The new asset (\datasetname{} dataset) is documented along multiple complementary surfaces: (i) a full Datasheet (Appendix~\ref{sec:appendix:dataset-sheet:composition}) following \cite{gebru2021datasheets}, covering motivation, composition, collection process, recommended uses, distribution, and maintenance; (ii) a validated Croissant 1.1 + Responsible AI metadata file shipped alongside the Hugging Face dataset page; (iii) the Hugging Face dataset README and the GitHub top-level \texttt{README.md}; (iv) per-step documentation in the \texttt{README.md} files in the public repository (\url{https://github.com/klimzaporojets/emerge}). % \justificationTODO{}
    \item[] Guidelines:
    \begin{itemize}
        \item The answer \answerNA{} means that the paper does not release new assets.
        \item Researchers should communicate the details of the dataset\slash code\slash model as part of their submissions via structured templates. This includes details about training, license, limitations, etc. 
        \item The paper should discuss whether and how consent was obtained from people whose asset is used.
        \item At submission time, remember to anonymize your assets (if applicable). You can either create an anonymized URL or include an anonymized zip file.
    \end{itemize}

\item {\bf Crowdsourcing and research with human subjects}
    \item[] Question: For crowdsourcing experiments and research with human subjects, does the paper include the full text of instructions given to participants and screenshots, if applicable, as well as details about compensation (if any)? 
    \item[] Answer: \answerNA{} % \answerTODO{} % Replace by \answerYes{}, \answerNo{}, or \answerNA{}.
    \item[] Justification: No external crowdsourcing was used. The annotation reliability check (500 triple-text pairs, 100 per TKGU operation) was performed by two expert co-author annotators (researchers holding a PhD in computer science) on public Wikipedia/Wikidata-derived content, with no compensation arrangements or third-party participants. The full annotator instructions are provided in Appendix~\ref{sec:appendix:human-annotation-guidelines}; the annotation process and inter-annotator agreement statistics are in Appendices~\ref{sec:appendix:human-annotation} and~\ref{sec:appendix:human-annotation-agreement}.
    \item[] Guidelines:
    \begin{itemize}
        \item The answer \answerNA{} means that the paper does not involve crowdsourcing nor research with human subjects.
        \item Including this information in the supplemental material is fine, but if the main contribution of the paper involves human subjects, then as much detail as possible should be included in the main paper. 
        \item According to the NeurIPS Code of Ethics, workers involved in data collection, curation, or other labor should be paid at least the minimum wage in the country of the data collector. 
    \end{itemize}

\item {\bf Institutional review board (IRB) approvals or equivalent for research with human subjects}
    \item[] Question: Does the paper describe potential risks incurred by study participants, whether such risks were disclosed to the subjects, and whether Institutional Review Board (IRB) approvals (or an equivalent approval/review based on the requirements of your country or institution) were obtained?
    \item[] Answer: \answerNA{} % Replace by \answerYes{}, \answerNo{}, or \answerNA{}.
    \item[] Justification: The work does not involve human-subjects research in the IRB sense: no external participants, no personal-data collection, no behavioral or interventional study. The only ``annotators'' are two expert co-authors performing label-quality verification on publicly available Wikipedia/Wikidata-derived content (Appendix~\ref{sec:appendix:human-annotation}). There are no risks to study participants and no IRB approval is required.
    \item[] Guidelines:
    \begin{itemize}
        \item The answer \answerNA{} means that the paper does not involve crowdsourcing nor research with human subjects.
        \item Depending on the country in which research is conducted, IRB approval (or equivalent) may be required for any human subjects research. If you obtained IRB approval, you should clearly state this in the paper. 
        \item We recognize that the procedures for this may vary significantly between institutions and locations, and we expect authors to adhere to the NeurIPS Code of Ethics and the guidelines for their institution. 
        \item For initial submissions, do not include any information that would break anonymity (if applicable), such as the institution conducting the review.
    \end{itemize}

\item {\bf Declaration of LLM usage}
    \item[] Question: Does the paper describe the usage of LLMs if it is an important, original, or non-standard component of the core methods in this research? Note that if the LLM is used only for writing, editing, or formatting purposes and does \emph{not} impact the core methodology, scientific rigor, or originality of the research, declaration is not required.
    %this research? 
    \item[] Answer: \answerYes{} % \answerTODO{} % Replace by \answerYes{}, \answerNo{}, or \answerNA{}.
    \item[] Justification: LLMs are central to two parts of the work and are declared in detail. (i) \textbf{Dataset curation}: \texttt{Meta-Llama-3.1-405B} is used to validate that each TKGU operation can be derived from the corresponding textual passage during the curation step (Section~\ref{sec:quality-control}); the full prompts, examples, and per-operation triggers are reproduced in Appendix~\ref{sec:appendix:prompts}, and curation reliability against two human expert annotators is reported in Appendices~\ref{sec:appendix:human-annotation}--\ref{sec:appendix:human-annotation-agreement}. (ii) \textbf{Benchmark evaluation}: LLM-based IE methods (EDC+ as an extension of EDC, KGGen, RAKG) are evaluated with three back-end LLMs (\texttt{Mistral-small}, \texttt{Mistral-Large-2411}, \texttt{GPT-5.1}); the model list, configurations, and our EDC+ extension are described in Section~\ref{sec:models} and Appendix~\ref{sec:appendix:compared-models}, with the EDC+ prompts in Appendix~\ref{sec:appendix:edc-prompts}, and the in-context-vs-zero-shot ablation in Appendix~\ref{sec:appendix:icl-ablation}.
    \item[] Guidelines:
    \begin{itemize}
        \item The answer \answerNA{} means that the core method development in this research does not involve LLMs as any important, original, or non-standard components.
        \item Please refer to our LLM policy in the NeurIPS handbook for what should or should not be described.
    \end{itemize}

\end{enumerate}
    %%%%%%%%%%%%%%%%%%%%%%%%%%%%%%%%%%%%%%%%%%%%%%%%%%%%%%%%%%%%
    \newpage
    \appendix
    % \section*{Appendix}
    \section{Comparison of \datasetname~with Existing Information Extraction Benchmarks}
    \label{sec:appendix:comparison}
    \Tabref{tab:comparison-benchmarks} presents a detailed comparison of \datasetname~with existing information extraction (IE) benchmark datasets across the following key criteria:
    \begin{itemize}
        \item \textbf{Evolution:} indicates whether the dataset captures the natural evolution of knowledge in knowledge graph (\textit{KG}) and textual (\textit{Text}) sources.
        \item \textbf{Text-to-KG integration:} extent to which information extraction annotations are integrated with knowledge in a KG, broken down in:
        \begin{itemize}
            \item \textsc{KG Linked:} indicates whether the annotated entities in the triples are linked to a KG, supporting thus \textit{entity linking} task.
            \item \textsc{\opexists:} presence of triples aligned with facts already in a KG (\opexists ~TKGU operation; Section \ref{sec:problem-definition}).
            \item \textsc{\opadd:} whether a dataset can be used to extract triples from text that connect existing entities in a KG (\opadd~TKGU operation; Section~\ref{sec:problem-definition}).
            \item \textsc{\opmintadd:}
            coverage of triples involving emerging (non-existing) entities in a KG (\opmintadd~TKGU operation; Section~\ref{sec:problem-definition}).
            \item \textsc{\opinfer:} availability of annotations linking emerging entities in text to other entities in KG not mentioned in text (\opinfer~TKGU operation; Section~\ref{sec:problem-definition}).
            \item \textsc{\opdeprecate:} inclusion of annotations that mark deprecation of existing KG triples based on information in textual passage (\opdeprecate~TKGU operation; Section~\ref{sec:problem-definition}).
        \end{itemize}
    \end{itemize}
    From \Tabref{tab:comparison-benchmarks}, we observe that, to the best of our knowledge, none of the existing IE datasets support information extraction in a realistic knowledge evolution setting, where knowledge evolves simultaneously in both KG and textual sources (columns \textsc{Evolution-KG} and \textsc{Evolution-Text} in the table).
    Moreover, a number of datasets, such as TACRED \citep{zhang2017position}, BC5CDR \citep{li2016biocreative}, DDI \citep{herrero2013ddi}, and DWIE \citep{zaporojets2021dwie}, include \textit{entity linking} to a KG, but are not accompanied by an actual KG, and their extracted relations do not align directly with the relations defined in a KG schema.
    Finally, although \opadd~and \opmintadd~operations are nominally supported in some of the compared datasets, they do not capture genuinely emerging knowledge; instead, they rely on random subsampling of triples to approximate TKGU operations.
    \newpage
    \begin{table}
    \caption{Overview of major
    information extraction datasets from the past three decades across various domains, compared to our~\datasetname~dataset.
    }
    \label{tab:comparison-benchmarks}
    \rowcolors{5}{rowwhite}{rowgray} % start from row 2
    \centering
    \resizebox{\textwidth}{!}{\begin{tabular}{l|cc|cccccc}
    \toprule
    & \multicolumn{2}{c|}{Evolution} & \multicolumn{6}{c}{Text-to-KG integration} \\
    \cmidrule(lr){2-3}\cmidrule(lr){4-9}
    Dataset & KG & Text  & KG &  \opexists  &  \opadd & \opmintadd & \opinfer & \opdeprecate \\
            &        &   & Linked &  &  &  &   &  \\ \midrule
    % FB15K (\citeyear{bordes2013translating}) & \xmark & \xmark & \xmark & \cmark & \revklimtodo{?} & \revklimtodo{?} & \revklimtodo{?} \\
    MUC-7 \citep{chinchor1998muc} & \xmark & \xmark & \xmark & \xmark & \xmark & \xmark & \xmark & \xmark \\
    % ontonotes does not have entity linking
    % OntoNotes \revklimtodo{5.0} (\revklimtodo{TODO}) & \revklimtodo{?} & \revklimtodo{?} & \revklimtodo{?} & \revklimtodo{?} & \revklimtodo{?} & \revklimtodo{?} & \revklimtodo{?} \\
    CoNLL04 \citep{roth2004linear} & \xmark & \xmark & \xmark & \xmark & \xmark & \xmark & \xmark & \xmark \\
    % CoNLL-2003 (\revklimtodo{TODO}) & \revklimtodo{?} & \revklimtodo{?} & \revklimtodo{?} & \revklimtodo{?} & \revklimtodo{?} & \revklimtodo{?} & \revklimtodo{?} \\
    ACE 2005 \citep{walker2006ace} & \xmark & \xmark & \xmark & \xmark & \xmark & \xmark & \xmark & \xmark \\
    SemEval 2010 \citep{hendrickx2010semeval} & \xmark & \xmark & \xmark & \xmark & \xmark & \xmark & \xmark & \xmark \\
    NYT \citep{riedel2010modeling} & \xmark & \xmark & \cmark & \cmark & \cmark & \xmark & \xmark & \xmark \\
    ADE \citep{gurulingappa2012development} & \xmark & \xmark & \xmark & \xmark & \xmark & \xmark & \xmark & \xmark \\
    % FB15K (\citeyear{bordes2013translating}) & \revklimtodo{?} & \revklimtodo{?} & \revklimtodo{?} & \revklimtodo{?} & \revklimtodo{?} & \revklimtodo{?} & \revklimtodo{?} \\
    % WN18 (\citeyear{bordes2013translating}) & \revklimtodo{?} & \revklimtodo{?} & \revklimtodo{?} & \revklimtodo{?} & \revklimtodo{?} & \revklimtodo{?} & \revklimtodo{?} \\
    % in DDI the relations exist between entities from KG, yet they are disconnected from what is there in KGs, no concrete KG is used to update. Just extract without accounting for what is already in a KG.
    DDI \citep{herrero2013ddi}  & \xmark & \xmark & \cmark & \xmark & \xmark & \xmark & \xmark & \xmark \\
    % FB15K-237 (\citeyear{toutanova2015observed}) & \revklimtodo{?} & \revklimtodo{?} & \revklimtodo{?} & \revklimtodo{?} & \revklimtodo{?} & \revklimtodo{?} & \revklimtodo{?}  \\
    BC5CDR \citep{li2016biocreative} & \xmark & \xmark & \cmark & \xmark & \xmark & \xmark & \xmark & \xmark  \\
    % Wikireading can be excluded as it is not a typical relationn extraction dataset, more like a value extraction, also not sure if we should mark KG link as it is based on values.
    WikiReading \citep{hewlett2016wikireading} & \xmark & \xmark & \xmark & \xmark & \xmark & \xmark & \xmark & \xmark  \\
    ScienceIE \citep{augenstein2017semeval} & \xmark & \xmark & \xmark & \xmark & \xmark & \xmark & \xmark & \xmark \\
    WebNLG \citep{gardent2017webnlg} & \xmark & \xmark & \cmark & \cmark & \cmark & \xmark & \xmark & \xmark \\
    WNUT \citep{derczynski2017results} & \xmark & \xmark & \xmark & \xmark & \xmark & \xmark & \xmark & \xmark \\
    % WebNLG (\citeyear{gardent2017webnlg}) & \revklimtodo{?} & \revklimtodo{?} & \revklimtodo{?} & \revklimtodo{?} & \revklimtodo{?} & \revklimtodo{?} & \revklimtodo{?} \\
    TAC KBP \citep{getman2017overview} & \xmark & \xmark & \cmark & \cmark & \cmark & \cmark & \xmark & \xmark \\
    SciERC \citep{luan2018multi} & \xmark & \xmark & \xmark & \xmark & \xmark & \xmark & \xmark & \xmark \\
    % WN18RR (\citeyear{dettmers2018convolutional}) & \revklimtodo{?} & \revklimtodo{?} & \revklimtodo{?} & \revklimtodo{?} & \revklimtodo{?} & \revklimtodo{?} & \revklimtodo{?} & \revklimtodo{?} \\
    TACRED \citep{zhang2017position} & \xmark & \xmark & \cmark & \xmark & \xmark & \xmark & \xmark & \xmark \\
    FewRel \citep{han2018fewrel} & \xmark & \xmark & \cmark & \cmark & \cmark & \xmark & \xmark & \xmark \\
    FewRel 2.0 \citep{gao2019fewrel} & \xmark & \xmark & \cmark & \cmark & \cmark & \xmark & \xmark & \xmark \\
    Geo-NRE \citep{distiawan2019neural} & \xmark & \xmark & \cmark & \cmark & \cmark & \xmark & \xmark & \xmark \\
    Wiki-NRE \citep{distiawan2019neural} & \xmark & \xmark & \cmark & \cmark & \cmark & \xmark & \xmark & \xmark \\
    T-REX \citep{elsahar2019t} & \xmark & \xmark & \cmark & \cmark & \cmark & \xmark & \xmark & \xmark \\
    DocRED \citep{yao2019docred} & \xmark & \xmark & \cmark & \cmark & \cmark & \xmark & \xmark & \xmark \\
    Wiki80 \citep{han2019opennre} & \xmark & \xmark & \cmark & \cmark & \cmark & \xmark & \xmark & \xmark \\
    FOBIE \citep{kruiper2020layman} & \xmark & \xmark & \xmark & \xmark & \xmark & \xmark & \xmark & \xmark \\
    DialogueRE \citep{yu2020dialogue} & \xmark & \xmark & \xmark & \xmark & \xmark & \xmark & \xmark & \xmark \\
    BioRel \citep{xing2020biorel} & \xmark & \xmark & \cmark & \cmark & \cmark & \xmark & \xmark & \xmark \\
    Wiki20 \citep{han2020more} & \xmark & \xmark & \cmark & \cmark & \cmark & \xmark & \xmark & \xmark \\
    DWIE \citep{zaporojets2021dwie} & \xmark & \xmark & \cmark & \xmark & \xmark & \xmark & \xmark & \xmark \\
    KELM \citep{agarwal2021knowledge} & \xmark & \xmark & \cmark & \cmark &  \cmark & \xmark & \xmark & \xmark \\
    REBEL \citep{cabot2021rebel} & \xmark & \xmark & \cmark & \cmark & \cmark & \xmark & \xmark & \xmark \\
    Re-TACRED \citep{stoica2021re} & \xmark & \xmark & \cmark & \xmark & \xmark & \xmark & \xmark & \xmark \\
    SMiLER \citep{seganti2021multilingual} & \xmark & \xmark & \xmark & \xmark & \xmark & \xmark & \xmark & \xmark \\
    DrugProt \citep{miranda2021overview} & \xmark & \xmark & \cmark & \xmark & \xmark & \xmark & \xmark & \xmark \\
    mLAMA \citep{kassner2021multilingual} & \xmark & \xmark & \xmark & \xmark & \xmark & \xmark & \xmark & \xmark \\
    Re-DocRED \citep{tan2022revisiting} & \xmark & \xmark & \cmark & \cmark & \cmark & \xmark & \xmark & \xmark \\
    CDG \citep{zhang2022distant} & \xmark & \xmark & \cmark & \cmark & \cmark & \xmark & \xmark & \xmark \\
    \textsc{KD-DTI} \citep{hou2022discovering} & \xmark & \xmark & \cmark & \cmark & \cmark & \xmark & \xmark & \xmark \\
    FinRED \citep{sharma2022finred} & \xmark & \xmark & \cmark & \cmark & \cmark & \xmark & \xmark & \xmark \\
    BioRED \citep{luo2022biored} & \xmark & \xmark & \cmark & \xmark & \xmark & \xmark & \xmark & \xmark \\
    SynthIE-text \citep{josifoski2023exploiting}  & \xmark & \xmark & \cmark & \cmark & \cmark & \xmark & \xmark & \xmark \\
    REFinD \citep{kaur2023refind}  & \xmark & \xmark & \xmark & \xmark & \xmark & \xmark & \xmark & \xmark \\
    BioDEX \citep{d2023biodex} & \xmark & \xmark & \cmark & \xmark & \xmark & \xmark & \xmark & \xmark \\
    % WebNLG \citep{gardent2017webnlg}, KELM \citep{agarwal2021knowledge}, FewRel \citep{han2018fewrel}, DocRED \citep{yao2019docred}, Wiki/GEO-NRE \citep{distiawan2019neural}, BioRel \citep{xing2020biorel}, T-REX \citep{elsahar2019t} and REBEL \citep{cabot2021rebel}.
    % \textsc{Text2KGBench}
    \textsc{Text2KG} \citep{mihindukulasooriya2023text2kgbench} & \xmark & \xmark & \cmark & \cmark & \cmark & \xmark & \xmark & \xmark \\
    \textsc{SciER} \citep{zhang2024scier} & \xmark & \xmark & \xmark & \xmark & \xmark & \xmark & \xmark & \xmark \\
    \textsc{SciNLP} \citep{duan2025scinlp} & \xmark & \xmark & \cmark & \cmark & \cmark & \xmark & \xmark & \xmark \\

    \midrule
    \datasetname~(ours) & \cmark & \cmark & \cmark & \cmark & \cmark & \cmark & \cmark & \cmark \\

    \bottomrule
    \end{tabular}}
    \end{table}

    \section{Metrics}
    \label{sec:appendix:metrics}

    \subsection{\mprexecutable{} Evaluation Protocol}
    \label{sec:appendix:executable-protocol}

    The \textit{\prexecutable{}} protocol formalizes the requirement that a predicted update can be applied as edit operation to an external KG. Let $G_t = (V_t, R_t, T_t)$ denote the KG snapshot at time $t$ and $G_{t'}$ the snapshot at $t' > t$. A predicted triple $(s, p, o)$ is \emph{\prexecutable{}} iff each component is mapped to a canonical KG identifier and the implied operation can be applied to $G_t$ to yield the correct change toward $G_{t'}$.

Per-operation requirements differ in whether canonical-ID evaluation is possible at inference time (\Tabref{tab:exec-protocol-per-op}). For \opexists{}, \opadd{}, and \opdeprecate{}, this evaluation is well-defined. In \opexists{} and \opadd{}, both entities already exist in the current KG snapshot $V_t$, so predictions can be matched directly against canonical entity identifiers. In \opdeprecate{}, the model must identify a triple that is present in the current snapshot but marked as deprecated in the later snapshot, so exact matching against the corresponding canonical triple is also possible. These operations can therefore be evaluated directly with~\revklim{Executable-R (E-R; Section~\ref{sec:metrics-and-evaluation}); for \opdeprecate{}, E-R covers the 91.3\% of test triples present in $T_t$ (the others have a minted entity or entered the KG only after $t$).}

In contrast, \opmintadd{} and \opinfer{} involve at least one emerging entity that has no canonical identifier in $V_t$ at inference time. Canonical-ID exact match is therefore not directly available for these operations. It can only be computed \emph{post-hoc}, after the emerging entity appears in the later snapshot $V_{t'}$. One feasible approach is alignment-based canonicalization: predicted text-form triples are aligned to canonical gold triples, for example via Hungarian matching over a BERTScore similarity matrix, as in G-BERTScore-R (Section~\ref{sec:metrics-and-evaluation}). Each prediction can then inherit the canonical identifiers of its aligned gold triple, allowing standard E-R to be computed after alignment. The alignment score is also meaningful on its own: low similarity indicates weak extraction or poor entity-label generation, whereas high similarity indicates a faithful emerging-entity prediction. Thus, for \opmintadd{} and \opinfer{}, G-BERTScore-R provides a continuous measure of prediction quality, while thresholded alignment-based canonicalization gives a complementary discrete E-R-style measure.

    \begin{table}
    \caption{Per-operation requirements and feasibility under the \prexecutable{} protocol. Inference-time exact match is feasible when all entities involved have canonical identifiers in $V_t$ (\opexists, \opadd, \opdeprecate). Post-hoc match resolves the emerging entity to its identifier in $V_{t'}$ once it has been added to the KG (\opmintadd, \opinfer).}
    \label{tab:exec-protocol-per-op}
    \centering
    \small
    \begin{tabularx}{\linewidth}{l l X l}
    \toprule
    Operation & Exact-match feasibility & Additional capability required & Supported by \\
    \midrule
    \opexists    & inference-time & link both entities to $V_t$ & ReLiK cIE\revklim{, KGA} \\
    \opadd       & inference-time & link both entities to $V_t$ & ReLiK cIE\revklim{, KGA} \\
    \opmintadd   & post-hoc only  & generate and add emerging entity to $V_{t}$ & none \\
    \opinfer     & post-hoc only  & generate and resolve emerging entity; KG-side reasoning & none \\
    \opdeprecate & inference-time & identify deprecated $(s,p,o) \in T_t$
    % $(s,p,o) \in T_t \setminus T_{t'}$
    & \revklim{KGA} \\
    \bottomrule
    \end{tabularx}
    \end{table}

 % \revklim{We apply two uniform scoring refinements, both applied identically to all methods and leaving the qualitative findings unchanged: (i) instances with no assessor-validated gold triples of an operation are excluded from that operation's Executable-R, matching the per-triple metrics (Completeness, G-BERTScore-R), which already exclude them; and (ii) \opdeprecate{} gold triples absent from the current snapshot $T_t$ --- violating the \opdeprecate{} precondition $(s,p,o)\in T_t$ (emerging entities created after the snapshot, or edges added and removed within the same window) --- are removed, reducing the \opdeprecate{} evaluation set from 1628 to 1484 gold triples.}

    \subsection{Completeness}
    \label{sec:appendix:completeness}
    The completeness metric \citep{jiang2024genres} can be formalized as follows:
    \begin{equation*}
      c(\mathcal{T}'_{\mathcal{D}}, \mathcal{T}_{\mathcal{D}}) = \dfrac{\vert \{ \tau \in \mathcal{T}'_{\mathcal{D}}  | \exists\tau \in \mathcal{T}_{\mathcal{D}}, \text{sim}(\tau, \tau') \geq \phi \} \vert}{\vert \mathcal{T}'_{\mathcal{D}} \vert},
    \end{equation*}
    where $\mathcal{T}'_{\mathcal{D}}$ is the set of ground truth, and $\mathcal{T}_{\mathcal{D}}$ the set of predicted triples. $\text{sim}(\tau, \tau') = \text{CosSim}(emb(\tau),emb(\tau'))$. We use \texttt{SentenceTransformer(`all-mpnet-base-v2')} to calculate the embeddings $emb$.
    We set the threshold $\phi$ to $0.9$, which, based on our observations, provides accurate similarity matching.

    \subsection{Bootstrap Stability of Benchmark Results}
    \label{sec:appendix:bootstrap-stability}

    We assess the stability of the results reported in \Tabref{tab:dataset_re_results} via bootstrap resampling. From the 3{,}500-instance test set we draw $B = 100$ bootstrap samples with replacement; on each sample we re-aggregate Completeness and G-BERTScore-R for every (model, TKGU operation) entry, yielding a 100-value distribution.
    % per cell.
    \Tabref{tab:bootstrap-stability} reports, per (metric, operation), the median and maximum width of the 95\% confidence interval across the eleven benchmarked models, and the Spearman rank correlation $\rho$ between the full-dataset ranking and each bootstrap-sample ranking (minimum and mean over the 100 samples). $\rho = 1.0$ indicates a perfectly preserved ranking; lower values indicate rank changes.

    % ------------------------------------------------------------------------------------------------------------------
    % Table 6: emerge-stage:scripts/stats/camera_ready_reissue_20261028/bootstrap_table_reissue_20261028.py --pkl <D3> --resample with-replacement
    % Output: bootstrap__with-replacement__table_rows.tex (these rows, copied verbatim); test: check_bootstrap.py (full means = Table 3)
    % Note: Table 3 aggregation (405B filter, 11 models, C > 0.90), real bootstrap; the old bootstrap_table.py did neither
    % ------------------------------------------------------------------------------------------------------------------
    \begin{table}
    \caption{Bootstrap stability of \Tabref{tab:dataset_re_results} ($B = 100$ resamples with replacement of the 3{,}500-instance test set). For each (metric, TKGU operation) we report the median and maximum 95\% CI width across the eleven models, and the Spearman rank correlation $\rho$ (minimum and mean over the 100 samples) between the full-dataset and bootstrap-sample rankings.}
    \label{tab:bootstrap-stability}
    \centering
    \small
    \begin{tabular}{ll rr rr}
    \toprule
    Metric & TKGU operation & \multicolumn{2}{c}{95\% CI width} & \multicolumn{2}{c}{Rank $\rho$} \\
    \cmidrule(lr){3-4}\cmidrule(lr){5-6}
     & & median & max & min & mean \\
    \midrule
    \textsc{C}   & \opexists    & 2.9 & 3.2 & 0.87 & 0.97 \\
    \textsc{C}   & \opadd       & 4.3 & 5.2 & 0.87 & 0.96 \\
    \textsc{C}   & \opmintadd   & 3.3 & 4.6 & 0.90 & 0.97 \\
    \textsc{C}   & \opinfer     & 1.3 & 3.4 & 0.89 & 0.98 \\
    \textsc{C}   & \opdeprecate & 0.0 & 5.8 & 0.99 & 1.00 \\
    \midrule
    \textsc{G-R} & \opexists    & 1.5 & 2.4 & 0.99 & 1.00 \\
    \textsc{G-R} & \opadd       & 1.5 & 2.9 & 0.96 & 1.00 \\
    \textsc{G-R} & \opmintadd   & 1.3 & 3.1 & 0.96 & 1.00 \\
    \textsc{G-R} & \opinfer     & 1.2 & 2.3 & 0.99 & 1.00 \\
    \textsc{G-R} & \opdeprecate & 0.0 & 4.3 & 0.99 & 0.99 \\
    % \textsc{C}   & \opexists    & 1.5 & 3.1 & 0.92 & 0.98 \\
    % \textsc{C}   & \opadd       & 1.5 & 2.4 & 0.97 & 1.00 \\
    % \textsc{C}   & \opmintadd   & 1.7 & 2.5 & 0.98 & 1.00 \\
    % \textsc{C}   & \opinfer     & 1.9 & 3.0 & 0.96 & 0.99 \\
    % \textsc{C}   & \opdeprecate & 0.0 & 4.2 & 0.99 & 1.00 \\
    % \midrule
    % \textsc{G-R} & \opexists    & 1.3 & 2.0 & 0.99 & 1.00 \\
    % \textsc{G-R} & \opadd       & 1.2 & 2.3 & 0.96 & 0.99 \\
    % \textsc{G-R} & \opmintadd   & 1.1 & 2.1 & 0.97 & 1.00 \\
    % \textsc{G-R} & \opinfer     & 1.1 & 1.7 & 0.99 & 1.00 \\
    % \textsc{G-R} & \opdeprecate & 0.0 & 3.2 & 0.99 & 0.99 \\
    \bottomrule
    \end{tabular}
    \end{table}

    % Rankings are near-perfectly preserved across bootstrap resamples (Spearman $\rho \geq 0.92$, mean $\geq 0.98$), and 95\% CI widths are narrow (median $\leq 1.9$ pts), confirming that the conclusions drawn from \Tabref{tab:dataset_re_results} are robust to test-set resampling.
    \revklim{Rankings are preserved across bootstrap resamples (Spearman $\rho \geq 0.87$, mean $\geq 0.96$), and 95\% CI widths are narrow (median $\leq 4.3$ points), confirming that the conclusions drawn from \Tabref{tab:dataset_re_results} are robust to test-set resampling.}

    \section{Benchmarked Models}
    \label{sec:appendix:compared-models}
    This appendix provides supplementary descriptions of the information extraction (IE) models included in our benchmark. The model descriptions are given below.

    \noindent \textbf{ReLiK.}
    ReLiK \citep{orlando2024relik} is a highly scalable architecture designed to minimize resource usage while achieving state-of-the-art performance in both entity linking and relation extraction. In our study, we evaluate two variants of ReLiK: closed information extraction ReLiK (ReLiK cIE) and relation-extraction ReLiK (ReLiK RE). \textit{ReLiK cIE} operates under the closed IE assumption \citep{galarraga2014canonicalizing, chaganty2017importance, josifoski2023exploiting}, predicting relations only between entities already present in the KG. Consequently, it can handle only those TKGU operations involving known entities, namely, \opexists~and \opadd~as defined in Section~\ref{sec:problem-definition}. For each test snapshot $t$, both models are provided with the corresponding KG snapshot. Specifically, ReLiK cIE receives the dictionaries of entities ($V_t$) and relation types ($R_t$) present in $t$, while ReLiK RE is given only the relation types ($R_t$), as it predicts relations without linking extracted entity mentions.
    To generate predictions, we run ReLiK on each KG snapshot independently. In each run, ReLiK is provided with the dictionary of entities and relations specific to that snapshot. For relation encoding, we use the pre-trained ReLiK model available on Hugging Face:
    \texttt{relik-ie/encoder-e5-small-v2-wikipedia- relations}.
    These relation encodings are used by both ReLiK RE and ReLiK cIE. For each snapshot, we also encode the corresponding KG entities using the model
    \texttt{relik-ie/encoder-e5-small-v2-wikipedia- matryoshka}.
    For prediction, we use the pre-trained \texttt{relik-ie/relik-relation-extraction-large} model for ReLiK RE, and the pre-trained \texttt{relik-ie/relik-cie-large} model for ReLiK cIE.

    \noindent \textbf{EDC.} The \textit{extract, define, canonicalize (EDC)} framework, introduced by \cite{zhang2024extract}, is a state-of-the-art LLM-based approach that uses LLMs to canonicalize relation types.~In its original implementation it extracts only triples explicitly mentioned in the input text, and thus covers only the triples under \opexists, \opadd, and \opmintadd~TKGU operations.~To extend it to \opinfer~and \opdeprecate~TKGU operations, we modify the original EDC prompt to (i) extract triples that may involve entities not explicitly mentioned in the input text but present in a Wikidata snapshot, and (ii) propose candidate triples for deprecation from the KG. We term this adaptation \textbf{EDC+} in our experiments. Additional execution details and the prompts used are provided in Section~\ref{sec:appendix:edc-prompts}.

    \noindent \textbf{KGGen.} We evaluate KGGen \citep{mo2025kggen}, an LLM-based text-to-KG generation method, using the default configuration provided in the official KGGen repository,\footnote{\url{https://github.com/stair-lab/kg-gen}} including the recommended prompts and hyperparameters.

    \noindent \textbf{RAKG.} We evaluate RAKG \citep{zhang2025rakg}, which constructs document-level knowledge graphs via pre-entity extraction followed by retrieval-augmented refinement to incorporate global context from text chunks.~We use the configuration provided in the official RAKG repository.\footnote{\url{https://github.com/KnowledgeXLab/RAKG}}

    \noindent \textbf{REBEL.} We evaluate REBEL \citep{cabot2021rebel}, an autoregressive seq2seq relation extraction model that casts extraction as conditional generation by producing a linearized sequence of (head, relation, tail) triples from the input text. We use the default configuration provided in the official REBEL repository.\footnote{\url{https://huggingface.co/Babelscape/rebel-large}}

    \noindent\revklim{\textbf{KG-aware (KGA).} We use \texttt{gpt-5.1-2025-11-13} with operation-specific instructions and five fixed demonstrations drawn from text-supported operations outside the test set, requesting JSON predictions. For retrieval, we use the ReLiK \texttt{relik-cie-large} retriever with snapshot-specific entity and relation indices. For each passage, we retrieve the top 124 entities and relation types. Candidate triples are edges in $G_t$ connecting these entities and the entity associated with the passage's Wikipedia page via the retrieved relations, ranked by the sum of head, relation, and tail retrieval scores. KGA-RAG includes the top 32 candidate triples for \opexists{} and \opdeprecate{}, and the top 32 entities, relation types, and triples for the remaining operations. KGA-O draws its gold entities, relation types, and triples exclusively from text-supported operations. Since an instance may contain multiple TKGU operation types, this context can also include gold triples from other operations in the same instance, such as \opexists{} and \opdeprecate{}. For \opinfer{}, KGA-O additionally includes up to six $G_t$ pattern triples per gold \opinfer{} triple. These triples show how the gold relation links other entities to the same existing entity, guiding the model toward analogous \opinfer{} triples for the emerging entity. KGA-O+RAG combines gold and the top 32 retrieved triples for \opexists{} and \opdeprecate{}. For the remaining operations, it supplements gold entities and relation types with retrieved candidates (up to 32 each), retaining the oracle triples and \opinfer{} pattern triples. The combined context is randomly shuffled. The reported predictions were generated using KG context built before the v1.0.1 re-issue.}

    \noindent\revklim{\textbf{Reproducibility.} We release the prompts, model configurations, and evaluation code. The locally run baselines (REBEL and ReLiK) can be reproduced using the released configurations, whereas results from API-hosted models (\texttt{Mistral-small}, \texttt{Mistral-Large-2411}, and \texttt{GPT-5.1}) may vary due to nondeterminism and provider updates.}

    \section{Quality Control}
    In this section, we describe 
    \revklim{how LLMs are used to automatically flag triples that cannot be derived from textual passages}
    % how LLMs are used to automatically filter out triples that cannot be derived from textual passages 
    (Section \ref{sec:appendix:prompts}). We also detail the human annotation process used to validate the resulting LLM-generated annotations (Section \ref{sec:appendix:human-annotation}).

    \subsection{Quality Control Prompts and Examples}
    \label{sec:appendix:prompts}
    % We use two different prompts to filter out triples that cannot be inferred from a textual passage. 
    \revklim{We use two different prompts to flag triples that cannot be inferred from a textual passage.}
    The first is an \textit{assertion prompt} (see Section~\ref{sec:prompt-assertion}) applied to validate \opexists, \opadd, \opmintadd, and \opinfer~TKGU operations as defined in Section~\ref{sec:problem-definition}. The goal of this prompt is to verify whether a triple can be directly or indirectly derived from the text. The second prompt is a \textit{deprecation prompt} (see Section~\ref{sec:prompt-deprecation}), and is used to validate the deprecation of triples involved in \opdeprecate~TKGU operation.

    \subsubsection{Triple Assertion Prompt}
    \label{sec:prompt-assertion}
    The following is the structure of the prompt used to assert that the \opexists, \opadd, \opmintadd, and \opinfer~TKGU operations can be derived from the information in textual passages. The placeholder \texttt{<TEXT>} is replaced by the textual passage, and \texttt{<TRIPLES\_LIST>} by a list of triples.
    \begin{promptbox}
    \ttfamily
    You are given the following text: \\

    \textit{\texttt{<TEXT>}} \\

    Can the following triples be directly or indirectly (the text provides some hints) inferred from the text? Use common sense but not knowledge that cannot be inferred from the text above. \\

    \textit{\texttt{<TRIPLES\_LIST>}} \\

    Write a numbered list with the triples above, where each of the triples is followed by YES if the triple is represented in the text, and NO otherwise. Follow your `YES' or `NO' answer for each triple with a brief explanation.
    \end{promptbox}

    The following is a concrete example of the \textit{triple assertion} prompt outlined above:

    \begin{promptbox}
    \ttfamily
    You are given the following text: \\

    \textit{After the war, Abraham H. Albertson returned to complete his education and graduated, with scholarship assistance, from the Columbia School of Architecture with a Ph.B in 1895.} \\

    Can the following triples be directly or indirectly (the text provides some hints) inferred from the text? Use common sense but not knowledge that cannot be inferred from the text above. \\

    \textit{1.  [Abraham H. Albertson(Q80520646); educated at(P69); Columbia Graduate School of Architecture, Planning and Preservation(Q5149701)] \\
    2.  [Abraham H. Albertson(Q80520646); occupation(P106); architect(Q42973)] \\
    3. [Abraham H. Albertson(Q80520646); country of citizenship(P27); United States of America(Q30)] \\
    4. [Abraham H. Albertson(Q80520646); instance of(P31); human(Q5)] \\
    5. [Abraham H. Albertson(Q80520646); archives at(P485); University of Washington Libraries(Q7896575)]
    }
    \\  \\
    Write a numbered list with the triples above, where each of the triples is followed by YES if the triple is represented in the text, and NO otherwise. Follow your `YES' or `NO' answer for each triple with a brief explanation.
    \end{promptbox}

    Below is the response generated by \texttt{Meta-Llama-3.1-405B} when prompted with the example above:
    \begin{promptbox}
    \ttfamily

    1.  [Abraham H. Albertson(Q80520646); educated at(P69); Columbia Graduate School of Architecture, Planning and Preservation(Q5149701)] - \explain{\textbf{YES} - \textit{The text explicitly states that Abraham H. Albertson graduated from the Columbia School of Architecture.}}\\
    2.  [Abraham H. Albertson(Q80520646); occupation(P106); architect(Q42973)] - \explain{\textbf{YES} - \textit{Although the text does not directly state that Abraham H. Albertson was an architect, it can be inferred from the fact that he graduated from the Columbia School of Architecture.}} \\
    3. [Abraham H. Albertson(Q80520646); country of citizenship(P27); United States of America(Q30)] - \explain{\textbf{NO} - \textit{The text does not provide any information about Abraham H. Albertson's country of citizenship.}} \\
    4. [Abraham H. Albertson(Q80520646); instance of(P31); human(Q5)] - \explain{\textbf{YES} - \textit{It can be inferred from the context that Abraham H. Albertson is a human being, as he is described as graduating from a school and having a war experience.}} \\
    5. [Abraham H. Albertson(Q80520646); archives at(P485); University of Washington Libraries(Q7896575)]  - \explain{\textbf{NO} - The text does not provide any information about the archives of Abraham H. Albertson.}

    \end{promptbox}

    \subsubsection{Prompt for Triple Deprecation}
    \label{sec:prompt-deprecation}
    The following prompt is used to verify whether the textual passage supports the deprecation of a set of triples from the KG (\opdeprecate~TKGU operation): \\

    \begin{promptbox}
    \ttfamily
    You are given the following text: \\

    \textit{\texttt{<TEXT>}} \\
    \\
    Does this text imply that the following triples are no longer valid or do not hold at the moment the text was written?
    \\

    \textit{\texttt{<TRIPLES\_LIST>}} \\
    \\
    Write a numbered list with the triples above, where each of the triples is followed by YES if the text implies that the triple is no longer valid or do not hold, and NO otherwise.~Follow your `YES' or `NO' answer for each of the triples with a brief explanation.

    \end{promptbox}

    The following is a concrete example of the \textit{triple deprecation} prompt outlined above:

    \begin{promptbox}
    \ttfamily
    You are given the following text: \\

    \textit{Kate Lehrer (born Kate Tom Staples; December 17, 1937) is an American writer, novelist and book reviewer from Washington, D.C., and a panelist on the Diane Rehm Book Club on National Public Radio. She was married to fellow writer and journalist Jim Lehrer from 1960 until his death in 2020.} \\
    \\
    Does this text imply that the following triples are no longer valid or do not hold at the moment the text was written?
    \\ \\
    \textit{1. [Kate Lehrer(Q6375599); spouse(P26); Jim Lehrer(Q931148)]} \\
    \textit{2. [Jim Lehrer(Q931148); spouse(P26); Kate Lehrer(Q6375599)]} \\
    \textit{3. [Jim Lehrer(Q931148); residence(P551); Washington, D.C.(Q61)]} \\
    \\
    Write a numbered list with the triples above, where each of the triples is followed by YES if the text implies that the triple is no longer valid or do not hold, and NO otherwise.~Follow your `YES' or `NO' answer for each of the triples with a brief explanation.

    \end{promptbox}

    Below is the response generated by \texttt{Meta-Llama-3.1-405B} when prompted with the example above:
    \begin{promptbox}
    \ttfamily

    1. [Kate Lehrer(Q6375599); spouse(P26); Jim Lehrer(Q931148)] - \explain{\textbf{YES} - The text states that Kate Lehrer was married to Jim Lehrer ``until his death in 2020'', implying that the marriage ended with his death, and therefore the spouse relationship is no longer valid.} \\
    2. [Jim Lehrer(Q931148); spouse(P26); Kate Lehrer(Q6375599)] - \explain{\textbf{YES} - Similar to the previous triple, the text implies that the marriage between Jim Lehrer and Kate Lehrer ended with his death, so the spouse relationship is no longer valid.} \\
    3. [Jim Lehrer(Q931148); residence(P551); Washington, D.C.(Q61)] - \explain{\textbf{NO} - The text does not provide any information about Jim Lehrer's residence, let alone imply that he no longer resides in Washington, D.C. It only mentions that Kate Lehrer is from Washington, D.C.} \\

    \end{promptbox}

    \subsection{Human Annotation Process}
    \label{sec:appendix:human-annotation}
    To evaluate the accuracy of automatic LLM annotations, we employ two expert annotators (researchers holding a PhD with a background in computer science) to manually label a subsample of 500 (100 triples per each TKGU type defined in Section~\ref{sec:problem-definition}) triples from 166 randomly selected passages from our~\datasetname~dataset. To ensure consistency and avoid ambiguity, the annotators follow a detailed set of guidelines described in Section~\ref{sec:appendix:human-annotation-guidelines}.

    \subsubsection{Annotation Guidelines}
    \label{sec:appendix:human-annotation-guidelines}
    \textit{Annotators were provided with the following guideline:}

    \begin{promptbox}
    For each annotation instance, you are provided with a textual passage, a triple, and an assessment type, which can be either \textit{assert} or \textit{deprecate}. For \textit{assert} assessments, respond YES if the triple can be directly or indirectly inferred from the passage, and NO if it is not supported by the textual knowledge.
    For \textit{deprecate} assessment, respond YES if the triple can be deprecated based on information present or implied in the passage, and NO otherwise. Take into account the following considerations when annotating for \textit{assert} assessment type:
    \begin{enumerate}
        \item The triple may not be factually correct at the time the text was written, but it expresses a fact that holds true at some other point in time. For example, the triple
        \textit{$\langle$Barack Obama, president of, United States$\rangle$}
        should be assessed YES for the text passage ``Barack Obama served as the 44th President of the United States from 2009 to 2017''.
        \item Use common world knowledge and reasoning to induce triples from textual passage. For example, the triple \textit{$\langle$Renault, headquarters in, France$\rangle$} should be assessed YES for the text passage ``The headquarters of Renault are located in Boulogne-Billancourt, a suburb of Paris.'', as Paris is located in France.
        \item Mark with NO any concrete fact that cannot be inferred from text, even if some of the entities appear in the passage. For example, the triple
    \textit{$\langle$John Smith, participant in, Portland Climate Action Group protest$\rangle$} should be assessed NO for the passage
    ``Several members of the Portland Climate Action Group gathered downtown to protest against deforestation and climate inaction.'', as its factuality cannot be reliably inferred from the text.
        \item Assess with NO the triples that cannot be reliably inferred from a textual passage. For example, the triple \textit{$\langle$David Bronkie, sibling, Eva Bronkie$\rangle$} should be assessed as NO for the passage: ``David Bronkie and Eva Bronkie co-founded a sustainable home goods business focused on eco-friendly candle kits.'', since the sibling relationship cannot be reliably inferred from the text (e.g., sharing the same last name).
    \end{enumerate}

    Take into account the following considerations when annotating for \textit{deprecate} assessment type:
    \begin{enumerate}
        \item The deprecation of a triple should be valid from the information provided in the passage and not the current status of the knowledge. For example, the triple \textit{$\langle$ Donald Trump, president of, United States $\rangle$} should be assessed with YES for the passage ``Joe Biden is the President of the United States, having taken office recently and begun his tenure with notable public appearances and speeches.'', despite the fact that Donald Trump may be a current president of United States.
        \item The deprecation of a triple might not be explicitly stated in the text, but can be implied. For example, the deprecation of the triple
    \textit{$\langle$Hans Rausing, spouse, Julia Rausing$\rangle$} should be assessed as YES for the passage
    ``Julia Rausing, the philanthropist and business heiress, passed away on April 18, 2024, at the age of 63 after a long battle with cancer. She is survived by her husband, Hans Rausing, and their family.'',
    since the marital relationship is no longer current due to Julia Rausing’s death, which implies that the triple is deprecated.
        \item Assess with NO any triples whose deprecation can not be reliably inferred from text, even if some of the entities appear in the text.
    \end{enumerate}
    \end{promptbox}

    % ------------------------------------------------------------------------------------------------------------------
    % Figures 4, 5, 6 (this figure, pie charts, per-delta bars): emerge-stage:scripts/stats/camera_ready_reissue_20261028/figures_dataset_stats_reissue_20261028.py
    % Run: python <script> --merged-root <v1.0.1>/corpus --eval-set-root <v1.0.1>/evaluation_set --output-dir <out> --output-suffix _v2
    % Output: annotation_stats_405bv1_v2.pdf, pie_chart_nr_triples_both_405bv1_v2.pdf, tkgu_distribution_deltas_405bv1_v2.pdf; test: check_dataset_figures.py
    % ------------------------------------------------------------------------------------------------------------------
    \begin{figure}
    \centering
    \includegraphics[trim=0 0 0 0, clip,  width=0.8\linewidth] {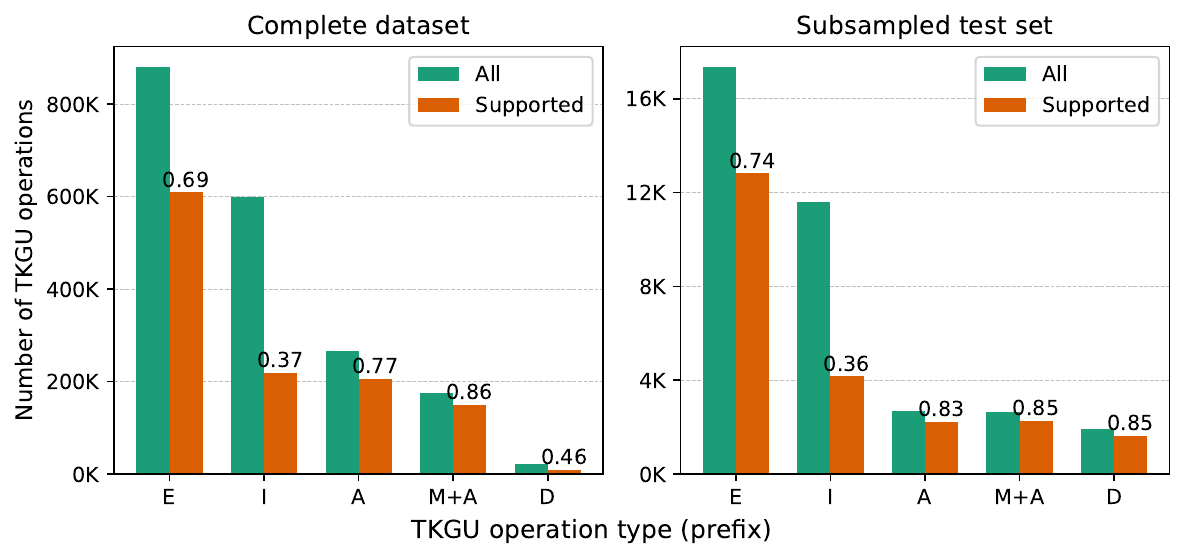}
    % {figures/2025.02.04.v3_MSCA_KG_deltas.pdf}
    % \caption{The ratio of TKGU operations supported by the LLM to the total number of TKGU operations mapped to textual passages during the alignment process.}
    \caption{The ratio of TKGU operations supported by the LLM to the total number of TKGU operations mapped to textual passages during the alignment process. TKGU operation types are denoted by the following prefixes: $\opexists \rightarrow \text{E}$, $\opadd \rightarrow \text{A}$, $\opmintadd \rightarrow \text{M{+}A}$, $\opinfer \rightarrow \text{I}$, and $\opdeprecate \rightarrow \text{D}$.}
    \label{fig:annotation_stats}
    \end{figure}
    \subsubsection{Annotation Agreement}
    \label{sec:appendix:human-annotation-agreement}
    We report annotation agreement between the two human annotators (\textsc{H–H Cohen's $\kappa$}), as well as between each human annotator and the LLM (\textsc{H1–LLM Cohen's $\kappa$} and \textsc{H2–LLM Cohen's $\kappa$}) in \Tabref{tab:annotation-agreement}. The Cohen’s $\kappa$ scores indicate strong agreement (0.6–0.8) to almost perfect agreement ($>$ 0.8). In addition, we compute Fleiss' $\kappa$ (\textsc{H+LLM Fleiss' $\kappa$}) and Krippendorff's $\alpha$ (\textsc{H+LLM Kripp. $\alpha$}) to assess agreement among all three annotators, both humans and the LLM. Consistent with Cohen's $\kappa$, these metrics also show strong to almost perfect agreement. This supports the use of the evaluated \texttt{Meta-Llama-3.1-405B} LLM to annotate full dataset using the prompts described in the Appendix \ref{sec:appendix:prompts}.
    %%%%%%%
    % script used to generate: /home/klim/work/w/repositories/emerge-stage/src/dataset/emerge/s09e_annotation_stats_for_paper.py
    %%%%%%%
    \begin{table}
    \caption{
    Annotation agreement per TKGU operation and overall. Columns show pairwise Cohen's $\kappa$ between humans (H-H) and between each human and the LLM (H1-LLM, H2-LLM), as well as multi-rater agreement including all three annotators (H+LLM) measured with Fleiss' $\kappa$ and Krippendorff's $\alpha$.
    }
    \label{tab:agreement_statistics}
        \centering
          \begin{tabular}{lccccc}
        \toprule
        \shortstack{TKGU \\ Operation } &
        \shortstack{H-H \\ Cohen's $\kappa$} &
        \shortstack{H1-LLM \\ Cohen's $\kappa$} &
        \shortstack{H2-LLM \\ Cohen's $\kappa$} &
        \shortstack{H+LLM \\ Fleiss' $\kappa$} &
        \shortstack{H+LLM \\ Kripp. $\alpha$} \\
        \midrule
        \opexists & 0.718 & 0.649 & 0.637 & 0.668 & 0.669 \\
    \opadd & 0.750 & 0.698 & 0.750 & 0.732 & 0.733 \\
    \opmintadd & 0.680 & 0.811 & 0.863 & 0.784 & 0.785 \\
    \opinfer & 0.880 & 0.840 & 0.761 & 0.827 & 0.827 \\
    \opdeprecate & 0.771 & 0.675 & 0.610 & 0.687 & 0.688 \\
    \midrule
    Overall & 0.792 & 0.765 & 0.744 & 0.767 & 0.767 \\
        \bottomrule
        \end{tabular}
        \label{tab:annotation-agreement}
    \end{table}
    \subsection{Triple Annotation Statistics}
    \label{sec:appendix:annotation-stats}
    \revklim{\Figref{fig:annotation_stats} shows, per operation type, the number of TKGU operations aligned with textual passages during the \textit{alignment} step (Section~\ref{sec:quality-control}) and how many of them the LLM curator marks as supported. The number above each pair of bars is the LLM-supported fraction (prompts in Appendix~\ref{sec:appendix:prompts}). The released data keep all aligned operations with these labels, and evaluation uses only the supported ones, \ie the \emph{Supported} bars (Section~\ref{sec:metrics-and-evaluation}). The supported fractions of the complete dataset and the test set are similar, except for \opdeprecate{}: only 46\% of its operations are supported in the complete dataset, but 85\% in the test set. This is because test-set subsampling deliberately selects instances that contain at least one supported \opdeprecate{} operation (Section~\ref{sec:experimental-setup}). The fraction stays below 100\% because selected \opdeprecate{} instances and instances sampled for the other TKGU operation types can also contain unsupported \opdeprecate{} operations.}
    % \Figref{fig:annotation_stats} illustrates the ratio of triples aligned with textual passages during the \textit{alignment} step described in Section~\ref{sec:quality-control} that were marked by automatic LLM annotations -- using the prompts detailed in Appendix~\ref{sec:appendix:prompts} -- as not representative of the passages.
    % This ratio is different between the \textit{complete} and \textit{subsampled} dataset used during testing. The reason is that during subsampling we retain instances with supported by LLM \opdeprecate~operations (see Section~\ref{sec:experimental-setup}).
    
    Additionally, we observe a lower fraction of \opinfer~operations supported by the LLM. This occurs because this TKGU operation type includes all entities in the KG, many of which are unrelated to the passage content but are connected to emerging entities mentioned in the text. Consequently, these triples are inherently less likely to be supported by the passages.
    A promising future direction is to develop information extraction methods that rely not only on textual evidence to extract triples but also integrate this content with existing knowledge and patterns in the KG. Such an approach could be particularly beneficial for incorporating emerging entities in \opinfer, even when they are not supported by textual passages, into the broader KG.

    \section{Dataset Statistics}
    \label{sec:appendix:additional-dataset-statistics}
    In this section we will present additional statistics of \datasetname.
    \subsection{Overall Statistics of \datasetname}
    % ------------------------------------------------------------------------------------------------------------------
    % Table 8: emerge-prod:src/stats/dataset/table_dataset_statistics.ipynb (layout; Entities = Wikipedia entity dictionaries,
    %   Rel. Types = yearly relation dictionaries data/indices/relik_edc_relation_indexes/<year>/documents.jsonl, all unchanged)
    % Camera-ready recounts on v1.0.1: Triples = KG snapshot rows (analysis_2021_dip_reissue_20261028.py, KG counts JSON);
    %   test-set Operations = 405B-supported operations per snapshot (one-off count, same rule as the stats macros; sum 23,107)
    % ------------------------------------------------------------------------------------------------------------------
    \begin{table}
    \caption{Statistics of our newly introduced \datasetname~dataset, organized by KG snapshots (rows). For each snapshot, we report the number of \textit{instances} and TKGU \textit{operations} in both the \textit{complete dataset} and the \textit{subsampled test set}. The \textit{KG statistics} section summarizes the number of entities, relation types, and triples in each KG snapshot.}
    \label{tab:dataset_statistics}
      \centering
    \begin{tabular}{c cccc ccc}
    \toprule
    & \multicolumn{2}{c}{Complete dataset} & \multicolumn{2}{c}{Subsampled test set} & \multicolumn{3}{c}{KG statistics} \\
    \cmidrule(lr){2-3}\cmidrule(lr){4-5}\cmidrule(l){6-8}
    Snapshot & Instances & Operations & Instances & Operations & Entities & Rel. Types & Triples  \\
    \midrule
    2019 & 37K & 173K & 500 & 3.24K & 5.96M & 5,646 & 26.35M  \\
    2020 & 31K & 159K & 500 & 3.21K & 6.14M & 7,017 & 29.47M  \\
    2021 & 40K & 232K & 500 & 3.82K & 6.34M & 8,216 & 31.63M  \\
    2022 & 30K & 155K & 500 & 2.86K & 6.54M & 9,425 & 34.30M  \\
    2023 & 26K & 126K & 500 & 2.91K & 6.67M & 10,599 & 35.93M  \\
    2024 & 32K & 169K & 500 & 3.45K & 6.80M & 11,409 & 37.32M  \\
    2025 & 33K & 180K & 500 & 3.62K & 6.93M & 12,304 & 38.59M  \\    % 2019 & 37K & 173K & 500 & 3.23K & 5.96M & 5,646 & 25.73M  \\
    % 2020 & 31K & 159K & 500 & 3.20K & 6.14M & 7,017 & 28.76M  \\
    % 2021 & 40K & 232K & 500 & 3.78K & 6.34M & 8,216 & 30.84M  \\
    % 2022 & 30K & 155K & 500 & 2.85K & 6.54M & 9,425 & 33.41M  \\
    % 2023 & 26K & 126K & 500 & 2.91K & 6.67M & 10,599 & 34.99M  \\
    % 2024 & 32K & 169K & 500 & 3.42K & 6.80M & 11,409 & 36.31M  \\
    % 2025 & 33K & 180K & 500 & 3.59K & 6.93M & 12,304 & 37.54M  \\
    % 2019 & 37K & 173K & 500 & 3.26K & 5.96M & 5,646 & 25.73M  \\
    % 2020 & 31K & 159K & 500 & 3.21K & 6.14M & 7,017 & 28.76M  \\
    % 2021 & 40K & 232K & 500 & 3.82K & 6.34M & 8,216 & 30.84M  \\
    % 2022 & 30K & 155K & 500 & 2.86K & 6.54M & 9,425 & 33.41M  \\
    % 2023 & 26K & 126K & 500 & 2.91K & 6.67M & 10,599 & 34.99M  \\
    % 2024 & 32K & 169K & 500 & 3.45K & 6.80M & 11,409 & 36.31M  \\
    % 2025 & 33K & 180K & 500 & 3.62K & 6.93M & 12,304 & 37.54M  \\
    \bottomrule
    \end{tabular}
      \vskip -0.1in
    \end{table}

    \Tabref{tab:dataset_statistics} presents key statistics of our newly introduced \datasetname~dataset, broken down by KG reference snapshots. For each snapshot, we report the number of \textit{instances} and TKGU \textit{operations} in both the full dataset and the subsampled test set. The table also summarizes \textit{KG} snapshots statistics, including the number of entities, relation types, and triples in each snapshot.
    We observe that the number of entities, relation types, and triples increases over time, reflecting the growth of Wikidata and the addition of new relations to the KG schema. This evolving structure creates a challenging scenario for future models, which must recognize these changes in the KG and adapt their predictions accordingly.
    \subsection{Number of TKGU Operations and Their Distribution}
    \label{sec:appendix:tkgu-operations-distribution}
    %
    %%%%%%%
    % Figure 5 (pie charts): same script as Figures 4-6 (block above Figure 4) -> pie_chart_nr_triples_both_405bv1_v2.pdf
    %%%%%%%
    \begin{figure}
    \centering
    \includegraphics[trim=0 0 0 0, clip,  width=1.0\linewidth] {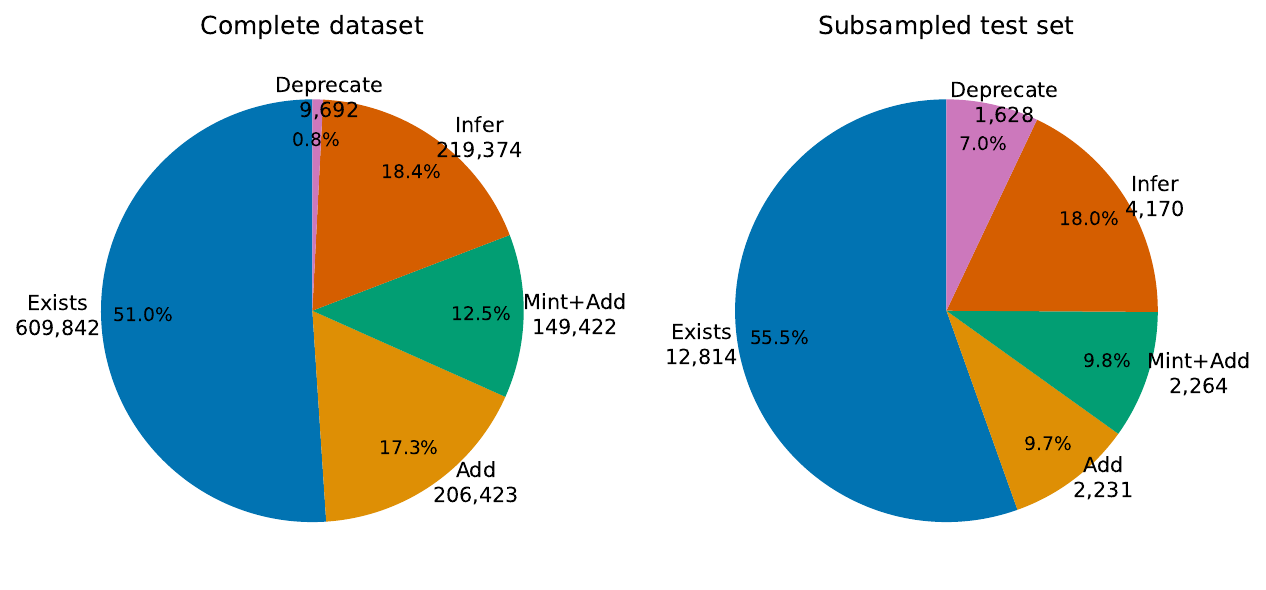}
    \caption{Distribution of TKGU operations defined in Section~\ref{sec:problem-definition} in \datasetname. The left subgraph shows the full dataset, while the right one shows the subsampled test set (see Section~\ref{sec:experimental-setup}). In the test set, \opdeprecate~TKGU operations are retained at higher frequency to ensure sufficient evaluation.}
    \label{fig:tkgu-types-distribution}
    \end{figure}

    %%%%%%%
    % Figure 6 (per-delta bars): same script as Figures 4-6 (block above Figure 4) -> tkgu_distribution_deltas_405bv1_v2.pdf
    %%%%%%%
    \begin{figure}[t]
    \centering
    \includegraphics[trim=0 0 0 0, clip,  width=1.0\linewidth] {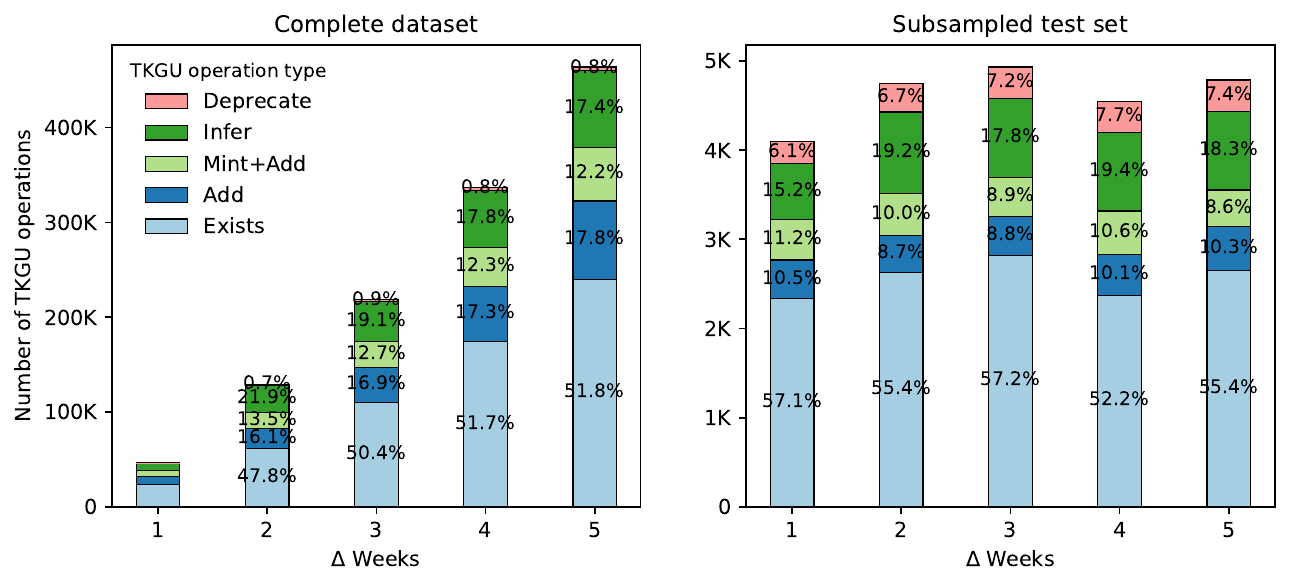}
    \caption{Distribution of TKGU operations across KG deltas up to 5 weeks defined in \datasetname.}
    \label{fig:tkgu-types-distribution-deltas}
    \end{figure}

    \Figref{fig:tkgu-types-distribution} illustrates the distribution of KG update operations for each TKGU type defined in Section~\ref{sec:problem-definition}. We report on both the complete dataset (left subgraph) and the subsampled test set (right subgraph). Furthermore, we display both the number as well as the percentage the operations of each of the TKGU types represent in \datasetname. \revklim{This distribution is similar between the complete dataset and subsampled test set, except for \opdeprecate~operations, which were retained at higher frequency in the test set to ensure sufficient evaluation (see Section~\ref{sec:experimental-setup}), and \opadd~operations, which are less frequent in the test set (9.7\% vs.\ 17.3\%).}
    % This distribution is very similar between the complete dataset and subsampled test set, except for \opdeprecate~operations, which were retained at higher frequency in the test set to ensure sufficient evaluation (see Section~\ref{sec:experimental-setup}).
    Additionally, \Figref{fig:tkgu-types-distribution-deltas} reports the distribution of TKGU update operations across weekly, temporally ordered KG deltas. In the \textit{Complete dataset} (left subplot), larger deltas induce more TKGU operations, yielding an increasing trend with delta size. In contrast, the \textit{subsampled test set} (right subplot) exhibits an approximately uniform distribution across deltas, since our subsampling retains exactly 100 instances per delta (see Section~\ref{sec:experimental-setup}), thereby equalizing the operation counts across delta sizes.
    % \newpage
    \section{Qualitative Analysis}
    \label{sec:appendix:qualitative-analysis}
    In this section, in \Tabrefs{tab:x-triples-examples}{tab:d-triples-examples} we present the five frequent factual triples from \datasetname~for each of the TKGU operation types, with an example of corresponding textual passage. The goal is to highlight representative cases that illustrate both the contents of the benchmark and the challenges it poses.
    The information in the tables contains the KG snapshot (\textit{Snap.}) used to compute weekly knowledge deltas aligned with each passage. We also report the number of occurrences of the triple in the \textit{Triple} column within \datasetname~(\textit{\#}), along with an example passage. The emerging entities in TKGU operations appear in bold. Due to space constraints, we selected the shortest passages; however, in \datasetname, passages consist of  full Wikipedia paragraphs.

    Our main observation is that the derived TKGU operations are closely aligned with the primary events occurring immediately after each KG snapshot (all snapshots are taken on January 1st of the corresponding year). We also note that the resulting triples are highly specific to the Wikidata KG structure. This is particularly evident in \Tabref{tab:ee-kg-triples-examples}, which shows examples of~\opinfer~TKGU operation, where an emerging entity must be connected to the existing KG. Consequently, we believe a promising future direction is to develop information extraction models that consider KG structure when proposing knowledge updates in it.

    Additionally, to illustrate the effectiveness of using an LLM (\texttt{Meta-Llama-3.1-405B}) to verify that all TKGU operations can be derived from their corresponding textual passages during the \textit{curation} step described in Section~\ref{sec:quality-control}, we present the most frequent factual triples from the \opinfer~TKGU operation in \datasetname~that were marked as \textit{not supported} by the LLM in \Tabref{tab:ee-kg-triples-filtered-out-examples}. These examples highlight triples that occurred frequently but were flagged because the LLM determined that their source textual passages did not support them. None of these triples are grounded in the corresponding text, demonstrating the reliability of the LLM-based validation process.
    \newpage

    \renewcommand{\arraystretch}{1.5} % 1.5 times the default row height
    % ------------------------------------------------------------------------------------------------------------------
    % Tables 9-14 (example tables): rows HAND-PICKED (kept from the submitted version); only the "#" column is computed:
    %   emerge-stage:scripts/stats/camera_ready_reissue_20261028/examples_update_counts_reissue_20261028.py --manuscript <this .tex> --corpus <v1.0.1>/corpus
    % "#" = distinct instances with the triple in that snapshot, 405B-supported (Table 13: 405B verdict False); passages that
    %   405B rejects were replaced by the shortest 405B-supported passage of the same triple. Old producer appendix_table_top_triples.ipynb used 8B.
    % ------------------------------------------------------------------------------------------------------------------
    \begin{table}
    \centering
    \caption{Example entries of the most frequent \opexists~TKGU operation instances in \datasetname, showing the snapshot (Snap.), triple, and number of instances (\#).}
    \label{tab:x-triples-examples}
    \begin{tabularx}{\linewidth}{l r >{\hsize=0.8\hsize}X >{\hsize=1.2\hsize}X}
    \toprule
    \textbf{Snap.} & \textbf{\#} & \textbf{Triple} &  \textbf{Example Passage} \\
    \midrule
    % 2021 & 834 & $\langle$Donald Trump; candidacy in election; 2020 United States presidential election$\rangle$ & Over the span of the 2020 presidential election, RSBN's coverage of Donald Trump's campaign rallies grossed over 127 million views on YouTube. \\
    % 2021 & 827 & $\langle$2020 United States presidential election; candidate; Donald Trump$\rangle$ &  In 2020, Pletts voiced support for Donald Trump and the Republican Party in the 2020 United States presidential election and Senate elections. \\
    % 2021 & 671 & $\langle$Joe Biden; candidacy in election; 2020 United States presidential election$\rangle$ & In September 2020, Kennedy Kent endorsed Republican President Donald Trump for reelection over Democratic nominee Joe Biden. \\
    % 2021 & 666 & $\langle$2020 United States presidential election; candidate; Joe Biden$\rangle$ & Despite being divorced, she remains good friends with her ex-husband, and she supported Joe Biden and Kamala Harris in the 2020 election. \\
    % 2021 & 586 & $\langle$midfielder; sport; association football$\rangle$& ``Niko Rak'' (born 26 July 2003) is a Croatian footballer who plays for Šibenik as a midfielder. \\
        2021 & 793 & $\langle$2020 United States presidential election; candidate; Donald Trump$\rangle$ &  In 2020, Pletts voiced support for Donald Trump and the Republican Party in the 2020 United States presidential election and Senate elections. \\
    2021 & 787 & $\langle$Donald Trump; candidacy in election; 2020 United States presidential election$\rangle$ & Over the span of the 2020 presidential election, RSBN's coverage of Donald Trump's campaign rallies grossed over 127 million views on YouTube. \\
    2021 & 666 & $\langle$2020 United States presidential election; candidate; Joe Biden$\rangle$ & Despite being divorced, she remains good friends with her ex-husband, and she supported Joe Biden and Kamala Harris in the 2020 election. \\
    2021 & 663 & $\langle$Joe Biden; candidacy in election; 2020 United States presidential election$\rangle$ & In September 2020, Kennedy Kent endorsed Republican President Donald Trump for reelection over Democratic nominee Joe Biden. \\
    2021 & 255 & $\langle$midfielder; sport; association football$\rangle$& ``Niko Rak'' (born 26 July 2003) is a Croatian footballer who plays for Šibenik as a midfielder. \\
    \bottomrule
    \end{tabularx}
    \end{table}
    \renewcommand{\arraystretch}{1}
    \renewcommand{\arraystretch}{1.5} % 1.5 times the default row height

    %%%%%%%
     % Table 10: see the block above Table 9 (hand-picked rows; "#" from examples_update_counts_reissue_20261028.py)
    %%%%%%%
    \begin{table}
    \centering
    \caption{Example entries of the most frequent \opadd~TKGU operation instances in \datasetname, showing the snapshot (Snap.), triple, and number of instances (\#).}
    \label{tab:e-triples-examples}
    \begin{tabularx}{\linewidth}{l r >{\hsize=0.8\hsize}X >{\hsize=1.2\hsize}X}
    \toprule
    \textbf{Snap.} & \textbf{\#} & \textbf{Triple} &  \textbf{Example Passage} \\
    \midrule
    2021 & 597 & $\langle$2020 United States presidential election; successful candidate; Joe Biden$\rangle$ & The winner of the 2020 election, President Joe Biden, is next in line to this purported curse. \\
    2020 & 161 & $\langle$Abu Mahdi al-Muhandis; military branch; Popular Mobilization Forces$\rangle$ & On 3 January 2020, a MQ-9 missile strike at the Baghdad International Airport killed Qasem Soleimani, the commander of Iran's Quds Force, and Abu Mahdi al-Muhandis, the commander of the Popular Mobilization Forces. \\
    2024 & 141 & $\langle$Houthi movement; country; Yemen$\rangle$ & On 8 November 2023, Houthi rebels in Yemen shot down a US Air Force MQ-9 over the Red Sea amid the attacks on U.S. bases in Iraq and Syria. \\
    2023 & 115 & $\langle$Kevin McCarthy; position held; Speaker of the United States House of Representatives$\rangle$ & On January 3, 2023, at the beginning of the 118th Congress, Boebert voted for Jim Jordan to be the U.S. House Speaker, in rebuke of House Minority Leader Kevin McCarthy. \\
    2020 & 114 & $\langle$Qasem Soleimani; place of death; Baghdad$\rangle$ & Soleimani was assassinated in a targeted U.S. drone strike on 3 January 2020 in Baghdad, which was approved by President Donald Trump on the grounds that Soleimani posed an ``imminent threat'' to American lives. \\
    % 2021 & 315 & $\langle$Joe Biden; position held; President of the United States$\rangle$ & On 20 January 2021, Joe Biden was sworn in as 46th President of the United States. \\
    % 2023 & 204 & $\langle$Kevin McCarthy; position held; Speaker of the United States House of Representatives$\rangle$ & On January 3, 2023, at the beginning of the 118th Congress, Boebert voted for Jim Jordan to be the U.S. House Speaker, in rebuke of House Minority Leader Kevin McCarthy. \\
    % 2020 & 168 & $\langle$Abu Mahdi al-Muhandis; military branch; Popular Mobilization Forces$\rangle$ & Abu Mahdi al-Muhandis returned to Iraq following the withdrawal of US troops (December 2011) to head the Kata'ib Hezbollah militia,; he then became deputy chief of the Popular Mobilization Forces. \\
    % 2024 & 164 & $\langle$Houthi movement; country; Yemen$\rangle$ &  On 28 March 2021, the Houthis forced 13 Jews to leave Yemen, they only allowed four elderly Jews to live in Yemen. \\
    % 2020 & 138 & $\langle$Qasem Soleimani; place of death; Baghdad$\rangle$ & Soleimani was assassinated in a targeted U.S. drone strike on 3 January 2020 in Baghdad, which was approved by President Donald Trump on the grounds that Soleimani posed an ``imminent threat'' to American lives. \\
    \bottomrule
    \end{tabularx}
    \end{table}
    \renewcommand{\arraystretch}{1}
    \renewcommand{\arraystretch}{1.5} % 1.5 times the default row height
    %%%%%%%
    % Table 11: see the block above Table 9 (hand-picked rows; "#" from examples_update_counts_reissue_20261028.py)
    %%%%%%%
    \begin{table}
    \centering
    % \small
    \caption{Example entries of the most frequent \opmintadd~TKGU operation instances in \datasetname~(emerging entities in bold), showing the snapshot (Snap.), triple, and number of instances (\#).}
    \label{tab:ee-triples-examples}
    \begin{tabularx}{\linewidth}{l r >{\hsize=0.8\hsize}X >{\hsize=1.2\hsize}X}
    \toprule
    \textbf{Snap.} & \textbf{\#} & \textbf{Triple} &  \textbf{Example Passage} \\
    \midrule
    2020 & 674 & $\langle$Qasem Soleimani; significant event; \textbf{assassination of Qasem Soleimani}$\rangle$ & He was killed by a targeted U.S. drone strike at the Baghdad International Airport on 3 January 2020, which also killed Iranian Armed Forces Major General Qasem Soleimani. \\
    2021 & 666 & $\langle$\textbf{January 6 United States Capitol attack}; significant person; Donald Trump$\rangle$ & CRP reported that President Trump's re-election campaign was financially related to the Capitol riot on January 6, 2021. \\
    2022 & 299 & $\langle$\textbf{Dawn FM}; performer; The Weeknd$\rangle$ & In 2022 the group also received credit for co producing songs off The Weeknds fifth studio album Dawn FM. \\
    2025 & 296 & $\langle$\textbf{2025 New Orleans truck attack}; located in the administrative territorial entity; New Orleans$\rangle$ & 2025 New Orleans truck attack: President Joe Biden has been briefed on the attack and has been in touch with New Orleans Mayor to offer support. \\
    2023 & 72 & $\langle$\textbf{Flowers}; performer; Miley Cyrus$\rangle$ & The chart's current number one as of the issue dated January 28, 2023, is ``Flowers'' by Miley Cyrus \\
    % 2021 & 848 & $\langle$\textbf{January 6 United States Capitol attack}; significant person; Donald Trump$\rangle$ & She called for the impeachment of President Donald Trump, in wake of the 2021 storming of the United States Capitol. \\
    % 2020 & 670 & $\langle$Qasem Soleimani; significant event; \textbf{assassination of Qasem Soleimani}$\rangle$ & He was killed by a targeted U.S. drone strike at the Baghdad International Airport on 3 January 2020, which also killed Iranian Armed Forces Major General Qasem Soleimani. \\
    % 2022 & 317 & $\langle$\textbf{Dawn FM}; performer; The Weeknd$\rangle$ & In 2022 the group also received credit for co producing songs off The Weeknds fifth studio album Dawn FM. \\
    % 2025 & 291 & $\langle$\textbf{2025 New Orleans truck attack}; located in the administrative territorial entity; New Orleans$\rangle$ & 2025 New Orleans truck attack: President Joe Biden has been briefed on the attack and has been in touch with New Orleans Mayor to offer support. \\
    % 2023 & 72 & $\langle$\textbf{Flowers}; performer; Miley Cyrus$\rangle$ & The chart's current number one as of the issue dated January 28, 2023, is ``Flowers'' by Miley Cyrus \\
    \bottomrule
    \end{tabularx}
    \end{table}
    \renewcommand{\arraystretch}{1}
    \renewcommand{\arraystretch}{1.5} % 1.5 times the default row height
    %%%%%%%
    % Table 12: see the block above Table 9 (hand-picked rows; "#" from examples_update_counts_reissue_20261028.py)
    %%%%%%%
    \begin{table}
    \centering
    % \small
    \caption{Example entries of the most frequent \opinfer~TKGU operation instances in \datasetname~(emerging entities in bold), showing the snapshot (Snap.), triple, and number of instances (\#).}
    \label{tab:ee-kg-triples-examples}
    \begin{tabularx}{\linewidth}{l r >{\hsize=0.8\hsize}X >{\hsize=1.2\hsize}X}
    \toprule
    \textbf{Snap.} & \textbf{\#} & \textbf{Triple} &  \textbf{Example Passage} \\
    \midrule
    2021 & 2169 & $\langle$\textbf{January 6 United States Capitol attack}; located in the administrative territorial entity; Washington, D.C.$\rangle$ & She called for the impeachment of President Donald Trump, in wake of the 2021 storming of the United States Capitol. \\
    2020 & 1065 & $\langle$\textbf{assassination of Qasem Soleimani}; instance of; assassination$\rangle$ & Assassination of Qasem Soleimani: the president called for restraint and said the events in Iraq were the result of previous ``terrorist acts''. \\
    2025 & 704 & $\langle$\textbf{Golden Age of Argentine cinema}; part of; history of film$\rangle$ & ``Huella'' is a 1940 Argentine film directed by Luis Moglia Barth during the Golden Age of Argentine cinema. \\
    2024 & 270 & $\langle$\textbf{South Africa v. Israel}; charge; genocide$\rangle$ & In 2023-24, he was appointed as a member of the South African legal team arguing ``South Africa v. Israel'' regarding the Genocide Convention. \\
    2019 & 153 & $\langle$\textbf{All Elite Wrestling}; instance of; business$\rangle$ & On January 1, 2019 Cody Rhodes unveiled a new promotion; All Elite Wrestling, in which he, along with Matt and Nick Jackson, will serve as Executive Vice President. \\
    \bottomrule
    \end{tabularx}
    \end{table}
    \renewcommand{\arraystretch}{1}

    \renewcommand{\arraystretch}{1.5} % 1.5 times the default row height
    %%%%%%%
    % Table 13: see the block above Table 9 (hand-picked rows; "#" = 405B verdict False, examples_update_counts_reissue_20261028.py)
    %%%%%%%
    \begin{table}
    \centering
    % \small
    \caption{Representative examples of the most frequent \opinfer~TKGU operation instances in \datasetname~\revklim{\textbf{marked as unsupported by the \texttt{Meta-Llama-3.1-405B} curator}}.~Emerging entities are in bold.~Each row shows the snapshot (Snap.), triple, and number of occurrences (\#). None of these triples are supported by the corresponding textual passages, illustrating the effectiveness of the LLM-based verification.}
    \label{tab:ee-kg-triples-filtered-out-examples}
    \begin{tabularx}{\linewidth}{l r >{\hsize=0.8\hsize}X >{\hsize=1.2\hsize}X}
    \toprule
    \textbf{Snap.} & \textbf{\#} & \textbf{Triple} &  \textbf{Example Passage} \\
    \midrule
    2021 & 3332 & $\langle$Proud Boys; significant event; \textbf{January 6 United States Capitol attack}$\rangle$ & Trump supporters infiltrated Capitol Hill in Washington DC., 5 people killed. \\
    2022 & 434 & $\langle$\textbf{Dawn FM}; distribution format; LP record$\rangle$ & In 2022 the group also received credit for co producing songs off The Weeknds fifth studio album Dawn FM. \\
    2024 & 290 & $\langle$\textbf{2024 Haneda Airport runway collision}; destination point; Niigata Airport$\rangle$ & 2024 Haneda Airport runway collision: All flights in and out of Haneda were suspended following the accident; operations currently remain suspended. \\
    2019 & 172 & $\langle$\textbf{All Elite Wrestling}; legal form; privately held company$\rangle$ & On January 1, 2019 Cody Rhodes unveiled a new promotion; All Elite Wrestling, in which he, along with Matt and Nick Jackson, will serve as Executive Vice President. \\
    2019 & 31 & $\langle$\textbf{@world\_record\_egg}; country; United Kingdom$\rangle$ & @world\_record\_egg is an account on social media platform Instagram, notable for holding the world records for both the most-liked Instagram post and most liked online post on any media platform in history. \\
    \bottomrule
    \end{tabularx}
    \end{table}
    \renewcommand{\arraystretch}{1}
    \renewcommand{\arraystretch}{1.5} % 1.5 times the default row height
    %%%%%%%
     % Table 14: see the block above Table 9 (hand-picked rows; "#" from examples_update_counts_reissue_20261028.py)
    %%%%%%%
    \begin{table}
    \centering
    % \small
    \caption{Example entries of the most frequent \opdeprecate~TKGU operation instances in \datasetname~(emerging entities in bold), showing the snapshot (Snap.), triple, and number of instances (\#).}
    \label{tab:d-triples-examples}
    \begin{tabularx}{\linewidth}{l r >{\hsize=0.8\hsize}X >{\hsize=1.2\hsize}X}
    \toprule
    \textbf{Snap.} & \textbf{\#} & \textbf{Triple} &  \textbf{Example Passage} \\
    \midrule
    2021 & 75 & $\langle$Parler; distributed by; Google Play$\rangle$ & After complaints that Parler was used to coordinate the 2021 storming of the U.S. Capitol, Apple and Google removed Parler's mobile app from their app stores. Parler went offline on January 10, 2021 at 11:59 PM (PST) after Amazon Web Services canceled its hosting services. \\
    2021 & 75 & $\langle$Mike Pence; position; Vice President of the United States$\rangle$ & ``Marlon Bundo'', also known as ``Bunny of the United States'' (``BOTUS''), is a rabbit, belonging to the family of Mike Pence, the 48th and former Vice President of the United States. \\
    2025 & 63 & $\langle$Vice President of the United States; position holder; Kamala Harris$\rangle$ & West is the brother-in-law of former Vice President Kamala Harris. He served as an advisor to her 2024 presidential campaign. \\
    2024 & 62 & $\langle$\textbf{Adam Peters}; member of sports team; San Francisco 49ers$\rangle$ & Peters joined the Denver Broncos as a scout in 2009. He was promoted to assistant director of college scouting in July 2014 and to director of college scouting in 2016. He was a member of the team that won Super Bowl 50 in 2015. \\
    2020 & 43 & $\langle$European Union; has part(s); United Kingdom$\rangle$ & Chris Davies was the chairman (2019 – 2020) - until the United Kingdom left the European Union. \\
    \bottomrule
    \end{tabularx}
    \end{table}
    \renewcommand{\arraystretch}{1}

    \section{EDC+ Prompt}
    \label{sec:appendix:edc-prompts}
    The following prompt is designed to identify all the operations to update the KG defined in Section~\ref{sec:problem-definition}. Concretely, it allows to identify triples explicitly mentioned in text under \textit{Triples in text} category. This includes triples covering \opexists, \opadd, and \opmintadd~TKGU operations. It also allows to classify \textit{Triples in text} in those that should be added to the KG (\ie with the ADD tag), and those that should be deprecated (\ie with the DEPRECATE tag). This way, the prompt also facilitates the identification of KG triples that may need to be deprecated (\ie \opdeprecate~TKGU operation). Finally, the prompt allows to generate triples covering \opinfer~TKGU operation under \textit{Triples not in text} category, by asking LLM to identify triples with only one single entity (head or tail) mentioned in text, and the other entity potentially existing in the KG.

    \begin{promptbox}
    \ttfamily
    Your task is to transform the given text into a semantic graph in the form of a list of triples. Two sets of triples are to be extracted: `Triples in text', which contain triples relating entities mentioned in text in the form of [Entity1, Relationship, Entity2, Action], where action indicates if the triple has to be added (action `ADD') or deprecated (action `DEPRECATE') from the graph based on the knowledge in text. The second set of triples is called `Triples not in text', and consists of triples with one entity (head or tail) mentioned in text and the other entity not mentioned in text but potentially existing in the graph.

    In your answer, please strictly only include the triples and do not include any explanation or apologies.

    Here are some examples:
    \\ \\
    \texttt{<FEW\_SHOT\_EXAMPLES>}
    \\ \\
    Now please extract triplets from the following text.
    \\ \\
    Text: \texttt{<INPUT\_TEXT>}

    \end{promptbox}

    % Running ReLiK on the subsampled \datasetname~test set takes about 5 hours on a single A100 GPU.
    \clearpage
    \section{Additional Analysis}
    \subsection{Model Performance Across Temporal Snapshots and Deltas}
    \label{sec:appendix:quantitative-predictions}
    %%%%%%%
     % Figure 7 (both panels): same notebook as Figure 3 (block above Figure 3) -> plot_deltas_overlay_both_metrics_v2_empty_pred_fix.pdf, plot_snapshots_overlay_both_metrics_v2_empty_pred_fix.pdf
    %%%%%%%
    \begin{figure}
    \centering
    \includegraphics[trim=0 0 0 0,clip,width=\textwidth]{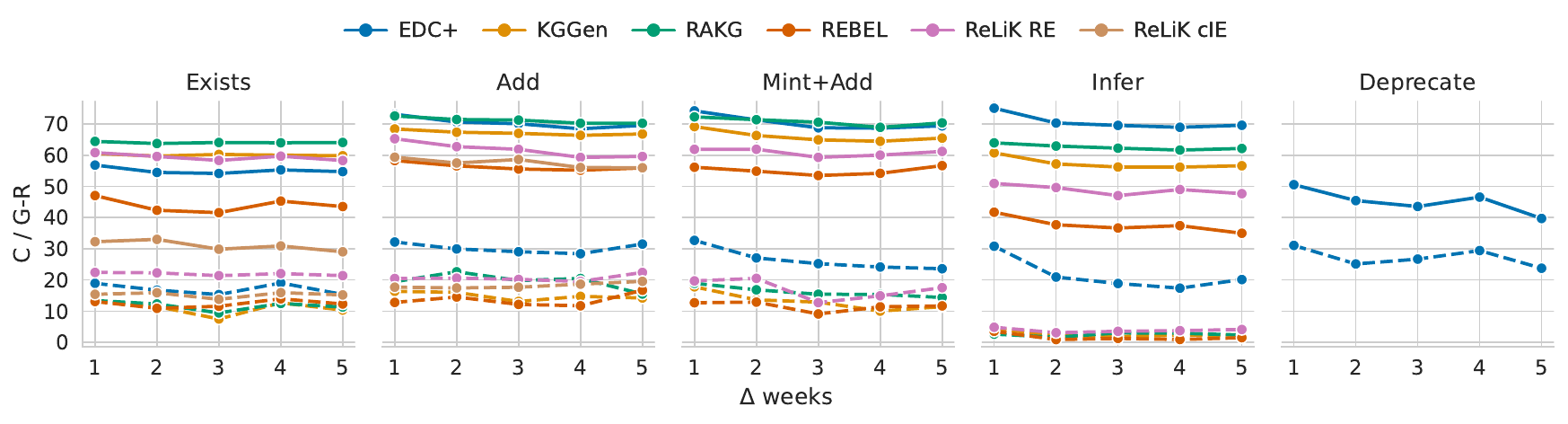}
    \vspace{-2mm} % tighten/adjust (try -1mm to -4mm)
    \includegraphics[trim=0 0 0 0,clip,width=\textwidth]{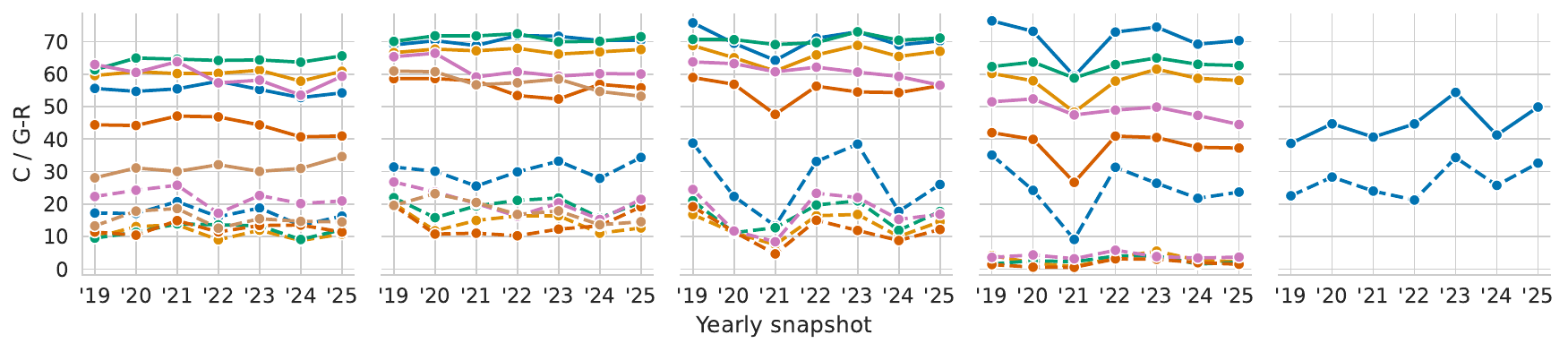}
    \caption{
    G-BERTScore-R (G-R, solid) and Completeness (C, dashed) across temporal KG deltas (top) and yearly KG snapshots (bottom). Performance decreases for larger weekly deltas, particularly for \opmintadd{}, \opinfer{}, and \opdeprecate{}, indicating increased difficulty as the KG becomes more stale. Across yearly snapshots, no consistent trend is observed, aside from a performance drop in 2021 for \opmintadd{} and \opinfer{} (see analysis in Section~\ref{sec:appendix:2021-dip}).}
    \label{fig:appendix:deltas-performance}
    \end{figure}
    \Figref{fig:appendix:deltas-performance} shows model performance (G-BERTScore-R) across temporal deltas (top) and KG snapshots (bottom). We observe a general degradation in performance for TKGU operations as the temporal delta between the KG snapshot and the updated KG increases. This effect is particularly pronounced for operations involving newly emerging entities (\opmintadd{} and \opinfer{}) and for triple deprecation (\opdeprecate{}), suggesting that models struggle increasingly to update older, stale KG snapshots.

    Considering model performance across yearly KG snapshots (bottom of \Figref{fig:appendix:deltas-performance}), we do not observe a monotonic degradation over time, indicating that models retain a relatively stable ability to extract relevant knowledge from text across snapshots. Notably, however, there is a consistent drop in performance around the 2021 KG snapshot.

%%%%%%%%%%%%%%%%%%%%%%%%%%%%%%%%%%%%

% \clearpage
\subsection{Effect of In-Context Learning on EDC+}
\label{sec:appendix:icl-ablation}
%%%%%%%
% Figure 8: emerge-stage:src/stats/evaluation/figure_zs_vs_icl_reissue_20261028.ipynb (PKL_PATH=<D3 results file> SAVE_DIR=<out>, run like Figure 3)
%   -> zs_vs_icl_v2_empty_pred_fix_hlegend.pdf; test: check_icl_values.py (40/40 bars = D3; main-text "33.4 / 50.0" = largest drops)
%%%%%%%
\begin{figure}[t]
\centering
\includegraphics[trim=0 10 0 10,clip,width=0.8\textwidth]{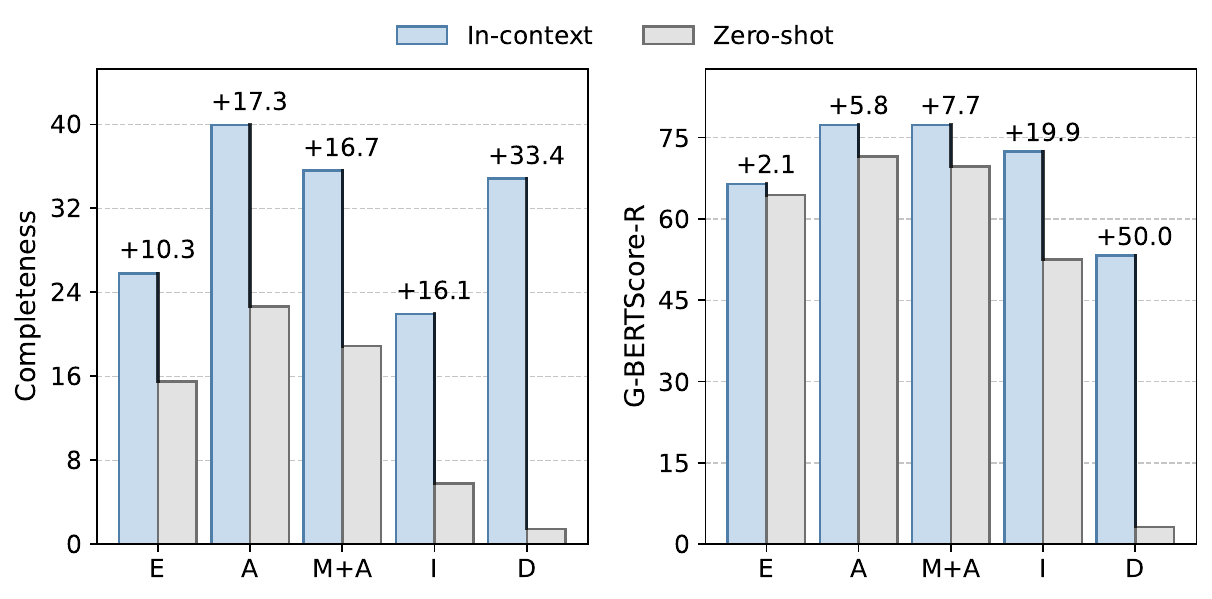}
\caption{
Completeness and G-BERTScore-R of EDC+ with \texttt{GPT-5.1} under \textit{in-context} learning versus \textit{zero-shot} prompting across TKGU operation types: \opexists~(\opexistsAcronym), \opadd~(\opaddAcronym), \opmintadd~(\opmintaddAcronym), \opinfer~(\opinferAcronym), and \opdeprecate~(\opdeprecateAcronym).}
\label{fig:in-context-zero-shot-performance}
\end{figure}

To complement the discussion in Section~\ref{sec:experimental-setup} on the contribution of in-context learning to EDC+ performance, \Figref{fig:in-context-zero-shot-performance} reports Completeness and G-BERTScore-R of EDC+ with \texttt{GPT-5.1} under in-context learning (five demonstrations drawn from outside the test set) versus zero-shot prompting, broken down by TKGU operation. Removing the demonstrations consistently degrades performance across all operations, with the largest absolute drop on \opdeprecate{} (33.4 points Completeness, 50.0 points G-BERTScore-R; full per-operation values are read off the figure).
As noted in the main text, the present ablation does not isolate whether the residual gains on the other operations stem from implicit exposure to KG content and structure or from generic few-shot effects; we leave this disentanglement to future work.

%%%%%%%%%%%%%%%%%%%%%%%%%%%%%%%%%%%%

    % \clearpage
    \subsection{Qualitative Analysis of Predictions}
    \label{sec:appendix:qualitative-predictions}
    \Tabref{table:appendix:prediction:examplels} showcases randomly selected EDC+ predictions generated with \texttt{GPT-5.1}. While the predicted triples are largely semantically correct, they often diverge from the ground truth, which is tightly coupled to the structure and schema of the underlying knowledge graph (KG). To facilitate comparison, each ground-truth triple is paired with its most similar predicted triple according to BERTScore.
    \renewcommand{\arraystretch}{1.12}
    %%%%%%%
    % script used to generate: /home/klim/work/w/repositories/emerge-stage/src/stats/evaluation/appendix_table_qualitative_predictions.ipynb
    %%%%%%%
    \begin{table}
    \centering
    \small

    \setlength{\tabcolsep}{2pt}
    \rowcolors{3}{gray!8}{white}
    \caption{Randomly selected examples of EDC+ predictions using \texttt{GPT-5.1}. While most predicted triples are semantically correct (\ie reflect the information in passage), they often differ from the ground truth triples, which depend strongly on the underlying knowledge graph schema. Each ground truth triple is paired with its most similar predicted triple according to BERTScore.}
    \begin{tabular}{
      p{0.11\textwidth}
      p{0.36\textwidth}
      p{0.37\textwidth}
      p{0.11\textwidth}
    }
    \toprule
    \multicolumn{4}{p{0.98\textwidth}}{\textbf{Passage 1. (KG Snapshot 2025)} Shafique Ahmed Siddique: == Personal life == Siddique married Sheikh Rehana in 1977 in London, sister of Prime Minister Sheikh Hasina. The couple have a son and two daughters, one of whom is Tulip Rizwana Siddiq.} \\

    \midrule
    \textbf{TKGU} & \textbf{Ground Truth} & \textbf{Predicted} & \textbf{BERTScore} \\
    \midrule
    \opexists & $\langle$Sheikh Rehana; sibling; Sheikh Hasina$\rangle$ & $\langle$Sheikh Rehana; sibling; Sheikh Hasina$\rangle$ & \cellcolor{blue!60}1.000 \\
    \opadd & $\langle$Shafiq Ahmed Siddiq; spouse; Sheikh Rehana$\rangle$ & $\langle$Shafique Ahmed Siddique; spouse; Sheikh Rehana$\rangle$ & \cellcolor{blue!54}0.956 \\
    \opadd & $\langle$Shafiq Ahmed Siddiq; child; Tulip Siddiq$\rangle$ & $\langle$Shafique Ahmed Siddique; child; Tulip Rizwana Siddiq$\rangle$ & \cellcolor{blue!47}0.898 \\
    \opadd & $\langle$Sheikh Rehana; spouse; Shafiq Ahmed Siddiq$\rangle$ & $\langle$Shafique Ahmed Siddique; spouse; Sheikh Rehana$\rangle$ & \cellcolor{blue!46}0.889 \\
    \opexists & $\langle$Sheikh Rehana; child; Tulip Siddiq$\rangle$ & $\langle$Shafique Ahmed Siddique; child; Tulip Rizwana Siddiq$\rangle$ & \cellcolor{blue!33}0.781 \\
    \opexists & $\langle$Tulip Siddiq; mother; Sheikh Rehana$\rangle$ & $\langle$Shafique Ahmed Siddique; spouse; Sheikh Rehana$\rangle$ & \cellcolor{blue!32}0.776 \\
    %\opexists & $\langle$Sheikh Hasina; relative; Tulip Siddiq$\rangle$ & $\langle$Shafique Ahmed Siddique; child; Tulip Rizwana Siddiq$\rangle$ & \cellcolor{blue!25}0.716 \\
    %\opexists & $\langle$Tulip Siddiq; relative; Sheikh Hasina$\rangle$ & $\langle$Shafique Ahmed Siddique; child; Tulip Rizwana Siddiq$\rangle$ & \cellcolor{blue!23}0.702 \\
    \bottomrule
    \toprule
    %  \vspace{-0.8ex}
    % \toprule
    \multicolumn{4}{p{0.98\textwidth}}{\textbf{Passage 2. (KG Snapshot 2025)} The Robert Bridges House was standing in Pacific Palisades, Los Angeles. The house stood on concrete pillars above Sunset Boulevard. It was destroyed in the January 2025 Southern California wildfires.} \\

    \midrule
    \textbf{TKGU} & \textbf{Ground Truth} & \textbf{Predicted} & \textbf{BERTScore} \\
    \midrule
    \opinfer & $\langle$Robert Bridges House; country; United States$\rangle$ & $\langle$Robert Bridges House; country; United States$\rangle$ & \cellcolor{blue!60}100.0 \\
    \opmintadd & $\langle$Robert Bridges House; significant event; January 2025 Southern California wildfires$\rangle$ & $\langle$Robert Bridges House; significant event; January 2025 Southern California wildfires$\rangle$ & \cellcolor{blue!59}100.0 \\
    \opinfer & $\langle$January 2025 Southern California wildfires; instance of; wildfire$\rangle$ & $\langle$January 2025 Southern California wildfires; instance of; wildfire$\rangle$ & \cellcolor{blue!59}100.0 \\
    \opmintadd & $\langle$Robert Bridges House; cause of destruction; January 2025 Southern California wildfires$\rangle$ & $\langle$Robert Bridges House; significant event; January 2025 Southern California wildfires$\rangle$ & \cellcolor{blue!40}85.0 \\
    \opinfer & $\langle$Robert Bridges House; located in the administrative territorial entity; Los Angeles$\rangle$ & $\langle$Sunset Boulevard; located in the administrative territorial entity; Los Angeles$\rangle$ & \cellcolor{blue!34}80.6 \\
    \opinfer & $\langle$January 2025 Southern California wildfires; country; United States$\rangle$ & $\langle$January 2025 Southern California wildfires; instance of; wildfire$\rangle$ & \cellcolor{blue!32}79.2 \\
    \opmintadd & $\langle$Robert Bridges House; location; Pacific Palisades$\rangle$ & $\langle$Robert Bridges House; located in the administrative territorial entity; Pacific Palisades, Los Angeles$\rangle$ & \cellcolor{blue!18}68.4 \\
    \opinfer & $\langle$Robert Bridges House; instance of; house$\rangle$ & $\langle$Robert Bridges House; country; United States$\rangle$ & \cellcolor{blue!10}62.1 \\
    \bottomrule
    \toprule
    \multicolumn{4}{p{0.98\textwidth}}{\textbf{Passage 3. (KG Snapshot 2021)}  President Joe Biden, who defeated Donald Trump in the 2020 presidential election, has vowed to bring the United States back into the Paris Agreement on his first day in office, as well as renewing America's commitment to mitigating climate change.} \\

    \midrule
    \textbf{TKGU} & \textbf{Ground Truth} & \textbf{Predicted} & \textbf{BERTScore} \\
    \midrule
    \opdeprecate & $\langle$Donald Trump; position held; President of the United States$\rangle$ & $\langle$Donald Trump; position held; President of the United States$\rangle$ & \cellcolor{blue!60}100.0 \\
    \opdeprecate & $\langle$United States of America; head of government; Donald Trump$\rangle$ & $\langle$Donald Trump; position held; President of the United States$\rangle$ & \cellcolor{blue!12}62.2 \\
    \opdeprecate & $\langle$United States of America; head of state; Donald Trump$\rangle$ & $\langle$Donald Trump; position held; President of the United States$\rangle$ & \cellcolor{blue!10}60.8 \\
    \bottomrule
    \end{tabular}
    \label{table:appendix:prediction:examplels}
    \end{table}
    \renewcommand{\arraystretch}{1}
    \clearpage

    \subsection{Analysis of the 2021 Performance Dip}
    \label{sec:appendix:2021-dip}

    The main-text analysis (right panel of \Figref{fig:deltas-performance}) shows a drop in G-BERTScore-R around the 2021 snapshot for \opinfer{} and \opmintadd{}. We investigate whether this drop is driven by a \emph{distributional shift} in relation types in the 2021 dataset relative to the underlying KG.

\begin{figure}
    \centering
    \includegraphics[width=0.95\linewidth]{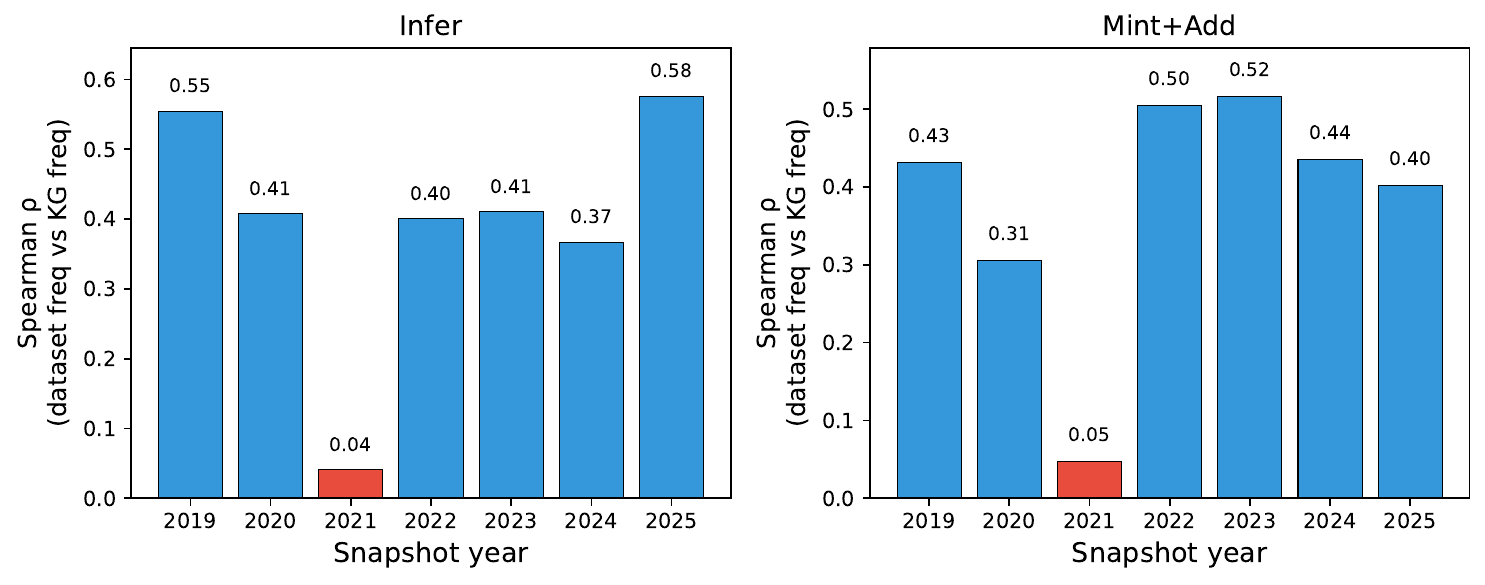}
    \caption{Spearman rank correlation between relation-type frequencies in the dataset and the KG, per yearly snapshot, for \opinfer{} and \opmintadd{}. The 2021 snapshot is a clear outlier ($\rho \approx 0.05$), indicating that relation types frequent in the KG are not the ones most frequent in the dataset.}
    \label{fig:appendix:2021-spearman}
    \end{figure}

\begin{figure}
    \centering
    \includegraphics[width=0.95\linewidth]{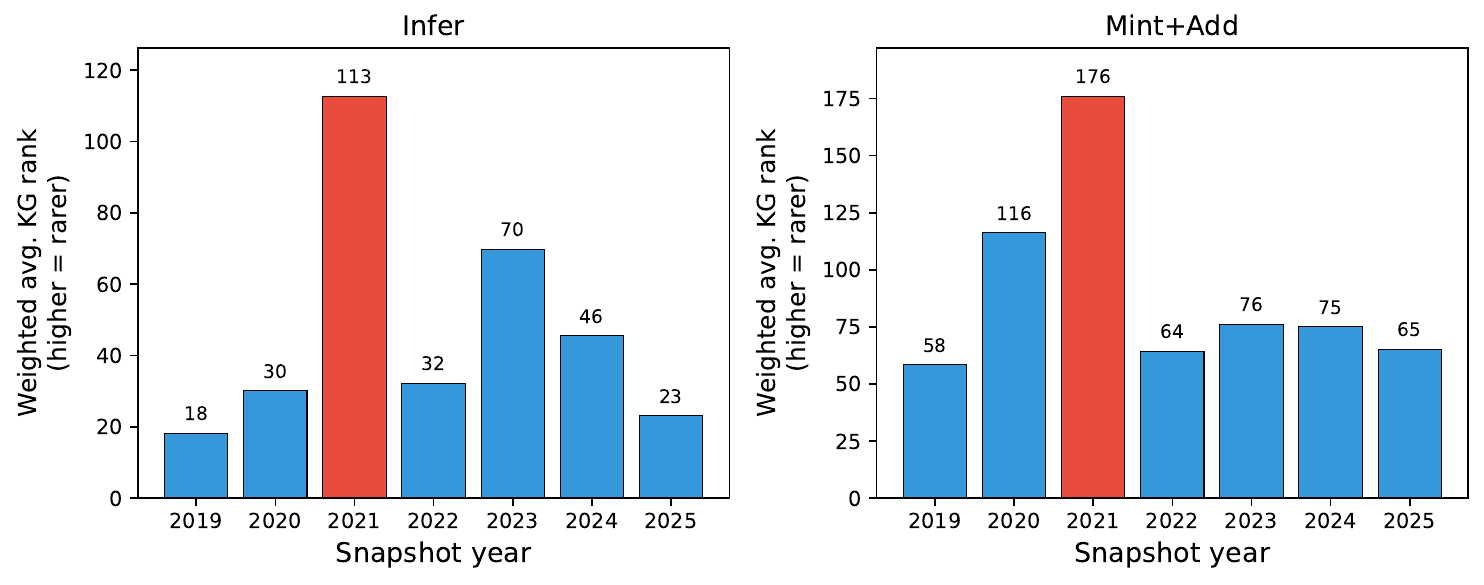}
    \caption{Weighted average KG rank of dataset relation types, per yearly snapshot, for \opinfer{} and \opmintadd{}. Higher values indicate rarer relation types in the KG. The 2021 snapshot stands out with substantially higher ranks, indicating that the dataset emphasizes rarer relation types compared to other years.}
    \label{fig:appendix:2021-kg-rank}
    \end{figure}

    \textbf{Distributional shift.} For each yearly snapshot $t$ and each TKGU operation $o \in \{\opmintadd, \opinfer\}$, we compare relation-type frequencies in the dataset triples and the KG snapshot $G_t$ along two axes:
    \begin{enumerate}
    \item The \textbf{Spearman rank correlation} between relation-type frequencies (\Figref{fig:appendix:2021-spearman}), which measures whether the same relation types are frequent in both the KG and the dataset.
    \item The \textbf{weighted average KG rank} of relation types in the dataset (\Figref{fig:appendix:2021-kg-rank}), computed as $\sum_r p_{\text{data}}(r)\,\text{rank}_{G_t}(r)$, where $p_{\text{data}}(r)$ is the empirical frequency of relation type $r$ in the dataset and $\text{rank}_{G_t}(r)$ ranks relation types by frequency in $G_t$ (higher rank = rarer).
    \end{enumerate}
    Both metrics show a pronounced outlier in 2021. For \opinfer{}, the Spearman correlation \revklim{drops from $0.37$--$0.58$} in other snapshots to $\rho = 0.04$ in 2021; for \opmintadd{}, \revklim{it drops from $0.31$--$0.52$ to} $\rho = 0.05$. The weighted KG rank rises correspondingly. Together, these results indicate that 2021 dataset triples overrepresent rarer KG relation types compared to other years.
    %%%%%%% END OLD VERSION

    %%%%%%%
% Figures 9-11, Tables 16-18 (2021 analysis): emerge-stage:scripts/stats/camera_ready_reissue_20261028/analysis_2021_dip_reissue_20261028.py
    %   --test-set <v1.0.1 test set> --kg-dir <v1.0.1 KG snapshots> --prop-labels <Elements .../s03_get_wikidata_labels/20250321/map_property_id_to_label.tsv> --pkl <D3> --manuscript <this .tex>
    %   -> scatter PDF + analysis2021__numbers.json (all numbers of this section and of the main-text 2021 sentence)
    % Figures 9-10 re-rendered with label headroom, same numbers: replot_2021_bars_reissue_20261029.py --numbers analysis2021__numbers.json
    %%%%%%%

    %%%%%%%
    % Figure 10: see block above Figure 9 -> fig_kg_rank_infer_mintadd_v3.pdf (replot_2021_bars_reissue_20261029.py)
    %%%%%%%

\begin{figure}
    \centering
    \includegraphics[width=1.0\linewidth]{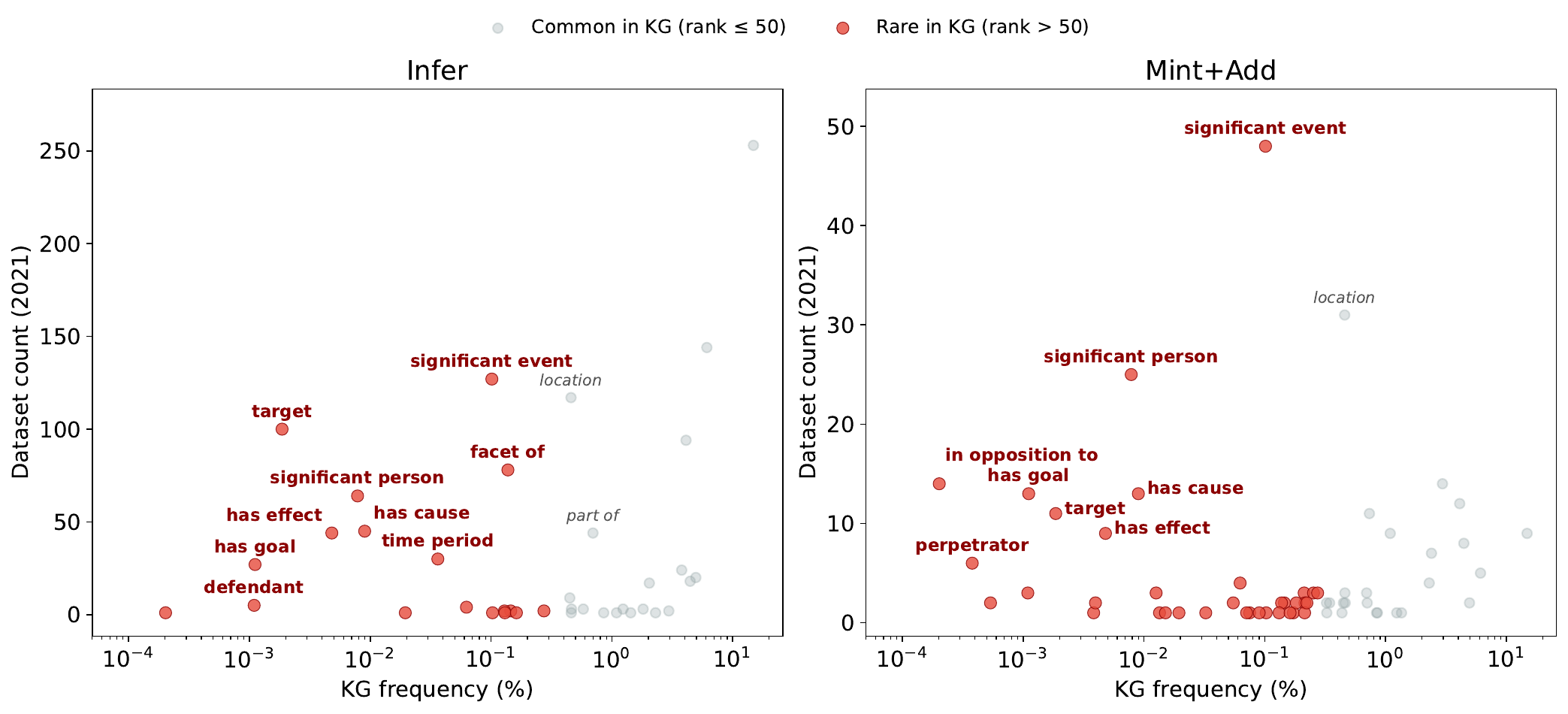}
    \caption{Relation-type frequencies in the 2021 KG snapshot (x-axis) vs.\ the 2021 \opinfer{} dataset (y-axis). The upper-left cluster (red) contains relation types that are rare in the KG (rank $> 50$) but occur at least five times in the 2021 dataset.
    % These relation types account for $\sim$45\% of 2021
    \revklim{These relation types account for $\sim$40\% of 2021}
    \opinfer{} triples and a similar share of 2021 \opmintadd{} triples.}
    \label{fig:appendix:2021-scatter}
    \end{figure}

    \textbf{KG-rare relation types concentrated in 2021.}
    To identify \emph{which} relation types drive the shift, \Figref{fig:appendix:2021-scatter} plots, for each relation type appearing in the 2021 \opinfer{} triples, its frequency in the 2021 KG snapshot against its frequency in the 2021 dataset. The upper-left cluster contains relation types that are rare in the 2021 KG (rank $> 50$) yet occur at least five times in the 2021 dataset: \textit{significant event}, \textit{target}, \textit{facet of}, \textit{significant person}, \revklim{\textit{has cause}, \textit{has effect}, \textit{time period}, \textit{has goal}, and \textit{defendant}.}
    % \textit{has cause}, \textit{has effect}, \textit{has goal}, \textit{perpetrator}, \textit{defendant}, and \textit{in opposition to}.
    \Tabref{tab:appendix:2021-cross-year} makes the year-over-year concentration explicit: these relation types appear almost exclusively in the 2021 snapshot, consistent with contemporaneous events such as the U.S. presidential transition and the January 2021 Capitol attack.

    %%%%%%%
    % Figure 11: see block above Figure 9 -> fig_scatter_2021_event_driven_v5_v5.pdf (replot_2021_scatter_reissue_20261030.py: legend outside, labels unclipped; data from analysis_2021_dip_reissue_20261028.py)
    %%%%%%%

    %%%%%%%
    % Table 16: see block above Figure 9 (rows = relations with KG rank > 50 and >= 5 Infer triples in 2021; numbers JSON "cross_year_infer")
    %%%%%%%
    \begin{table}
    \centering
    \small
    \caption{Cross-year counts of LLM-curated \opinfer{} triples whose relation type has KG rank $> 50$ in the 2021 snapshot and occurs at least five times in the 2021 dataset. These relation types are concentrated almost entirely in 2021, indicating a strong shift toward rare relation types in that snapshot.}
    \label{tab:appendix:2021-cross-year}
    \begin{tabular}{lrrrrrrr}
    \toprule
    \textbf{Relation} & \textbf{2019} & \textbf{2020} & \textbf{2021} & \textbf{2022} & \textbf{2023} & \textbf{2024} & \textbf{2025} \\
    \midrule
    significant event   & 1 & 17 & 127 & 3 & 0 & 6 & 5 \\
    target              & 1 &  6 & 100 & 0 & 3 & 0 & 0 \\
    facet of            & 0 & 18 &  78 & 0 & 1 & 2 & 12 \\
    significant person  & 0 &  0 &  64 & 0 & 0 & 0 & 3 \\
    has cause           & 0 &  0 &  45 & 0 & 1 & 0 & 1 \\
    has effect          & 0 &  0 &  44 & 5 & 0 & 0 & 2 \\
    time period         & 0 &  0 &  30 & 0 & 1 & 0 & 2 \\
    has goal            & 0 &  0 &  27 & 0 & 3 & 0 & 0 \\
    defendant           & 0 &  0 &   5 & 0 & 0 & 5 & 0 \\
    \bottomrule
    \end{tabular}
    \end{table}

    \textbf{KG-rare relation types explain the performance dip.}
    We next test whether the KG-rare relation types identified above are the ones that decrease performance in 2021. We classify each 2021 test instance by the KG rank of its relation type into \emph{rare} (rank $> 50$) or \emph{common} (rank $\leq 50$), and compare mean G-BERTScore-R across all benchmarked models. Instances involving KG-rare relation types
    % score on average $17.7$ points lower
    \revklim{score on average $17.8$ points lower}
    for \opinfer{} and $8.0$ points lower for \opmintadd{} than instances with common relation types. This per-instance gap is consistent with, and large enough to account for, the snapshot-level dip in \Figref{fig:deltas-performance}.

    \textbf{Representative triples.}
    \Tabref{tab:appendix:2021-infer-examples} and \Tabref{tab:appendix:2021-mintadd-examples} show representative 2021 \opinfer{} and \opmintadd{} triples whose relation type has KG rank $> 50$ in the 2021 snapshot. The examples are characteristic of contemporaneous 2021 events (e.g., the Capitol attack, the second impeachment of Donald Trump, the 2020 U.S. presidential election, and the 2021 Myanmar coup).

    %%%%%%%
    % Table 17: rows hand-picked (submitted); KG ranks from analysis_2021_dip_reissue_20261028.py (numbers JSON "examples")
    %%%%%%%
    \begin{table}
    \centering
    \small
    \caption{Representative \opinfer{} triples from the 2021 snapshot whose relation type has KG rank $> 50$. The examples reflect clusters of 2021 political events (e.g., the Capitol attack, second impeachment, and 2020 election) associated with the concentration reported in \Tabref{tab:appendix:2021-cross-year}.}
    \label{tab:appendix:2021-infer-examples}
    \begin{tabularx}{\linewidth}{X l X r}
    \toprule
    \textbf{Head} & \textbf{Relation} & \textbf{Tail} & \textbf{KG rank} \\
    \midrule
    % United States Capitol & significant event & January 6 United States Capitol attack & 103 \\
    % Donald Trump & significant event & second impeachment of Donald Trump & 103 \\
    % first presidency of Donald Trump & significant event & January 6 United States Capitol attack & 103 \\
    % Tatmadaw & significant event & 2021 Myanmar coup d'\'etat & 103 \\
    % January 6 United States Capitol attack & target & United States Congress & 452 \\
    % January 6 United States Capitol attack & facet of & presidential transition of Joe Biden & 86 \\
    % second impeachment of Donald Trump & facet of & presidential transition of Joe Biden & 86 \\
    % second impeachment of Donald Trump & significant person & Donald Trump & 312 \\
    % January 6 United States Capitol attack & significant person & Donald Trump & 312 \\
    % January 6 United States Capitol attack & has cause & 2020 United States presidential election & 299 \\
    % 2020 United States presidential election & has effect & January 6 United States Capitol attack & 358 \\
    % January 6 United States Capitol attack & has goal & attempts to overturn the 2020 United States presidential election & 507 \\
    % January 6 United States Capitol attack & time period & first presidency of Donald Trump & 171 \\
    % second impeachment of Donald Trump & defendant & Donald Trump & 510 \\
    United States Capitol & significant event & January 6 United States Capitol attack & 103 \\
    Donald Trump & significant event & second impeachment of Donald Trump & 103 \\
    first presidency of Donald Trump & significant event & January 6 United States Capitol attack & 103 \\
    Tatmadaw & significant event & 2021 Myanmar coup d'\'etat & 103 \\
    January 6 United States Capitol attack & target & United States Congress & 453 \\
    January 6 United States Capitol attack & facet of & presidential transition of Joe Biden & 86 \\
    second impeachment of Donald Trump & facet of & presidential transition of Joe Biden & 86 \\
    second impeachment of Donald Trump & significant person & Donald Trump & 313 \\
    January 6 United States Capitol attack & significant person & Donald Trump & 313 \\
    January 6 United States Capitol attack & has cause & 2020 United States presidential election & 300 \\
    2020 United States presidential election & has effect & January 6 United States Capitol attack & 358 \\
    January 6 United States Capitol attack & has goal & attempts to overturn the 2020 United States presidential election & 507 \\
    January 6 United States Capitol attack & time period & first presidency of Donald Trump & 172 \\
    second impeachment of Donald Trump & defendant & Donald Trump & 511 \\
    \bottomrule
    \end{tabularx}
    \end{table}

    %%%%%%%
    % Table 18: rows hand-picked (submitted); KG ranks from analysis_2021_dip_reissue_20261028.py (numbers JSON "examples")
    %%%%%%%
    \begin{table}
    \centering
    \small
    \caption{Representative \opmintadd{} triples from the 2021 snapshot whose relation type has KG rank $> 50$. The examples center on the January 2021 Capitol attack and the surrounding U.S. presidential transition.}
    \label{tab:appendix:2021-mintadd-examples}
    \begin{tabularx}{\linewidth}{X l X r}
    \toprule
    \textbf{Head} & \textbf{Relation} & \textbf{Tail} & \textbf{KG rank} \\
    \midrule
    % 2020 US presidential election Electoral College count & significant event & January 6 United States Capitol attack & 103 \\
    % Lauren Boebert & significant event & January 6 United States Capitol attack & 103 \\
    % Donald Trump & significant event & second impeachment of Donald Trump & 103 \\
    % January 6 United States Capitol attack & significant person & Donald Trump & 312 \\
    % second impeachment of Donald Trump & significant person & Donald Trump & 312 \\
    % January 6 United States Capitol attack & in opposition to & 2020 US presidential election Electoral College count & 712 \\
    % January 6 United States Capitol attack & has goal & attempts to overturn the 2020 US presidential election & 507 \\
    % January 6 United States Capitol attack & has cause & 2020 United States presidential election & 299 \\
    % second impeachment of Donald Trump & has cause & January 6 United States Capitol attack & 299 \\
    % international reactions to the 2021 storming of the US Capitol & has cause & January 6 United States Capitol attack & 299 \\
    % January 6 United States Capitol attack & target & United States Congress & 452 \\
    % January 6 United States Capitol attack & has effect & second impeachment of Donald Trump & 358 \\
    % January 6 United States Capitol attack & perpetrator & Klete Keller & 643 \\
    % second impeachment of Donald Trump & defendant & Donald Trump & 510 \\
    % timeline of the 2021 storming of the US Capitol & facet of & January 6 United States Capitol attack & 86 \\
    2020 US presidential election Electoral College count & significant event & January 6 United States Capitol attack & 103 \\
    Lauren Boebert & significant event & January 6 United States Capitol attack & 103 \\
    Donald Trump & significant event & second impeachment of Donald Trump & 103 \\
    January 6 United States Capitol attack & significant person & Donald Trump & 313 \\
    second impeachment of Donald Trump & significant person & Donald Trump & 313 \\
    January 6 United States Capitol attack & in opposition to & 2020 US presidential election Electoral College count & 718 \\
    January 6 United States Capitol attack & has goal & attempts to overturn the 2020 US presidential election & 507 \\
    January 6 United States Capitol attack & has cause & 2020 United States presidential election & 300 \\
    second impeachment of Donald Trump & has cause & January 6 United States Capitol attack & 300 \\
    international reactions to the 2021 storming of the US Capitol & has cause & January 6 United States Capitol attack & 300 \\
    January 6 United States Capitol attack & target & United States Congress & 453 \\
    January 6 United States Capitol attack & has effect & second impeachment of Donald Trump & 358 \\
    January 6 United States Capitol attack & perpetrator & Klete Keller & 640 \\
    second impeachment of Donald Trump & defendant & Donald Trump & 511 \\
    timeline of the 2021 storming of the US Capitol & facet of & January 6 United States Capitol attack & 86 \\
    \bottomrule
    \end{tabularx}
    \end{table}

    \textbf{Takeaway.}
    The 2021 dip is therefore not simply a change in model capability over time, but a snapshot-specific distributional shift driven by an increase in KG-rare relation types (associated with contemporaneous events such as the U.S. presidential transition and the January 2021 Capitol attack) in Wikidata. Instances involving these relation types are those on which models score lowest. This suggests that TKGU performance is closely tied to KG relation-type coverage, reinforcing the future-work direction highlighted in the main text (i.e., methods that explicitly access and reason over KG structure).

    \section{Wikidata Qualifiers to Detect Deprecation of Triples}
    \label{sec:appendix:qualifiers}
    The following is the list of Wikidata qualifiers we use to detect the deprecation of triples when creating \datasetname:
    \begin{enumerate}
        \item P582: end time.
        \item P1326: latest date.
        \item P576: dissolved, abolished or demolished date.
        \item P570: date of death.
        \item P730: service retirement.
        \item P2032: work period (end).
        \item P2669: discontinued date.
        \item P3999: date of official closure.
        \item P7125: date of the latest one.
    \end{enumerate}
    \section{Limitations and Future Work}
    \label{sec:appendix:limitations}

    In this work, we focus on leveraging external textual sources to enhance KGs. However, textual data represents only one type of external knowledge. Other modalities, such as video (e.g., podcasts), images, and audio, also contain rich, complementary information that can contribute to KG enrichment. As such, a promising direction for future research is to explore the integration of knowledge from these multimodal sources to address this limitation.

    During the generation of~\datasetname, we use the same temporal delta window for both, the extraction of changes in Wikidata and the emerging passages from Wikipedia. However, certain pieces of knowledge do not always appear within the same time frame in the two sources. For example, events such as Brexit or the election of a president are often documented in Wikipedia months or even years before they are incorporated into the Wikidata knowledge graph.
    \revklim{Such temporal discrepancies can reduce coverage by preventing otherwise valid text-triple alignments. The subsequent LLM-based curation (Section~\ref{sec:quality-control}) flags unsupported TKGU operations, which are excluded from evaluation. Future work could quantify these temporal offsets and extend alignment to adjacent delta windows, followed by LLM-based curation to assess textual support.}
    % In future work, we plan to investigate this temporal discrepancy between the two knowledge sources more thoroughly.

\revklim{Our experimental setup has further limitations. The EDC+ in-context ablation cannot distinguish general few-shot gains from possible exposure to KG content or schema (Appendix~\ref{sec:appendix:icl-ablation}). The higher frequency of \opdeprecate{} in the test set supports per-operation evaluation but makes its operation distribution less representative of the full dataset (Section~\ref{sec:experimental-setup}). Since all metrics are recall-based, over-generating KG candidates (\eg all snapshot triples between linked entities for \opexists{}) can yield high recall, and the 1,628 supported \opdeprecate{} operations cover only 681 distinct triples. Finally, the IE baselines rely on text alone, which is why we complement them with the KG-aware baseline (Section~\ref{sec:experimental-setup}).}

This instantiation of \datasetname{} focuses on KG structure defined by entities and relations, restricting the benchmark to triples where both subject and object are entities in the catalog. Literal-valued attributes, such as dates of birth, lengths, sizes, or employee counts \citep{mesquita2019knowledgenet}, are therefore out of scope. As with the choice of Wikidata as the target KG and Wikipedia hyperlinks for alignment (see Section~\ref{sec:data-collection}), this reflects the current instantiation rather than a limitation of the underlying methodology. Extending to literal-valued triples is a natural next step, e.g., via normalization of temporal expressions, unit-aware matching of numerical values, and fuzzy alignment of string attributes.

    Finally, a limitation of EMERGE is that it covers only the subset of Wikidata changes that can be reliably grounded in Wikipedia text. This stems from the fact that Wikidata is crowdsourced and not fully determined by Wikipedia content, meaning that many Wikidata updates have no corresponding textual evidence. In addition, EMERGE is restricted to Wikipedia paragraphs in which annotated entity mentions can be reliably identified through hyperlinks, as described in Section~\ref{sec:data-collection}. As future work, an alternative dataset could be constructed using text-to-data generation methods \citep{hu2025gptkb, edge2024local, hofer2024construction} to create a synthetic KG that mirrors all knowledge found in text, thereby achieving complete coverage of updates. While such an approach would ensure full alignment between text and KG, it would also introduce challenges such as potential errors in entity disambiguation and the substantial computational cost of generating an entire KG from text.

    Another direction for future work is to incorporate explicit start and end dates for TKGU operations that imply changes in the KG, such as triple addition and deprecation. In the current version of EMERGE, deprecated facts are identified through the delta interval, as our work primarily focuses on updating the KG at a specific point in time rather than modeling full temporal validity. Adding explicit temporal qualifiers would more precisely capture when a fact begins and ceases to hold, aligning EMERGE more closely with the way temporal information is handled in Wikidata. This extension would also enable richer modeling of fact evolution and support downstream methods that rely on explicit temporal boundaries.

    Finally, a further promising future direction is to develop information extraction methods that rely not only on textual evidence to extract triples, but also integrate this content with existing knowledge and patterns in the KG. Such an approach could be particularly beneficial for incorporating emerging entities in \opinfer~TKGU operation, even when they are not supported by textual passages, into the broader KG.~\revklim{Our \textit{KG-aware} baseline highlights two remaining challenges for such methods: retrieving relevant KG context and reasoning over it alongside textual evidence to produce executable updates.}

    \section{Dataset Documentation: Datasheet}
    We describe our dataset following the datasheets for datasets guidelines introduced in \citep{gebru2021datasheets}, detailing its motivation, composition, collection process, and recommended uses.~This documentation supports transparency, reproducibility, and responsible dataset use in machine learning research.

    \subsection{Motivation}
    \paragraph{For what purpose was the dataset created?} The \datasetname~dataset was created to address the lack of integration between changes in textual knowledge and their effect on knowledge graph content. The proposed benchmark enables evaluation of KG updates driven by newly emerging knowledge in textual sources over temporally increasing KG deltas.
    Moreover, because the dataset is generated via an automatic annotation pipeline, it can be continuously extended to include more recent knowledge, thereby allowing evaluation of model robustness to ever-evolving and novel information and KG structures.
    This contrasts with existing benchmarks (see \Tabref{tab:comparison-benchmarks} in the Appendix~\ref{sec:appendix:comparison}), which are static in nature and unable to emulate the evolution of knowledge in textual and KG sources (columns \textit{Evolution-KG} and \textit{Evolution-Text} in \Tabref{tab:comparison-benchmarks}). Furthermore, existing benchmarks do not cover all the necessary text-driven knowledge graph update (TKGU) operations necessary to keep them updated (columns \textsc{\opexists}, \textsc{\opadd}, \textsc{\opmintadd}, \textsc{\opinfer}~and \textsc{\opdeprecate}~in \Tabref{tab:comparison-benchmarks}).

    We expect \datasetname~will encourage the research on methods that are not limited to extracting knowledge from textual sources, but also capable of effectively maintaining KGs by integrating that knowledge into existing KGs. This contrasts with current state-of-the-art IE methods (see Section~\ref{sec:related-work} and \Tabref{tab:comparison_current_architectures}) limited to the extraction of knowledge purely from text without the ability to effectively integrate that knowledge into existing knowledge in KGs.

    \paragraph{Who created the dataset (\eg which team, research group) and on behalf of which entity (\eg company, institution, organization)?} The dataset was developed by academic researchers through an international, cross-institutional collaboration. The contributing researchers bring extensive expertise in information extraction methods and dataset construction.

    \paragraph{Who funded the creation of the dataset?} The dataset was created with funding from, among others, the European Union Marie Curie Actions Postdoctoral Grant.

    \subsection{Composition}
    \label{sec:appendix:dataset-sheet:composition}
    \paragraph{What do the instances that comprise the dataset represent (\eg documents, photos, people, countries)?} The instances that comprise the dataset represent general-domain passages from Wikipedia, KG triples representing the knowledge contained in those passages, and TKGU operations (see Section~\ref{sec:problem-definition}) with respect to the respective general-domain Wikidata KG snapshot.

    \paragraph{How many instances are there in total (of each type, if appropriate)?}
    %%% these number are spread throughout http://localhost:8888/notebooks/src/s14_dataset_stats_v4.ipynb
    % 1449396 TKGU ops
    Our \datasetname~contains in total
    % 233,113
    \statsNuminstances~instances, with a total of \statsTotnrtkgus~TKGU operations: \revklim{610K \opexists, 206K \opadd, 149K \opmintadd, 219K \opinfer, and 9.7K \opdeprecate}. % \statsNuminstances~instances, with a total of 1.45M TKGU operations: 727K \opexists, 240K \opadd, 161K \opmintadd, 313K \opinfer, and 8K \opdeprecate.

    \paragraph{Does the dataset contain all possible instances or is it a sample (not necessarily random) of instances from a larger set?}
    We include a set with all possible instances that can be used for training. For testing (on which we report our results), we subsampled 100 instances per delta per snapshot.

    \paragraph{What data does each instance consist of?} Each of the instances in the dataset consists of a textual passage with an annotated set of entity mentions linked to a particular KG snapshot. In addition, each instance includes a list of triples together with the corresponding TKGU operations that update the KG snapshot, as described in Section~\ref{sec:problem-definition}. Each triple is further annotated with an LLM-based assessment indicating whether the knowledge it represents can be inferred from the textual passage. See Appendix \ref{sec:appendix:prompts} for details on the prompt and examples.
    The dataset spans seven yearly KG snapshots covering 2019-2025. For each snapshot, TKGU updates are annotated over five progressively larger weekly KG deltas, thereby capturing different levels of knowledge staleness in the KG.

    \paragraph{Is there a label or target associated with each instance?} Yes, the target consists of all the triples with corresponding TKGU operations associated with the textual passage of an instance. These operations specify the updates to be applied to a KG snapshot to ensure consistency with the textual passage.

    \paragraph{Is any information missing from individual instances?} All the instances are consistently annotated. However, the triples involved in TKGU operations associated with a passage are restricted to the entities of mentions explicitly annotated with hyperlinks in Wikipedia (see Section~\ref{sec:data-collection} for further details on annotation process). As such, there might be TKGU operations not covered by our dataset. This is also discussed in the limitations sections (see Appendix~\ref{sec:appendix:limitations}).

    \paragraph{Are relationships between individual instances made explicit (\eg users' movie ratings, social network links)?} Yes, all the detected TKGU operations during the annotation process are made explicit. We further mark each of these operations as supported or no by the content of textual passage using LLM automatic annotation process described in Section~\ref{sec:quality-control}.

    \paragraph{Are there recommended data
    splits (\eg training, development/validation, testing)?} Yes. We recommend training and validating models on earlier snapshots (e.g., from 2019 and 2020) and testing on later snapshots (i.e., from 2021–2025). This setup prevents knowledge leakage, since earlier KG snapshots do not contain information from later ones.

    \paragraph{Are there any errors, sources of noise, or redundancies in the dataset?}
    We applied several quality-control measures, including removing duplicate or highly similar passages and filtering out passages with a low proportion of English words, among others described in Section~\ref{sec:quality-control}. In addition, we manually annotated and verified a random subset of the dataset (see Section~\ref{sec:quality-control}). Nevertheless, we do not consider \datasetname~as entirely error-free, as it may contain factual inaccuracies resulting from erroneous edits in Wikipedia or Wikidata. Finally, the annotation agreement scores between the LLMs and human annotators, as well as between humans, are very strong (see Section~\ref{sec:quality-control}) but not perfect, reflecting the complexity and intricacy of error detection in the dataset.

    \paragraph{Is the dataset self-contained, or does it link to or otherwise rely on external resources (\eg websites, tweets, other datasets)?} Yes, the introduced \datasetname~dataset is self-contained and consists of:
    \begin{enumerate}
        \item Annotated instances containing passages with associated KG triples and TKGU operations.
        \item Wikidata KG snapshots to which the annotated TKGU updates are applied.
    \end{enumerate}

    \paragraph{Does the dataset contain data that might be considered confidential (\eg data that is protected by legal privilege or by doctor–patient confidentiality, data that includes the content of individuals' non-public communications)?} No, Wikidata and Wikipedia are public resources.

    \paragraph{Does the dataset contain data that, if viewed directly, might be offensive, insulting, threatening, or might otherwise cause anxiety?} No, no such instances were observed in \datasetname.

    \paragraph{Does the dataset identify any subpopulations (\eg by age, gender)?} While Wikipedia and Wikidata contain entities from various subpopulations, when building \datasetname, we do not focus on identifying and annotating any one in particular.

    \paragraph{Is it possible to identify individuals (that is, one or more natural persons), either directly or indirectly (that is, in combination with other data) from the dataset?} It is possible to identify individuals publicly described in Wikipedia pages or represented in Wikidata entities. However, we do not save other personal information, such as details of the editors involved in Wikipedia and Wikidata updates.
    %  Individuals publicly described in Wikipedia pages or represented in Wikidata entities can be identified. However, \datasetname~does not include other personal information, such as details of the editors involved in Wikipedia or Wikidata updates.

    \paragraph{Does the dataset contain data that might be considered sensitive in any way (\eg data that reveals race or ethnic origins, sexual orientations, religious beliefs, political opinions or union memberships, or locations; financial or health data; biometric or genetic data; forms of government identification, such as social security numbers; criminal history)?} Since Wikipedia and Wikidata are public resources intended to be factual, this concern can be disregarded for the majority of instances in \datasetname.

    \subsection{Collection Process}

    \paragraph{How was the data associated with each instance acquired?~Was the data directly observable (\eg raw text, movie ratings), reported by subjects (\eg survey responses), or indirectly inferred/derived from other data (\eg part-of-speech tags, model-based guesses for age or language)?} The \datasetname~dataset was annotated using publicly available entity mentions in Wikipedia pages, as described in Section~\ref{sec:data-collection}. These hyperlinked mentions are visible to any Wikipedia visitor as links to other pages. To annotate the TKGU operations, we relied on actual updates in Wikidata. Generative models (i.e., LLMs) were used only to verify whether the detected TKGU operations are reflected in the textual content of the passages (see Section~\ref{sec:quality-control}).

    \paragraph{What mechanisms or procedures were used to collect the data (\eg hardware apparatuses or sensors, manual human curation, software programs, software APIs)?} The \datasetname~dataset was generated from the Wikipedia and Wikidata dumps of March 2025. A computing cluster with 64 CPUs and 128 GB of RAM was used to process and generate the dataset. Additionally, a cluster with 4 H100 GPUs was used to run \texttt{Meta-Llama-3.1-405B} for verifying that the TKGU operations are effectively represented in the textual passages (see Section~\ref{sec:quality-control}).

    \paragraph{If the dataset is a sample from a larger set, what was the sampling strategy (\eg deterministic, probabilistic with specific sampling probabilities)?}
    The test set used in our experiments was randomly sampled from the larger dataset, with exactly 100 instances per snapshot per KG delta. The sampling procedure, described in detail in Section~\ref{sec:experimental-setup}, first retains up to 40 instances per delta for each operation that requires an actual update to the KG (i.e., \opdeprecate, \opadd, \opmintadd, and \opinfer). This ensures that the models are evaluated on a sufficiently large number of such instances. This is particularly important for \opdeprecate~TKGU operations, which are very scarce in the original dataset; without this retention, a purely random subsample would contain only a few instances, potentially leading to high variability in the results.

    \paragraph{Who was involved in the data collection process (\eg students, crowdworkers, contractors) and how were they compensated (\eg how much were crowdworkers paid)?}
    The dataset was generated automatically from real-world updates to Wikidata and changes in Wikipedia articles. LLMs were used to assess each TKGU operation with respect to the knowledge contained in the textual passages. The only human involvement was the annotation of a subsample of the dataset to measure agreement with the LLM annotations. For this purpose, two researchers acted as annotators and were credited as co-authors of the paper.

    \paragraph{Over what timeframe was the data collected?} The data were collected from seven yearly snapshots, spanning January 1, 2019, to January 1, 2025. For each snapshot, KG deltas were extracted for up to five weeks, ending on February 5 of the corresponding year.

    \paragraph{Were any ethical review processes conducted (\eg by an institutional review board)?} No, the public nature of the data, consisting of Wikipedia pages and Wikidata KG updates, meant that no formal ethical review was required.

    \paragraph{Did you collect the data from the individuals in question directly, or obtain it via third parties or other sources (\eg websites)?} The data were obtained from publicly available Wikipedia and Wikidata repository dumps (\url{https://dumps.wikimedia.org/}).

    \paragraph{Were the individuals in question notified about the data collection?} No individuals were directly involved in the data collection.

    \paragraph{Did the individuals in question consent to the collection and use of their data?} No individuals were directly involved in the data collection.

    \paragraph{If consent was obtained, were the consenting individuals provided with a mechanism to revoke their consent in the future or for certain uses?} No individuals were directly involved in the data collection.

    \paragraph{Has an analysis of the potential impact of the dataset and its use on data subjects (\eg a data protection impact analysis) been conducted?} No formal data protection impact analysis was conducted, as the dataset is derived entirely from publicly available Wikipedia pages and Wikidata KG updates and does not include private or sensitive information about individuals.

    \subsection{Preprocessing/Cleaning/Labeling}
    \label{sec:appendix:cleaning}
    \paragraph{Was any preprocessing/cleaning/labeling of the data done (\eg discretization or bucketing,
    tokenization, part-of-speech tagging,
    SIFT feature extraction, removal of
    instances, processing of missing values)? }
    Yes. The original raw data from the Wikipedia and Wikidata dumps underwent several preprocessing steps:
    \begin{enumerate}
        \item Preprocessed Wikipedia wikitext, retaining only lists and textual paragraphs as dataset inputs, while excluding tables, figures, and other multimodal elements.
        \item Extracted only Wikipedia text containing explicitly annotated entity mentions by editors, which could be mapped to Wikidata updates within a given time window in the KG delta.
        \item Constrained Wikipedia passages to lengths between 30 and 1,000 tokens.
        \item Filtered out passages with fewer than 30\% English words, using the Python \texttt{nltk} package.
        \item Applied stability constraints by discarding changes in Wikidata and Wikipedia that were quickly rolled back (often indicating incorrect knowledge). Specifically, we retained Wikidata KG updates persisting at least 7 days and Wikipedia edits not followed by another change within 30 minutes.
        \item Ensured diversity by requiring passages aligned to similar updates in Wikipedia to differ in content, measured by edit distance (minimum 0.15 for texts under 2,500 characters and 0.25 for texts 2,500 characters or longer).
        \item Validated the alignment of TKGU operations to textual passages with LLMs, explicitly marking operations that could be grounded in the passage content (see Section~\ref{sec:quality-control} for further details).
        \end{enumerate}

    \paragraph{Was the ``raw'' data saved in addition to the preprocessed/cleaned/labeled data (\eg to support unanticipated future uses)?} Yes. We preserved all input and output data from each preprocessing step, beginning with the raw Wikipedia and Wikidata dumps used to construct \datasetname.

    \paragraph{Is the software that was used to preprocess/clean/label the data available?} Yes, all the software that was used to preprocess/clean/label can be accessed through the public \datasetname{} repository (\url{https://github.com/klimzaporojets/emerge}).

    \subsection{Uses}

    \paragraph{Has the dataset been used for any tasks already?} Yes, in Section~\ref{sec:experimental-setup} we experiment with various current state-of-the-art information extraction models.

    \paragraph{Is there a repository that links to any or all papers or systems that use the dataset?} Yes, \url{https://github.com/klimzaporojets/emerge}.

    \paragraph{What (other) tasks could the dataset be used for?}
    Beyond the KG updating task presented in this paper, \datasetname~could be directly applied to at least the following tasks:
    \begin{enumerate}
        \item Question answering over novel and emerging knowledge derived from the TKGU operations introduced here.
        \item General knowledge graph completion, where certain changes may trigger additional updates that are not limited to entities mentioned in textual passages but instead depend on the evolving KG structure. To support this, we will release all KG changes, not only those aligned with textual passages, which form the core of \datasetname.
    \end{enumerate}

    \paragraph{Is there anything about the composition of the dataset or the way it was collected and preprocessed/cleaned/labeled that might impact future uses?} No.

    \paragraph{Are there tasks for which the dataset should not be used?} No.

    \subsection{Distribution}

    \paragraph{Will the dataset be distributed to third parties outside of the entity (\eg company, institution, organization) on behalf of which the dataset was created?} \revklim{Yes, the dataset is publicly available on Hugging Face.}

    \paragraph{How will the dataset be distributed (\eg tarball on website, API, GitHub)?}
    % The \datasetname~dataset will be distributed via Hugging Face (\url{https://huggingface.co/}), and the code for generating the dataset will be released on GitHub (\url{https://github.com/}).
    \revklim{The \datasetname~dataset is distributed via Hugging Face (\url{https://huggingface.co/datasets/klimzaporojets/emerge-benchmark}), and the code for generating the dataset is released on GitHub (\url{https://github.com/klimzaporojets/emerge}).}

    \paragraph{When will the dataset be distributed?}
    The \datasetname~dataset is publicly released on Hugging Face (\url{https://huggingface.co/datasets/klimzaporojets/emerge-benchmark}).

    \paragraph{Will the dataset be distributed under a copyright or other intellectual property (IP) license, and/or under applicable terms of use (ToU)?}
    To support openness and collaboration in research, we release the datasets under the Creative Commons % Attribution 4.0 International
    Attribution-ShareAlike 4.0 International (CC BY-SA 4.0) license. The full terms of this license can be found on the Creative Commons website: \url{https://creativecommons.org/licenses/by-sa/4.0/}.

    \paragraph{Have any third parties imposed IP-based or other restrictions on the data associated with the instances?} No, the dataset is derived from publicly available Wikipedia and Wikidata knowledge repositories and is not subject to any third-party IP restrictions.

    \paragraph{Do any export controls or other regulatory restrictions apply to the dataset or to individual instances?}
    No, the dataset and its individual instances are based on publicly available Wikipedia and Wikidata content and are not subject to export controls or other regulatory restrictions.

    \subsection{Maintenance}
    \paragraph{Who will be supporting/hosting/maintaining the dataset?} The dataset will be supported, hosted, and maintained by the authors of this paper.

    \paragraph{How can the owner/curator/manager of the dataset be contacted (\eg email address)?}
    The dataset is curated and managed by the authors of this paper. Inquiries regarding the dataset, including access, usage, and reporting issues, can be directed to the corresponding authors via email. Additionally, users can submit questions or report issues through the GitHub repository hosting the dataset generation code.

    \paragraph{Is there an erratum?}
    \revklim{Yes. Corrections are documented in the changelog on the dataset card (\url{https://huggingface.co/datasets/klimzaporojets/emerge-benchmark}).}
    % No erratum has been issued for the \datasetname~dataset.

    \paragraph{Will the dataset be updated (\eg to correct labeling errors, add new instances, delete instances)?}
    Yes, \datasetname~will be regularly updated with emerging knowledge through yearly snapshots. Announcements regarding new versions will be communicated via the \datasetname~GitHub repository. Additionally,
    % as described in \cref{sec:dataset-extension},
    users can generate customized versions of \datasetname~by adjusting relevant hyperparameters, as well as personalized snapshots of different granularity (\eg daily, weekly, monthly).

    \paragraph{If the dataset relates to people, are there applicable limits on the retention of the data associated with the instances (\eg were the individuals in question told that their data would be retained for a fixed period of time and then deleted)?}

    The \datasetname~dataset does not contain private or personally identifiable information about individuals. It is derived entirely from publicly available Wikipedia pages and Wikidata entities, and no retention limits for individual consent were applicable.

    \paragraph{Will older versions of the dataset continue to be supported/hosted/maintained?}
    Yes, all previous versions of \datasetname~will continue to be supported, hosted, and maintained. Each version will be assigned a unique version number, and we will provide persistent links to access every version through Hugging Face storage server. This will ensure reproducibility of experiments and will enable users to reference or use specific dataset versions as needed.

    \paragraph{If others want to extend/augment/build on/contribute to the dataset, is there a mechanism for them to do so?}
    Yes.
    % As described in \cref{sec:dataset-extension},
    \datasetname~users will have access to all necessary scripts to re-generate the dataset with customized settings. This includes adjusting hyperparameters such as the maximum passage length, generating the dataset for newer snapshots, and specifying the number and granularity of KG deltas.

\section{Compute Resources}
\label{sec:appendix:compute-resources}

\textbf{Dataset construction (one-time, not required to use \datasetname{}).} The end-to-end pipeline that produced \datasetname{} runs on a SLURM cluster equipped with NVIDIA H100 80\,GB GPUs. The most expensive stage is the LLM curation pass (Section~\ref{sec:quality-control}), which serves \texttt{Meta-Llama-3.1-405B} via Hugging Face Text Generation Inference (TGI) 3.3.4 across 4$\times$H100 80\,GB GPUs per node. The 233\,K-passage corpus was sharded into seven jobs (one per yearly snapshot, 2019--2025), each consuming roughly 24 hours of wall-clock on a single 4-GPU node, for a total of roughly 7$\times$24$\times$4 $\approx$ 670 H100-GPU-hours; nodes additionally request 100\,GB system memory and 10 CPU cores per task. Earlier (smaller) Llama-3.1-8B passes used the same hardware profile with a single GPU per shard. The remaining pipeline stages (Wikidata/Wikipedia delta extraction, alignment, deduplication, subsampling, statistics) are CPU-bound and run in a few minutes to a few hours on a standard compute node. The full SLURM batch wrappers (partition, GPU count, memory, wall-clock budget) are released in the public github repository (\url{https://github.com/klimzaporojets/emerge}).

\textbf{Benchmark evaluation.} Once \datasetname{} is published, downstream evaluation does not require the dataset-construction compute. Evaluation needs only 1$\times$GPU with 16\,GB+ VRAM and 32\,GB system memory (180\,GB+ if loading full Wikidata KG snapshots into memory). Concretely, our reported numbers were obtained with: a single A100 GPU for ReLiK and REBEL inference on the 3{,}500-instance test set ($\sim$5 hours on A100 for ReLiK); LLM-API calls for the LLM-backed baselines using \texttt{GPT-5.1} (OpenAI), \texttt{Mistral-Large-2411} and \texttt{Mistral-small} (Mistral via Azure), with no author-side GPU usage; the ICL/zero-shot ablation (Appendix~\ref{sec:appendix:icl-ablation}) reuses the same API calls. Total API budget for the full benchmark sweep is on the order of a few hundred USD, dominated by GPT-5.1.

\section{Broader Impact}
\label{sec:appendix:broader-impact}

\textbf{Positive impacts.} \datasetname{} supports research on keeping public knowledge bases (e.g., Wikidata) in sync with newly emerging textual evidence, an underserved direction relative to text-only IE benchmarks. Advances in text-driven KG-update methods improve the timeliness, Completeness, and verifiability of public knowledge sources used in downstream applications such as fact-checking, search, question answering, and education. The dataset is built end-to-end from openly licensed public sources (Wikipedia under CC BY-SA 3.0+, Wikidata under CC0), and is released under CC BY-SA 4.0 with documentation (Appendix~\ref{sec:appendix:dataset-sheet:composition}, Croissant + RAI metadata) so it can be extended to other KGs and other languages with minimal pipeline change (see also Section~\ref{sec:our-dataset} on extensibility).

\textbf{Potential negative impacts and mitigations.}~Three considerations are worth flagging. First, models that automatically propose KG edits could, if deployed without human review, propagate errors from text into curated knowledge bases; we
do not advocate fully unsupervised KG maintenance. Second, our LLM-curation step inherits any biases in the underlying LLM (\texttt{Meta-Llama-3.1-405B}); we mitigate by reporting human--LLM agreement statistics on a 500-pair sample (Appendix~\ref{sec:appendix:human-annotation-agreement}). This way, users can audit the reliability of the curation labels. Third, because the dataset is grounded in Wikipedia/Wikidata, sub-population coverage inherits the known coverage gaps of those sources \citep{abian2022analysis, field2022controlled}.

    %%%%%%%%%%%%%%%%%%%%%%%%%%%%%%%%%%%%%%%%%%%%%%%%%%%%%%%%%%%%

    \end{document}